\documentclass[Afour,sageh,times]{sagej}
\usepackage[ruled,vlined]{algorithm2e}  
\usepackage{amsmath,amsfonts,amssymb}
\usepackage{array}
\usepackage{textcomp}
\usepackage{stfloats}
\usepackage{url}
\usepackage{verbatim}
\usepackage{graphicx}
\usepackage{moreverb,url}
\usepackage{times}
\usepackage{epsfig}
\usepackage{subfigure}
\usepackage{multirow}
\usepackage{multicol}
\usepackage{caption}
\usepackage{gensymb}
\usepackage[colorlinks,bookmarksopen,bookmarksnumbered,citecolor=red,urlcolor=red]{hyperref}
\usepackage{mathtools}
\usepackage{booktabs}
\usepackage{pifont}
\usepackage{xspace}
\usepackage{tabularx} 
\usepackage{enumerate}
\usepackage{enumitem} 
\usepackage{bm}
\newcommand{\trsp}{{\scriptscriptstyle\top}}

\usepackage{pifont}
\usepackage[table]{xcolor}

\newcommand\BibTeX{{\rmfamily B\kern-.05em 
\textsc{i\kern-.025em b}\kern-.08em
T\kern-.1667em\lower.7ex\hbox{E}\kern-.125emX}}

\begin{document}


\title{A Riemannian Take on Distance Fields and Geodesic Flows in Robotics}

\author{Yiming Li\affilnum{1,2,*}, Jiacheng Qiu\affilnum{1,*} and Sylvain Calinon\affilnum{1,2}}

\affiliation{\affilnum{1}Idiap Research Institute, Switzerland\\
\affilnum{2}École Polytechnique Fédérale de Lausanne (EPFL), Switzerland\\
\affilnum{*}Equal Contribution}

\corrauth{Yiming Li, Idiap Research Institute, Centre du Parc, Rue Marconi 19, CH-1920 Martigny, Switzerland.}

\email{yiming.li@idiap.ch}

\begin{abstract}
Distance functions are crucial in robotics for representing spatial relationships between a robot and its environment. They provide an implicit, continuous, and differentiable representation that integrates seamlessly with control, optimization, and learning. While standard distance fields rely on the Euclidean metric, many robotic tasks inherently involve non-Euclidean structures. To this end, we generalize Euclidean distance fields to more general metric spaces by solving the Riemannian eikonal equation, a first-order partial differential equation whose solution defines a distance field and its associated gradient flow on the manifold, enabling the computation of geodesics and globally length-minimizing paths. We demonstrate that \emph{geodesic distance fields}—the classical Riemannian distance function represented as a global, continuous, and queryable field—are effective for a broad class of robotic problems where Riemannian geometry naturally arises. To realize this, we present a neural Riemannian eikonal solver (NES) that solves the equation as a mesh-free implicit representation without grid discretization, scaling to high-dimensional robot manipulators. Training leverages a physics-informed neural network (PINN) objective that constrains spatial derivatives via the PDE residual and boundary/metric conditions, so the model is supervised by the governing equation and requires no labeled distances or geodesics. We propose two NES variants, conditioned on boundary data and on spatially varying Riemannian metrics, underscoring the flexibility of the neural parameterization. We validate the effectiveness of our approach through extensive examples, yielding minimal-length geodesics across diverse robot tasks involving Riemannian geometry. Additionally, we validate the method in a dynamics-aware motion-planning task for energy-efficient trajectory generation, with comparisons to baseline approaches.\\
Project website: \href{https://sites.google.com/view/geodf}{https://sites.google.com/view/geodf}
\end{abstract}

\keywords{Differential Geometry, Riemannian Manifold, Distance Field, Geodesic, Eikonal Equation, Gradient Flow, \textcolor{black}{Energy Conservation}}

\maketitle

\section{Introduction}

In robotics, measuring distances constitutes a fundamental concept for determining spatial relationships and enabling effective physical and non-physical interactions with the environment. These metrics provide a systematic means for quantifying the geometric relationships between various entities, such as points, poses, shapes or trajectories. They are widely applicable across robotic tasks, including inverse kinematics \citep{chiacchio1991closed} and motion planning \citep{ratliff2009chomp}. Signed distance fields (SDFs), in particular, have gained popularity for representing geometries using implicit functions, as they enable efficient distance and gradient queries which are suitable to integrate into learning \citep{weng2022neural}, optimization \citep{li2024representing} and control \citep{liu2022regularized}.

SDFs are conventionally employed in Euclidean spaces, representing the shortest distance from any point in the environment to the boundary of a given object or surface \citep{park2019deepsdf}. However, many robot tasks inherently operate in non-Euclidean spaces, with manifolds that can be described implicitly by a smoothly varying weighting matrix, which locally measures distances. For example, distance fields can also be applied to joint configuration space, indicating the minimum joint motion required by the robot to establish contact with a given point or object \citep{Li24CDF}. In this case, \emph{geodesic distance fields} enable the consideration of inertia, stiffness, or manipulability ellipsoids in the processing, by providing a Riemannian metric constructed with a smoothly varying, symmetric, positive definite (SPD) weighting matrix in the robot configuration space. Riemannian geometry provides a principled and systematic way to generalize algorithms from Euclidean spaces to more general manifolds \citep{calinon2020gaussians}. Figure~\ref{fig:intro} shows minimal-distance paths in Euclidean space and on a Riemannian manifold. In the latter, geodesics are shaped by the manifold’s geometry—for example, when the Riemannian metric is defined by the robot’s inertia  matrix to reflect its dynamic properties.

\begin{table}[t]
    \begin{center}
        \begin{tabular}{cc}
            \centering
            \includegraphics[width =0.41\linewidth
            ]{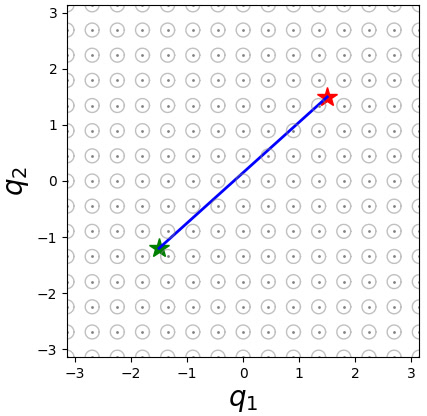}   &
            \includegraphics[width =0.55\linewidth]{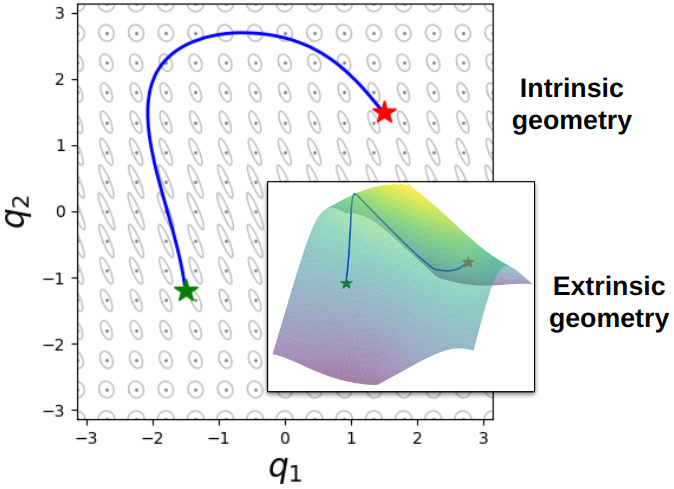}
 \\
            (a) & (b)\\
        \end{tabular}
        \captionof{figure}{Minimal distance paths as geodesics in the Euclidean space (a) and in another Riemannian metric space (b). The ellipses depict the SPD weighting matrices used to locally compute distances with this metric (isocontours of inverse matrices). A Riemannian manifold can be described intrinsically by the depicted metric. For visualization, it can also be depicted with corresponding extrinsic geometry in a higher-dimensional space (see inset), but geodesic computation does not require this construction and instead only requires the metric as an intrinsic geometry representation.}
        \label{fig:intro}
    \end{center}
\end{table}

\begin{figure}
    \centering
    \includegraphics[width=\linewidth,trim=4pt 2pt 2pt 2pt,clip]{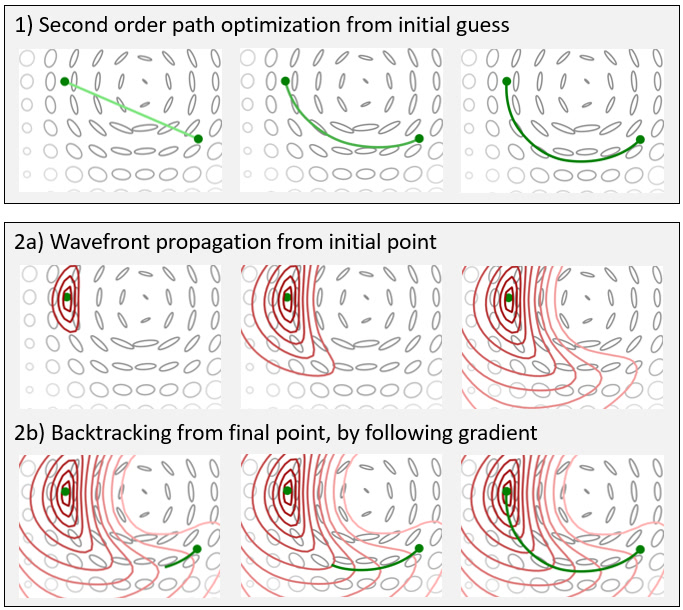}
    \caption{\textcolor{black}{Two approaches for computing geodesics on inhomogeneous manifolds. The traditional method (1) formulates the problem as an iterative solution to a differential equation, often using second-order optimization (e.g., Gauss-Newton path optimization). This approach requires a good initial guess and must be solved separately for each point pair. In contrast, we propose a wavefront propagation approach that first computes a geodesic distance field from a source point by solving the Riemannian eikonal equation (2a), then retrieves geodesics by backtracking along the gradient of this field (2b). We employ physics-informed neural networks (PINNs) to solve the eikonal equation, enabling scalable solutions in high-dimensional settings. This method encodes the manifold’s intrinsic geometry and yields globally optimal geodesics. It also offers modularity and efficiency: the distance field can be trained offline and used online for fast distance and geodesic queries.
    }}
    \label{fig:compare_eik_opt}
\end{figure}

While the geometric nature of robot problems on Riemannian manifolds is well established, many methods for computing geodesics operate on prescribed endpoint pairs and rely on local optimization \citep{jaquier2022riemannian,klein2023design,cabrera2024optimal}. These approaches are computationally intensive, must be re-run for every new endpoint pair, and do not provide a global representation of the manifold. \textcolor{black}{As an alternative, we represent the manifold with a \emph{geodesic distance field}, a global and continuous value function that implicitly encodes geodesics \citep{CraneGeodesicSurvey2020} and from which minimal paths are recovered by backtracking. However, constructing such fields on Riemannian manifolds is challenging due to nonlinear metrics and complex topology; classical geodesic ray tracing and geodesic ordinary differential equation (ODE) integration/shooting are computationally expensive, sampling dependent, and lack global guarantees \citep{rawlinson2005fast}.} Motivated by wavefront propagation methods that are well established for distance fields in Euclidean space, we compute the distance field by solving the Riemannian eikonal equation, a first-order partial differential equation (PDE) for wave travel time, which yields an efficient and globally consistent single source solution \citep{KimmelSethian1998}. This perspective is closely related to recent implicit shape models based on SDFs, which appear as viscosity solutions of the eikonal equation \citep{gropp2020implicit,Maric24}. The resulting field captures the manifold’s global structure, provides gradient information for fast queries, and reveals geodesic flow for straightforward trajectory backtracking. Beyond geodesic computation, this representation enables learning, optimization, and control methods based on distance fields on the manifold. An illustration is given in Figure~\ref{fig:compare_eik_opt}.

Classical PDE solvers for wavefront propagation include the Fast Marching Method (FMM) \citep{sethian1996fast} and fast sweeping methods \citep{zhao2005fast}, widely used in level set and Hamilton–Jacobi settings \citep{osher2004level}. \textcolor{black}{FMM computes the entire single-source distance field in a single, monotone outward pass, after which geodesics to any target are obtained by inexpensive backtracking.} These schemes run on discretized grids or meshes and yield deterministic solutions under standard conditions, though fixed resolution limits scalability in high-dimensional spaces. We build on this framework along two axes. First, we generalize the eikonal equation from Euclidean spaces to Riemannian manifolds, enabling geodesic distance fields and flows in robot configuration spaces with curved, non-Euclidean geometry. Second, to overcome grid scalability, we introduce a neural Riemannian eikonal solver (NES) based on physics-informed neural networks (PINN) \citep{raissi2019physics,kelshaw2024computing}. NES replaces the grid with a coordinate-based neural representation, computes gradients via automatic differentiation, and yields a continuous field that can be queried at arbitrary resolution. Crucially, it can be conditioned on start–goal pairs to produce geodesic flows— which is a challenging problem for fixed-source PDE solvers. Once trained, NES outputs arbitrary state-to-goal distance queries with millisecond-level latency. Training is self-supervised by the eikonal constraint and boundary conditions, avoiding precomputed distance labels. \textcolor{black}{We also introduce two NES variants that can be conditioned on boundary points or on the Riemannian metric, enabled by the flexibility and differentiability of neural network representation.} In summary, our approach leverages deep learning to solve the Riemannian eikonal equation, delivering a scalable, flexible, and real-time representation of distance fields and geodesic flows that integrates naturally with learning, optimization, and control in robotics.

{\color{black}
To demonstrate the advantages of our neural Riemannian eikonal solver for distance fields and geodesic flows, we present examples across common robotics geometries: energy-aware (kinetic/potential) metrics for dynamics-aware motion, pullback metrics for task-space minimization, and tailored metrics for stability shaping and collision avoidance.} We also study dynamics-aware motion formulated as geodesic minimization under an energy-aware metric that encodes the system dynamics. \textcolor{black}{NES computes geodesic flows efficiently for arbitrary source–target pairs, providing a global, geometry-aware prior toward lower-effort motions; in practice, the resulting shortest paths often align with energy-efficient trajectories and are easier to track.
} We demonstrate efficient planning and simple tracking on planar robots and a 7-DoF Franka arm, highlighting scalability, efficiency, and flexibility.
The main contributions are summarized as follows:

{\color{black}
\begin{itemize}
  \item We propose solving the \emph{eikonal equation} to obtain distance fields and geodesic flows for robot problems on Riemannian manifolds. Unlike geodesic shooting or optimal-control methods, this approach reframes geodesic computation as a \emph{single global} implicit representation of the manifold, enabling fast backtracking and direct integration with planning and control.

  \item We introduce a Neural Eikonal Solver (NES) that learns the eikonal PDE in a physics-informed manner without distance labels, yielding a continuous, differentiable field with efficient gradient queries. NES scales to high-dimensional spaces beyond grid-based solvers. We also present two variants conditioned on boundary data and on spatially varying Riemannian metrics, underscoring the flexibility of the neural parameterization.

  \item We provide extensive examples across kinematics, dynamics, motion planning, and control. On energy-aware manifolds, NES yields shorter paths under the chosen Riemannian metric and, in our experiments, achieves lower energy costs than baselines on high-dimensional systems, with both quantitative and qualitative comparisons.

\end{itemize}
}

The rest of the article is structured as follows: Section~\ref{sec:background} introduces the necessary mathematical background, and Section~\ref{sec:rw} reviews related work. Section~\ref{sec:res} describes our NES for solving the Riemannian eikonal equation to compute geodesic distance fields and flows.  Section~\ref{sec:examples} presents extensive examples using our approach across robot problems involving Riemannian geometry. Section~\ref{sec:exp} reports experimental results on dynamics-aware motion generation, demonstrating the effectiveness of our method. Finally, Section~\ref{sec:conclusion} concludes the paper and outlines potential applications and future directions.

\section{Background}\label{sec:background}

In this section, we introduce the mathematical background of Riemannian manifolds, geodesics, the eikonal equation \textcolor{black}{and introduce commonly used Riemannian metrics in robotics.}

\subsection{Riemannian Metrics and Geodesics}
A $d$-dimensional Riemannian manifold $\mathcal{M}$ is a topological space equipped with a smooth metric tensor $\bm{G}(\bm{x})$ defined at each point $\bm{x} \in \mathcal{M}$. The metric tensor $\bm{G}(\bm{x})$ is a symmetric positive definite matrix that defines the Riemannian metric, allowing us to calculate distances and angles on the manifold. For each point $\bm{x} \in \mathcal{M}$, there exists a tangent space $T_{\bm{x}} \mathcal{M}$, which locally linearizes the manifold.

The inner product of two velocity vectors, $\bm{u}$ and $\bm{v}$, in the tangent space $T_{\bm{x}} \mathcal{M}$ at a point $\bm{x} \in \mathcal{M}$ is given by
\begin{equation}
    \begin{aligned}
    \label{eq: inner}
    \langle \bm{u}, \bm{v} \rangle_{\bm{G}}= \bm{u}^\trsp \bm{G}(\bm{x}) \bm{v}.
    \end{aligned}
\end{equation}
Using this inner product, we define the Riemannian norm of a vector $\bm{u}$ as
\begin{equation}
    \label{eq: norm}
    \| \bm{u} \|_{\bm{G}} = \sqrt{\langle \bm{u}, \bm{u} \rangle_{\bm{G}}}.
\end{equation}
These definitions allow us to measure vector lengths and angles within the tangent space $T_{\bm{x}} \mathcal{M}$. With this, we can define the Riemannian distance between two points, $\bm{x}_1$ and $\bm{x}_2$, on the given manifold as
\begin{equation}
    \label{eq:geodesic}
    \begin{aligned}
        U(\bm{x}_1, \bm{x}_2) = \int_{s_0}^{s_1} \| \dot{\gamma}(s) \|_{\bm{G}(\gamma(s))}\;\mathrm{d}s,
    \end{aligned}
\end{equation}
where $\gamma(s)$ is a smooth curve connecting $\bm{x}_1$ and $\bm{x}_2$, with $\gamma(s_0) = \bm{x}_1$ and $\gamma(s_1) = \bm{x}_2$. The minimization of this expression allows us to define geodesic distances and shortest geodesic paths between two points on the manifold. 

\subsection{Eikonal Equation}
\label{sec:eikonal}
\subsubsection{Isotropic eikonal equation}

The eikonal equation is a nonlinear first-order partial differential equation (PDE) that models wavefront propagation. \textcolor{black}{Let \(\Omega\subset\mathbb{R}^n\) be the domain, \(\bm{x}_1\in\Omega\) a prescribed source, and \(c:\Omega\to(0,\infty)\) a positive scalar function specifying the travel speed.} The standard form is
\begin{equation}
\label{eq:eikonal}
\|\nabla U(\bm{x}_2)\| = c(\bm{x}_2) \quad \text{s.t.} \quad U(\bm{x}_1) = 0,
\end{equation}
where \(U:\Omega\to\mathbb{R}\) is the distance-to-source function. Here, \(U(\bm{x}_2)\) is used as a notation instead of \(U(\bm{x}_1,\bm{x}_2)\) when the source \(\bm{x}_1\) is fixed, corresponding to the geodesic \(\gamma\):
\begin{equation}
U(\bm{x}_2) = \min_{\gamma} \int_0^1 c(\gamma(s)) \left\|\dot{\gamma}(s)\right\| \, ds,
\end{equation}
Here \(\gamma:[0,1]\to\Omega\) is a path parameterized by \(s\in[0,1]\) with endpoints \(\gamma(0)=\bm{x}_1\) (source) and \(\gamma(1)=\bm{x}_2\) (target), and \(\dot{\gamma}(s)\) denotes its velocity.
\textcolor{black}{\(c(\cdot)\) specifies the local speed that shapes the arrival time/distance.} The function \(U(\bm{x}_2)\) with fixed source \(\bm{x}_1\) is the \textit{single-point} solution; \textcolor{black}{for global distance fields, one can encode \(U(\bm{x}_1,\bm{x}_2)\) over arbitrary \textit{source–goal} pairs.} 

\subsubsection{Riemannian eikonal equation}
\label{subsecsec:REE}
The isotropic eikonal equation describes the distance field and yields shortest paths in Euclidean space. \textcolor{black}{This generalizes to Riemannian manifolds \citep{mirebeau2017anisotropic, peyre2010geodesic}}:


\begin{equation} \label{eq: eikonal_riemannian}
\|\nabla U(\bm{x}_2)\|_{\bm{G}^{-1}(\bm{x}_2)} = 1 \quad \text{s.t.} \quad U(\bm{x}_1) = 0,
\end{equation}
which characterizes wavefront propagation and minimal distance on a manifold endowed with a Riemannian metric defined by $\bm{G}$. \textcolor{black}{Unless noted otherwise, we adopt the unit-weight case $c(\bm{x}_2)=1$; any strictly positive field can be absorbed by rescaling the metric.} The corresponding geodesic distance is
\begin{equation} \label{eq: eikonal_riemannian_velocity}
U(\bm{x}_2) \;=\; \min_{\gamma}\; \int_0^1 \big\|\dot{\gamma}(s)\big\|_{\,\bm{G}(\gamma(s))}\, ds,
\end{equation}
\textcolor{black}{which is the minimal Riemannian distance~\eqref{eq:geodesic}. Equation~\eqref{eq: eikonal_riemannian} measures the norm of the covector $\nabla U$ with the inverse matrix $\bm{G}^{-1}$, whereas \eqref{eq: eikonal_riemannian_velocity} measures the speed of the vector $\dot{\gamma}$ with the metric tensor $\bm{G}$. These are \emph{dual norms}, induced by $\bm{G}$ on vectors and by $\bm{G}^{-1}$ on covectors. This dual pairing underlies the Hamilton–Jacobi characterization of Riemannian distance \citep{mantegazza2002hamilton} and is made explicit in dual-metric formulations for eikonal solvers and geodesic backtracking \citep{mirebeau2017anisotropic}.}

\textcolor{black}{To backtrack geodesics, we can define the geodesic flow:}
\begin{equation}\label{eq:riem-direction-V}
\bm{V}(\bm{x}_2) := \bm{G}^{-1}(\bm{x}_2)\,\nabla U(\bm{x}_2).
\end{equation}
\textcolor{black}{From \eqref{eq: eikonal_riemannian} it follows that $\bm{V}$ is unit-speed with respect to $\bm{G}$:}
\begin{equation}\label{eq:unit-speed-G-V}
\resizebox{.98\linewidth}{!}{$
\begin{aligned}
\|\bm{V}(\bm{x}_2)\|_{\bm{G}(\bm{x}_2)}^2
&= \bm{V}(\bm{x}_2)^{\!\top}\,\bm{G}(\bm{x}_2)\,\bm{V}(\bm{x}_2) \\
&= \big(\bm{G}^{-1}(\bm{x}_2)\,\nabla U(\bm{x}_2)\big)^{\!\top}\,\bm{G}(\bm{x}_2)\,\big(\bm{G}^{-1}(\bm{x}_2)\,\nabla U(\bm{x}_2)\big) \\
&= \nabla U(\bm{x}_2)^{\!\top}\,\bm{G}^{-1}(\bm{x}_2)\,\nabla U(\bm{x}_2) \\
&= 1.
\end{aligned}
$}
\end{equation}
\textcolor{black}{Given a target $\bm{x}_2$ and source $\bm{x}_1$, a geodesic is recovered by integrating the flow backward in time}
\begin{equation}\label{eq:backtrack-V}
\dot{\gamma}(s) = -\,\bm{V}\big(\gamma(s)\big), \qquad \gamma(0)=\bm{x}_2,\ \ \gamma(T)=\bm{x}_1.
\end{equation}
\textcolor{black}{Along this backward flow, $U$ decreases at a unit rate. Starting from $U(\bm{x}_2)$ and terminating at the source where $U(\bm{x}_1)=0$, the elapsed time equals the initial value, i.e., $T=U(\bm{x}_2)$.}

A common numerical approach for solving eikonal equations is the Fast Marching Method (FMM). Like Dijkstra's algorithm, FMM relies on discretization and sequentially propagates information from the boundary or solved nodes on the mesh~\citep{sethian1996fast}. This method extends to non-Euclidean domains, allowing computation of geodesic distances on manifolds~\citep{KimmelSethian1998}. More recent work leverages machine learning to approximate solutions, producing continuous, differentiable distance fields in Euclidean spaces~\citep{grubas2023neural} as well as on nonlinear manifolds~\citep{kelshaw2024computing}.

{\color{black}
\subsection{Common Riemannian Metrics in Robotics}
\label{sec:riemann-props}

In this section, we introduce commonly used Riemannian metrics that furnish a geometric perspective on robotic problems. These include metrics that encode intrinsic kinematic and dynamic properties of the robot, as well as task-shaped designs. We also clarify the practical meaning of geodesics on the resulting manifolds and indicate how they can be computed via solving the Riemannian eikonal equation.
}

\subsubsection{Kinetic–energy metric.}
Let \(\bm{q}\) denote the joint configuration and \(\bm{M}(\bm{q})\) the inertia matrix. The kinetic–energy metric is the smooth, symmetric positive-definite matrix that defines the Riemannian metric \citep{bullo2004geometric}
\begin{equation}
\bm{G}_{\mathrm{ke}}(\bm{q}) \coloneqq \bm{M}(\bm{q}).
\end{equation}
The kinetic energy is defined as
\begin{equation}
T(\bm{q},\dot{\bm{q}})
=\tfrac{1}{2}\,\dot{\bm{q}}^{\top} \bm{M}(\bm{q})\,\dot{\bm{q}}
=\tfrac{1}{2}\,\|\dot{\bm{q}}\|_{\bm{G}_{\mathrm{ke}}(\bm{q})}^{2}.
\end{equation}
Under the unit–speed condition \(\|\dot{\bm{q}}\|_{\bm{G}_{\mathrm{ke}}}= 1\), geodesics on this Riemannian manifold coincide with trajectories that minimize the kinetic–energy functional.

\subsubsection{Jacobi metric.}
\label{sec:jacobi}
The metric defined by \(\bm{G}_{\mathrm{ke}}(\bm{q})=\bm{M}(\bm{q})\) captures only the kinetic part of the dynamics and neglects the potential energy \(P(\bm{q})\).
For a conservative system with fixed total energy \(H\), the Jacobi metric is defined by \citep{casetti2000geometric}:
\begin{equation} 
\bm{G}_{\mathrm{Jac}}(\bm{q}) = 2\bigl(H - P(\bm{q})\bigr)\, \bm{M}(\bm{q}), \qquad H > P(\bm{q}).
\end{equation}
The Hamiltonian satisfies \(H=T+P\) and remains constant along conservative motion~\citep{lutter2023combining}. 
The Jacobi metric is a conformal transformation of the kinetic–energy metric and is therefore smooth, symmetric, and positive-definite on the configuration space manifold.
By the Maupertuis–Jacobi principle, trajectories with fixed energy \(H\) correspond to geodesics on the Riemannian manifold endowed with this metric \citep{albu2022can}.

{\color{black}
\subsubsection{Pullback metric.}
The pullback operation induces a configuration–space metric from a task–space metric~\citep{ratliff2018riemannian}.
Let \(\varphi:\mathcal{Q}\to\mathcal{X}\) be a smooth task map with Jacobian \(\bm{J}(\bm{q})\), and let \(\bm{G}_{\mathcal{X}}\) be a smooth positive–definite matrix on \(\mathcal{X}\).
We define
\begin{equation} 
\bm{G}_{\mathrm{pb}}(\bm{q}) \coloneqq \bm{J}(\bm{q})^{\top}\, \bm{G}_{\mathcal{X}}\, \bm{J}(\bm{q}).
\end{equation}
as a smooth positive semidefinite matrix. One may add a small joint–space regularization,
\begin{equation} 
\tilde{\bm{G}}_{\mathrm{pb}}(\bm{q}) \coloneqq \bm{G}_{\mathrm{pb}}(\bm{q}) + \lambda \bm{I} \quad \text{with } \lambda>0,
\end{equation}
which yields a strictly SPD matrix while preserving the task–induced geometry. Common choices for \(\bm{G}_{\mathcal{X}}\) include \emph{stiffness} metrics with \(\bm{G}_{\mathcal{X}}=\bm{K}_x\), where \(\bm{K}_x\succ 0\) may be diagonal for per–axis stiffness or a full SPD matrix capturing cross–axis coupling \citep{Hogan1985ImpedanceI}; and \emph{manipulability} metrics, either kinematic with \(\bm{G}_{\mathcal{X}}=\bm{J}(\bm{q})\bm{J}(\bm{q})^{\top}\) ~\citep{Yoshikawa1985Manipulability} or more general with \(\bm{G}_{\mathcal{X}}=\bm{J}(\bm{q})\,\bm{\Lambda}(\bm{q})^{-1}\,\bm{J}(\bm{q})^{\top}\), where \(\bm{\Lambda}(\bm{q})\succ 0\) is a joint–space weighting, for example \(\bm{\Lambda}(\bm{q})=\bm{M}(\bm{q})\) to account for inertia in dynamic manipulability~\citep{lachner2020influence}, though other choices are possible. Additionally, setting \(\bm{G}_{\mathcal{X}}=\bm{I}\) yields the Euclidean metric in task space; its pullback \(\bm{J}(\bm{q})^{\top}\bm{J}(\bm{q})\) measures how joint–space motions map to task–space displacements. Geodesics under this metric correspond to joint–space paths that minimize task–space path length.
}

\subsubsection{Other task-specific metrics.}
Apart from the metrics above that reflect the robot’s inherent geometry, one can also design problem-specific metrics to encode task objectives directly. These task-specific metrics reshape the manifold so that geodesics realize desired behaviors. We illustrate this idea with two examples:

\paragraph{Obstacle avoidance.}
Obstacle avoidance typically relies on a collision checker or a distance to the nearest obstacle. To obtain adaptive behavior that is fast when far from obstacles and slow near boundaries, we introduce a smoothly varying scale \(\phi(\bm{x})\) and a distance-aware Riemannian metric tensor that penalizes motion along the normal direction to the obstacle ~\citep{klein2023design}:
\begin{equation}
\begin{aligned}
\phi(\bm{x}) \;&=\; f\!\left(\alpha\bigl(\tau- d(\bm{x})\bigr)\right)\,(1-\phi_{\min}) + \phi_{\min}, \\
\bm{G}_{\text{obs}}(\bm{x}) \;&=\; \phi(\bm{x}) \, \bm{r}_1(\bm{x}) \bm{r}_1(\bm{x})^\top \;+\; \bm{r}_2(\bm{x}) \bm{r}_2(\bm{x})^\top,
\end{aligned}
\end{equation}
where \(f\) is a smooth monotone function (e.g., the sigmoid \(f(z)=1/(1+e^{-z})\)), \(d(\bm{x})\) is the signed distance function, \(\alpha>0\) controls the transition sharpness, \(\tau\) sets a safety margin, and \(\phi_{\min}\ll 1\) enforces a minimum scale near obstacles. The vectors \(\{\bm{r}_1(\bm{x}),\bm{r}_2(\bm{x})\}\) form a local orthonormal frame with \(\bm{r}_1(\bm{x})\) aligned with \(\nabla d(\bm{x})\). Here, we use $\bm{x}$ to denote a task–space point, however, the formulation extends to configuration space $\bm{q}$ by employing distance fields for articulated robots \citep{li2024representing,Li24CDF}. With \(\phi\) increasing as proximity grows, the metric amplifies the cost of motion toward obstacles while leaving tangential motion comparatively less penalized. From a geometric perspective, this metric warps the local geometry near obstacles, causing geodesics to bend around them while maintaining clearance. The induced speed scaling further slows motion in close proximity, enhancing safety.

\paragraph{Stability-aware passive motion.}
In motion planning it is often desirable to bias trajectories toward passively stable, low–potential–energy regions~\citep{ortega2002interconnection}. A simple and effective construction is to scale the inertia by a normalized potential, yielding a position-dependent Riemannian metric tensor
\begin{equation}
\label{eq:passive_energy}
\begin{aligned}
\bm{G}_{\text{sta}}(\bm{q}) \;&=\; \bm{M}(\bm{q})\,\bigl[1 + \epsilon\,\tilde P(\bm{q})\bigr],\\
\tilde P(\bm{q}) \;&=\; \frac{P(\bm{q})-P_{\min}}{P_{\max}-P_{\min}} \in [0,1],\ \ \epsilon>0,
\end{aligned}
\end{equation}
where $\bm{M}(\bm{q})$ is the physical inertia matrix, $P(\bm{q})$ the potential energy, and $\epsilon>0$ controls the strength of the stability bias. $P_{\max}$ and $P_{\min}$ are the maximum and minimum potential energy, respectively. Larger \(P(\bm{q})\) produces a larger effective metric and hence longer paths, so geodesics on this manifold naturally avoid unstable regions while preserving the coupling encoded by \(\bm{M}(\bm{q})\). This metric preserves natural dynamics via uniform inertia modulation, regulates speed by slowing motion in high–potential zones, and guides the system toward low–energy basins.

Embedding stability in the metric makes robustness intrinsic: the eikonal is solved on a geometry already biased toward low–energy, stable states. The resulting distance field and geodesic flow integrate naturally with learning, planning, and control approaches to incorporate additional task–specific objectives and constraints.

\section{Related Work}\label{sec:rw}

\subsection{Distance Fields in Robotics}
Distance fields are fundamental representations in robotics, due to their capacity to implicitly encode spatial information while offering continuous, differentiable representations and efficient computational properties. This versatility has led to extensive exploration of signed distance fields (SDFs) for representing scenes and objects~\citep{millane2024nvblox,Maric24}, with demonstrated applications in collision detection~\citep{macklin2020local}, grasp synthesis~\citep{liu2021synthesizing}, motion generation~\citep{ratliff2009chomp}, and manipulation planning~\citep{yang2024contactsdf}. Moreover, distance fields are increasingly utilized as latent geometric features for downstream tasks such as dynamics models learning~\citep{driess2022learning}, grasp pose estimation~\citep{breyer2021volumetric,weng2022neural}, and motion policy generation~\citep{fishman2023motion}. Recent advances have introduced distance fields encoded with joint angles~\citep{liu2022regularized,koptev2022neural,li2024representing}, enabling efficient distance queries between arbitrary points and the surfaces of articulated robots. Building on this foundation, our previous work~\citep{Li24CDF} extended the concept of distance fields to the configuration space, wherein the representation measures the minimal joint motion required for a robot to reach specified points. 
The representation of articulated robots using distance fields can be interpreted as an implicit forward/inverse kinematics model, facilitating the utilization of distance and gradient information directly in joint space. By inherently capturing joint positions and velocities, this approach opens up new possibilities for advancing applications in reactive motion planning and control~\citep{koptev2024reactive}.
                         
\subsection{Solving the Eikonal Equation for Distance Fields}

\textcolor{black}{Early work computed travel-time (distance) fields by integrating geodesic ODEs as initial/boundary-value problems using shooting or ray tracing~\citep{julian1977three}. While effective locally, ray-based methods are sensitive to initialization and step size. They provide limited coverage away from sampled rays and can suffer from efficiency and convergence issues on manifolds with complex geometry or strong anisotropy~\citep{rawlinson2005fast}.} PDE-based solvers take a different route: they solve the eikonal equation, whose viscosity solution is the shortest-path distance from prescribed sources and whose characteristics encode front propagation. Such solvers are widely used in seismic tomography~\citep{lin2009eikonal}, rendering~\citep{ihrke2007eikonal}, image segmentation~\citep{alvino2007efficient}, and collision avoidance~\citep{garrido2013application}. The Fast Marching Method (FMM)~\citep{sethian1996fast} computes eikonal solutions on discretized grids with near-linear complexity, though grid discretizations face memory and resolution limits in high-dimensional spaces. Unlike ODE ray tracing, which returns distances only along sampled geodesics, PDE formulations yield a \emph{global} distance field whose gradient flow induces geodesics. 

Recent advances in physics-informed neural networks (PINNs) enable grid-free eikonal solvers by representing the solution as a differentiable neural field and training it via the PDE residual and boundary conditions, with gradients obtained through backpropagation~\citep{raissi2019physics}. In this view, the network provides a continuous scalar field, while the eikonal constraint in the loss drives it to behave as a distance-to-go function even without supervised distance labels, making its gradients directly usable for geodesic backtracking. This approach has been applied to isotropic eikonals (e.g., EikoNet)~\citep{smith2020eikonet}. In robotics, the eikonal constraint also appears when training implicit signed distance fields for shape representation~\citep{gropp2020implicit,xie2022neural}. For motion planning, NTFields solves neural eikonal equations under collision-avoidance constraints and demonstrates fast, scalable generation in high-dimensional spaces~\citep{ni2022ntfields,ni2023progressive}. Beyond Euclidean settings, the eikonal constraint has been extended to manifolds~\citep{ni2024physics}. The Riemannian Fast Marching method adapts fast marching to anisotropic (Riemannian) metrics~\citep{mirebeau2017anisotropic}, and the heat method computes geodesic distances via short-time heat flow~\citep{crane2013geodesics}. More recently, neural eikonal solvers have been studied directly on manifolds to compute geodesics and distance fields, further broadening applicability~\citep{kelshaw2024computing}.
\\
\subsection{Motion Planning on Manifolds}
Recent advances in motion planning leveraged Riemannian manifolds to tackle complex challenges. Obstacles are often treated as features that reshape the geometry of the space, allowing geodesics to naturally navigate around them and achieve collision-free motion~\citep{ratliff2015understanding,laux2021robot}. Building on this concept, Riemannian motion policies can be used in joint space using a pullback metric~\citep{ratliff2018riemannian}. This framework was later extended to Geometric Fabrics~\citep{van2022geometric} by incorporating principles of classical mechanics for more adaptable motion planning. Beyond static obstacle avoidance, dynamic-aware motions have been explored through kinetic energy-based Riemannian metrics~\citep{jaquier2022riemannian,klein2023design}, with further extensions to the Jacobi metric that account for both kinetic and potential energy, enabling energy-conserving paths~\citep{albu2022can}.
Additionally, Riemannian metrics have been applied to human motion modeling, where geodesics represent minimum-effort paths in configuration space~\citep{neilson2015riemannian}. These ideas have inspired methods to transfer human arm motions to robots, facilitating more natural and human-like behavior~\citep{klein2022riemannian}. Unlike these approaches that focus on local policies or optimizing for the shortest geodesics, our method emphasizes constructing a comprehensive distance field over the entire configuration space, allowing for more flexible and efficient motion planning.

\section{Riemannian Eikonal Solver}\label{sec:res}

In this section, we present the methodology for solving the Riemannian eikonal equation to compute geodesic distances and flows. {\color{black} For clarity, we first describe a numerical PDE solver based on Riemannian Fast Marching (RFM), and then introduce our \emph{Neural Riemannian Eikonal Solver (NES)}, which parameterizes the PDE and learns the distance field and geodesic flows directly. We also present two NES variants that condition NES on boundary points and metrics. Because our tasks primarily involve solving the equation on the configuration-space manifold, we denote configurations by $\bm{q}$ unless we explicitly refer to task-space points, which we denote by $\bm{x}$. 
}

\subsection{Riemannian Fast Marching}
\label{sec:RFM}
Riemannian Fast Marching (RFM)~\citep{mirebeau2017anisotropic} is an extension of classical Fast Marching Methods for solving the eikonal equation on anisotropic, inhomogeneous manifolds endowed with a Riemannian metric. {\color{black}As described in Section~\ref{subsecsec:REE}, we seek the single-point solution \(U(\bm{q})\) such that \(\|\nabla U(\bm{q})\|_{\bm{G}^{-1}(\bm{q})}=1\), and we recover geodesics by backtracking the unit-\(\bm{G}\)-speed flow~\eqref{eq:riem-direction-V}}.
Like FMM, RFM tackles single-source problems by discretizing the manifold into a grid and applying an upwind finite-difference scheme. The wavefront starts from a fixed source \(\bm{q}_s\) and marches outward. RFM updates the travel time at each neighbor \(\bm{q}_n\) of a current front point \(\bm{q}_e\) via
\begin{equation}
\label{eq: rfm_update}
    U(\bm{q}_n) \;=\; \underset{\bm{q}_e \in \bm{Q}_e}{\min}\;
    \Bigl( U(\bm{q}_e) + \|\bm{q}_n - \bm{q}_e \|_{\,\bm{G}(\bm{q}_e)} \Bigr),
\end{equation}
where \(\bm{Q}_e\) is the evolving wavefront, \(\bm{q}_n \in \bm{Q}_n = N(\bm{q}_e)\) is a neighbor, and
\(\|\bm{q}_n - \bm{q}_e \|_{\,\bm{G}(\bm{q}_e)}\)
is the local metric-induced norm. A detailed algorithm for computing geodesic distances on configuration-space manifolds is given in Algorithm~\ref{alg:RFM}. After obtaining \(U(\bm{q})\), geodesics from an arbitrary point \(\bm{q}\) to the source \(\bm{q}_s\) are retrieved efficiently by backtracking using \eqref{eq:backtrack-V} and the geodesic field \eqref{eq:riem-direction-V}.

\begin{algorithm}
\caption{Riemannian Fast Marching}
\label{alg:RFM}
\KwIn{Discretized grid \(\mathcal{M}\); matrix \(\bm{G}(\bm{q})\); source \(\bm{q}_s\)}
\KwOut{\(U(\bm{q})\): geodesic distance from \(\bm{q}_s\) to \(\bm{q}\in\mathcal{M}\)}

\textbf{Initialization:}\\
\textit{set} \(U(\bm{q}_s)=0\); \ \(\,U(\bm{q})=\infty\) for all \(\bm{q}\neq \bm{q}_s\).\\
\textit{set} \(\bm{Q}_e=\{\bm{q}_s\}\) \CommentSty{wavefront}\\

\textbf{Propagation Step:}\\
\While{there exists \(\bm{q}\in\mathcal{M}\) with \(U(\bm{q})=\infty\)}{
  \For{each \(\bm{q}_e\in\bm{Q}_e\)}{
    find neighbors \(\bm{Q}_n=N(\bm{q}_e)\) \\
    \For{each \(\bm{q}_n\in\bm{Q}_n\)}{
      \If{\(U(\bm{q}_n)=\infty\)}{
         update \(U(\bm{q}_n)\) via \eqref{eq: rfm_update} \\
         \(\bm{Q}_e \leftarrow \bm{Q}_e \cup \{\bm{q}_n\}\)
      }
    }
    \(\bm{Q}_e \leftarrow \bm{Q}_e \setminus \{\bm{q}_e\}\)
  }
}
\textbf{Termination:} for all \(\bm{q}\in\mathcal{M}\), \(U(\bm{q})\neq \infty\).
\end{algorithm}

As a finite-difference, single-pass method, RFM is computationally efficient and numerically accurate on grids. However, the dependence on an explicit discretization limits scalability for high-dimensional manipulators. We therefore propose a neural parameterization in the next section to address this limitation.

\subsection{Neural Riemannian Eikonal Solver (NES)}
\label{sec:nes}
The Neural Riemannian Eikonal Solver (NES) is inspired by recent developments in solving PDEs through deep neural networks.
Different from classical grid-based approaches, it operates in a continuous space without explicit discretization. Gradients are calculated through network backpropagation by automatic differentiation, allowing for high-dimensional manifolds with continuously varying metrics. Instead of seeking \textit{single-point} solutions, NES allows for global geodesic distances for \textit{source-goal} pairs, where $U(\bm{q}_s,\bm{q}_e)$ is a function of both source $\bm{q}_s$ and goal $\bm{q}_e$ points, thanks to the flexible structure provided by neural networks.
Therefore, the Riemannian eikonal equation to be solved is written as
\begin{equation}
    \label{eq: eikonal_whole}
    \|\nabla_{\bm{q}_e} U(\bm{q}_s,\bm{q}_e) \|_{\bm{G}^{-1}(\bm{q}_e)}=1, \quad \text{s.t.} \quad U(\bm{q}_s,\bm{q}_s) = 0.
\end{equation}

This equation lets us train a neural field $U_\theta(\bm{q}_s,\bm{q}_e)$ directly from the eikonal constraint: we sample source--goal pairs and optimize a PDE-residual loss that acts on $\nabla_{\bm{q}_e} U_\theta$. To obtain a consistent and continuous distance field, we further impose the following physical constraints:

\textbf{Symmetry}. The geodesic distance  
from a source point $\bm{q}_s$ to the destination point $\bm{q}_e$, and in the other direction are identical, following the symmetry property that $U(\bm{q}_s,\bm{q}_e) = U(\bm{q}_e,\bm{q}_s)$. It also applies to their partial derivatives:
$\nabla_{\bm{q}_e} U(\bm{q}_s,\bm{q}_e) = \nabla_{\bm{q}_e} U(\bm{q}_e,\bm{q}_s)$. To impose this constraint, we define a symmetric function $u_{\bm{\theta}}^\text{sym}$~\citep{ni2022ntfields}
\begin{equation}
    \label{eq: sym}
    u_{\bm{\theta}}^\text{sym}(\bm{q}_s, \bm{q}_e     
            ) = \frac{u_{\bm{\theta}}(\bm{q}_s, \bm{q}_e)+u_{\bm{\theta}}(\bm{q}_e, \bm{q}_s)}{2},
\end{equation}
where $u_{\bm{\theta}}$ is the output of neural network parameterized by $\bm{\theta}$. This equation ensures the symmetry of the network output with respect to permuted source-to-goal pairs.

\textbf{Non-negativity and non-singularity.} The geodesic distance is strictly positive between any two distinct points, and zero when the points coincide. It can be simply achieved by adding a non-negative activation function $\sigma$. However, the geodesic distance should approach zero for points close to one another, which might cause singularity issues, leading to numerical errors in distances and gradients for points close to the source point. To overcome this problem, we follow the approach in~\citep{kelshaw2024computing,smith2020eikonet} that factorizes the distance function as
\begin{equation}
    \label{eq: isfactor}
    U_{\bm{\theta}}(\bm{q}_s, \bm{q}_e) = \lVert \bm{q}_e - \bm{q}_s \rVert \sigma\left( u_{\bm{\theta}}^\text{sym}(\bm{q}_s, \bm{q}_e     
            )\right),
\end{equation} 
where $\sigma(\cdot)$ is a non-negative activation function and  
$ \lVert \bm{q}_e - \bm{q}_s \rVert$ is the Euclidean term between two joint configurations for non-singularity.
This equation guarantees the non-negativity of geodesic distance and implicitly constrains the gradient pointing to the destination.

\textbf{Loss Functions.} The parameterization involves a multi-layer perceptron (MLP) neural network, with a batch of concatenated joint configuration pairs $\bm{q}_s$ and $\bm{q}_e$ as input and outputs the predicted geodesic distance $U_{\bm{\theta}}(\bm{q}_s, \bm{q}_e) $. Ground truth geodesic distances are unknown, and the neural network is supervised through the physical law defined by the Riemannian eikonal equation~\eqref{eq: eikonal_whole}. Therefore, for each source-to-goal point pair, we minimize the  loss function
\begin{equation}
    \label{eq: eik_loss}
    \mathbb{L}_\text{eik}(\bm{q}_s,\bm{q}_e) = \left(\lVert\nabla_{\bm{q}_e}U_{\bm{\theta}}(\bm{q}_s, \bm{q}_e)\rVert_{\bm{G}^{-1}(\bm{q}_e)}  - 1\right)^2
\end{equation}
to construct the geodesic distance field. The partial derivatives $\nabla_{\bm{q}_e}U_{\bm{\theta}}(\bm{q}_s, \bm{q}_e)$ are computed through automatic differentiation. In addition, to produce a smooth geodesic distance field, we add a regularization term based on the Laplace-Beltrami operator, which defines the divergence of the vector field:
\begin{equation}
    \label{eq: div_loss}
    \mathbb{L}_\text{div}(\bm{q}_s, \bm{q}_e) = \left(G^{ij} \left(\frac{\partial^2 U(\bm{q}_s, \bm{q}_e)}{\partial q_{e, i}\partial q_{e, j}} - {}\Gamma^k_{ij} \frac{\partial U(\bm{q}_s, \bm{q}_e)}{\partial q_{e, k}}\right)\right)^2.
\end{equation}
 The Laplace–Beltrami term discourages spurious high curvature and reduces gradient oscillations, yielding more stable geodesic backtracking. For details on the derivation of the divergence term, please refer to Appendix \ref{sec:laplace-beltrami operator}.

The total loss is
\begin{equation}
    \label{eq: total_loss}
    \mathbb{L}_{\mathrm{total}} 
    = \frac{1}{N}\sum_{n=1}^{N}\big(\mathbb{L}_{\mathrm{eik}} + \lambda\,\mathbb{L}_{\mathrm{div}}\big),
\end{equation}
where \(N\) is the batch size and \(\lambda\) weights the divergence regularization term.

{\color{black}
\subsection{Conditioned NES}
\label{sec:nes-variants}

The key advantage of NES over traditional numerical PDE solvers and optimal control approaches is its grid-free and end-to-end differentiability, which provides great flexibility for the neural network parameterization. Building on this property, we introduce conditioned NES (C-NES), which contains two variants, either conditioned on boundaries or conditioned on Riemannian metrics.
}
\subsubsection{Conditioned on boundaries.}
In the standard NES formulation, both source and goal lie in the same space (e.g., the robot’s configuration space), and the solution traces a minimal–distance curve on that manifold. Many manipulation tasks, however, involve coupling the task space and joint space. We address this by conditioning the boundary on a task–space specification while still propagating the wavefront in joint space—a hybrid process that implicitly handles inverse kinematics.

Here, we condition the boundary on a task–space source point \(\bm{x}_s\). Conceptually, the boundary in configuration space is the set of configurations satisfying \(f(\bm{q})=\bm{x}_s\), where \(f\) is the forward kinematics. In practice, we do \emph{not} enumerate these IK solutions. Instead, we use \(f\) directly in the network to anchor the boundary implicitly. The neural network is parameterized as:
\begin{equation}
\label{eq:nn_ik}
U_{\bm{\theta}}(\bm{x}_s,\bm{q}_e)
=
\big\|f(\bm{q}_e)-\bm{x}_s\big\|\;\sigma\!\big(u_{\bm{\theta}}(\bm{x}_s,\bm{q}_e)\big),
\end{equation}
where \(\sigma(\cdot)\) enforces nonnegativity and the Euclidean factor \(\|f(\bm{q}_e)-\bm{x}_s\|\) drives \(U_{\bm{\theta}}\) to zero whenever the end–effector reaches the task–space source \(\bm{x}_s\). Because \(\bm{x}_s\) (task space) and \(\bm{q}_e\) (configuration space) live in different spaces, symmetry does not apply; we therefore use the raw output \(u_{\bm{\theta}}\) rather than a symmetrized variant. The corresponding Riemannian eikonal equation we aim to solve is
\begin{equation}
\label{eq:eikonal_ik}
\begin{aligned}
&\|\nabla_{\bm{q}_e} U(\bm{x}_s,\bm{q}_e)\|_{\bm{G}^{-1}(\bm{q}_e)} = 1,\\
&U(\bm{x}_s,\bm{q}) = 0 \quad \text{whenever } f(\bm{q})=\bm{x}_s .
\end{aligned}
\end{equation}

The wavefront originates on the boundary set \(\{\bm{q}: f(\bm{q})=\bm{x}_s\}\) in configuration space and propagates outward; geodesics are recovered by backtracking the flow from the query configuration \(\bm{q}_e\).
Under this formulation, \(U(\bm{x}_s,\bm{q}_e)\) equals the Riemannian distance (under \(\bm{G}\)) from \(\bm{q}_e\) to the IK set. Backtracking the geodesic flow from \(\bm{q}_e\) therefore terminates at an IK configuration that is \emph{geodesically closest} to \(\bm{q}_e\). Crucially, this selection is obtained \emph{without} computing or sampling all IK solutions—the PDE solution together with \eqref{eq:nn_ik} imposes the boundary/source implicitly and recovers the minimizing IK endpoint via geodesics.

The training is identical to standard NES \eqref{eq: eik_loss}–\eqref{eq: div_loss}, except the source is specified in task space. The losses are
\begin{equation}
\label{eq:ik_losses}
\resizebox{\linewidth}{!}{$
\begin{aligned}
\mathbb{L}_{\mathrm{eik}}(\bm{x}_s,\bm{q}_e)
&= \Big(\|\nabla_{\bm{q}_e} U_{\bm{\theta}}(\bm{x}_s,\bm{q}_e)\|_{\bm{G}^{-1}(\bm{q}_e)} - 1\Big)^2,\\
\mathbb{L}_{\mathrm{div}}(\bm{x}_s,\bm{q}_e)
&= \Big(
G^{ij}(\bm{q}_e)\big[
\frac{\partial^2 U(\bm{x}_s,\bm{q}_e)}{\partial q_{e,i}\partial q_{e,j}}
-\Gamma^{k}_{ij}(\bm{q}_e)\frac{\partial U(\bm{x}_s,\bm{q}_e)}{\partial q_{e,k}}
\big]\Big)^2.
\end{aligned}
$}
\end{equation}

{\color{black}
\subsubsection{Conditioned on metrics.}
Another variant of C–NES conditions the eikonal on a \emph{parameterized} Riemannian metric. Let the metric depend on a task/physics parameter \(\omega\), written \(\bm{G}(\bm{q};\omega)\), and let the corresponding distance be \(U(\bm{q}_s,\bm{q}_e \mid \omega)\). 
The neural parameterization remains as in \eqref{eq: sym}–\eqref{eq: isfactor}; we simply augment the network inputs with \(\omega\), i.e., \(u_{\bm{\theta}}=u_{\bm{\theta}}(\bm{q}_s,\bm{q}_e,\omega)\), while enforcing symmetry only over \((\bm{q}_s,\bm{q}_e)\). 
The parameter \(\omega\) acts through the metric, so the eikonal becomes
\begin{equation}
\label{eq:eikonal_whole_metric}
\Big\|\nabla_{\bm{q}_e} U(\bm{q}_s,\bm{q}_e \mid \omega)\Big\|_{\bm{G}^{-1}(\bm{q}_e;\omega)} = 1,
\qquad
U(\bm{q}_s,\bm{q}_s \mid \omega) = 0 .
\end{equation}
The training objectives are identical to \eqref{eq: eik_loss}–\eqref{eq: div_loss}, evaluated with the augmented input \(\omega\). Conditioning NES on the Riemannian metric allows a single network to represent a \emph{family} of distance fields and geodesic flows. At test time, we adapt to different metrics by adjusting the parameter \(\omega\), allowing us to interpolate smoothly across geometries by varying \(\omega\).

\subsection{Benefits of NES}
Building on the RFM approach introduced earlier for solving the Riemannian eikonal equation, we summarize the key advantages of NES over classical numerical PDE solvers for computing distance fields and geodesics:

\textbf{Grid-free and end-to-end differentiable.}
NES operates in continuous space without explicit meshes and is trained via automatic differentiation. The resulting distance fields and geodesic flows are end-to-end differentiable, enabling gradient-based optimization, policy learning, and closed-loop control within a unified framework.

\textbf{Scales to high dimensions.}
By avoiding discretization and leveraging neural parameterization, NES remains tractable on high-degree-of-freedom configuration manifolds with continuously varying metrics, where grid- or mesh-based methods become impractical due to resolution and memory constraints. The inference time is also fast once the model is trained. 

\textbf{Globality.}
Motivated by Fast Marching, NES enforces the eikonal PDE as a derivative constraint on \(U\) over the whole domain and approximates its viscosity solution, producing a single, globally consistent distance potential on the configuration manifold. In contrast, shooting/BVP methods require explicit geodesic solves and can become trapped in local minima.

\textbf{Flexibility.}
NES encodes a two-point value function \(U(\bm{q}_s,\bm{q}_e)\) rather than a single-source field, and it naturally supports conditioning on boundary specifications as well as on parameterized Riemannian metrics \(\bm{G}(\cdot;\omega)\). These changes are handled by inputs to a single model, whereas numerical PDE solvers typically require re-discretization or re-solving the problem to accommodate them.
}

\section{Examples}
\label{sec:examples}

\begin{table*}[t]
    \begin{center}
        \begin{tabular}{cccc}
            \centering
            \includegraphics[width =0.22\linewidth
            ]{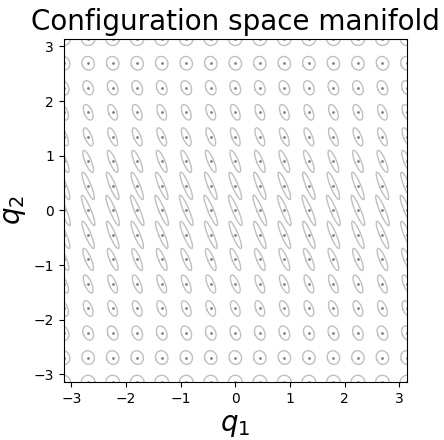}   &
            \includegraphics[width =0.205\linewidth]{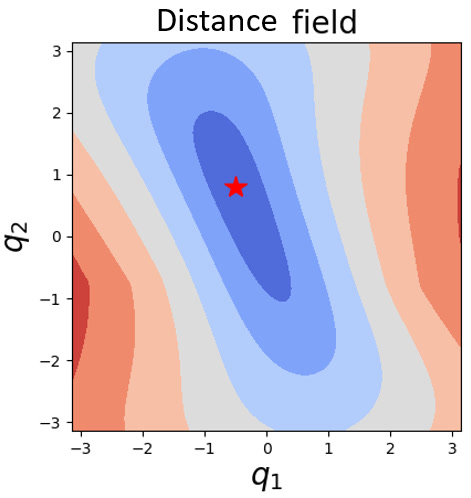}  &
            \includegraphics[width =0.205\linewidth]{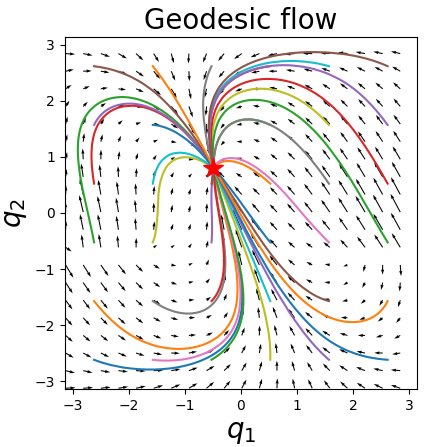}  &
            \includegraphics[width =0.21\linewidth]{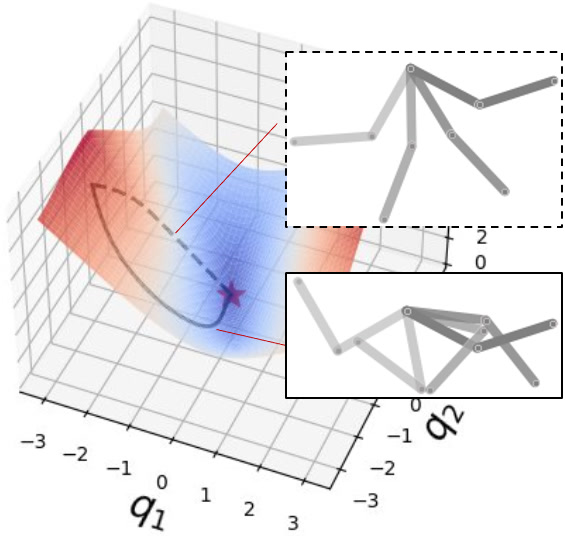}  \\
            (a) & (b) & (c) & (d)
        \end{tabular}
        \captionof{figure}{(a) Configuration space manifold endowed with a Riemannian metric using inertia as weighting matrix (visualized as isocontours of inverse matrices). The geodesics on this manifold correspond to minimal kinetic energy paths. By starting from a given point (red star), we can solve the eikonal equation on this manifold, accounting for the distance field (b) and gradient flow (c), which can then be used to backtrack geodesics in a very rapid manner (in milliseconds), see colored paths for examples of retrieved trajectories. Here, the source point is fixed for visualization. By using the proposed Neural Riemannian eikonal Solver (NES), these points are given as inputs, meaning that geodesics from any starting point to any final point are considered altogether. 
        (d) Geodesic path (solid line) and Euclidean path (dashed line) on this manifold with corresponding robot motions. }
        \label{fig:RFM}
    \end{center}
\end{table*}

\begin{table*}[t]
    \begin{center}
        \begin{tabular}{cccc}
            \centering
            \includegraphics[width =0.22\linewidth]{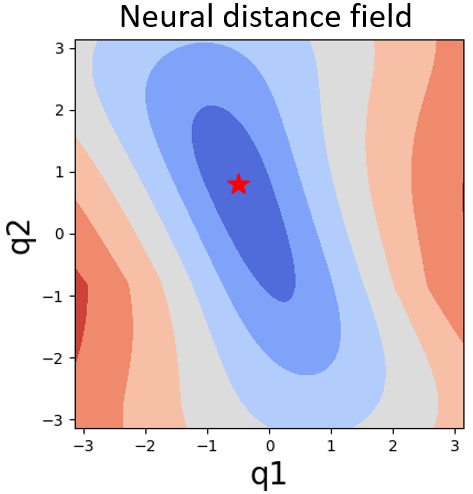}  &
            \includegraphics[width =0.22\linewidth]{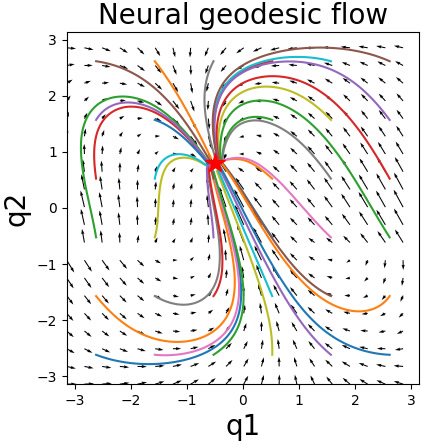}  &
            \includegraphics[width =0.22\linewidth]{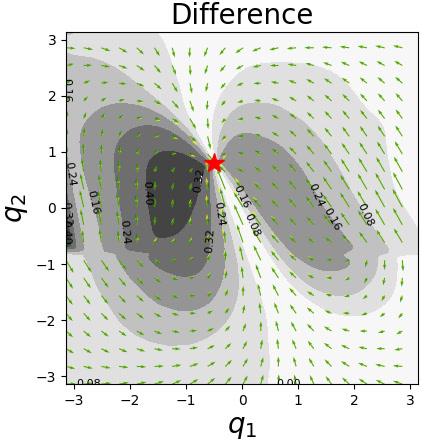}  &
            \includegraphics[width =0.22\linewidth]{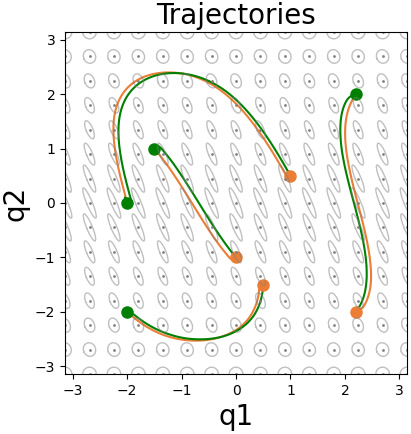}  \\
            (a) & (b) & (c) & (d)
        \end{tabular}
        \captionof{figure}{Solutions of Neural Riemannian Eikonal Solver (NES). (a) and (b) show the distance field and geodesic flow with the same parameters as Riemannian Fast Marching (RFM). (c) compares the difference with Figure~\ref{fig:RFM} (b) (c), where geodesic flows produced by NES and RFM are shown in yellow and green, respectively. These three figures demonstrate that the neural network parameterization can solve the Riemannian eikonal equation, yielding results similar to those of RFM. (d) shows trajectories from source (green) to goal (orange) points and vice versa, highlighting the generalizability and symmetry of NES for arbitrary joint angle configuration pairs.}
        \label{fig:nes_2d}
    \end{center}
\end{table*}
{\color{black}
In this section we illustrate how the proposed methods produce distance fields and geodesic flows that are directly useful for robot problems. We present two groups of examples. First, we validate NES and its conditioned variant (C–NES) under an energy-related metric that reflects the robot's dynamics. Second, we show how different Riemannian metrics shape distance fields and motions for various robot applications. We provide quantitative visualizations to clarify the behavior of the learned fields and flows.
}
\subsection{NES for Minimal Energy Geodesics}
\label{sec:nes-meg}
We consider a 2D planar manipulator with two links, each having a length of $l_1 = l_2 = 2$ and masses $m_1 =m_2=1$ concentrated at each articulation. The joint angle ranges from $-\pi$ to $\pi$ for both links. Consequently, the corresponding inertial mass matrix $\bm{M}(\bm{q})$ is expressed as:
\begin{equation}
\resizebox{\linewidth}{!}{%
$\bm{M}(\bm{q})\!=\! 
\begin{bmatrix}
(m_1 + m_2) l_1^2 + m_2 l_2^2 + 2 m_2 l_1 l_2 \cos(q_2) & m_2 l_2^2 + m_2 l_1 l_2 \cos(q_2) \\
m_2 l_2^2 + m_2 l_1 l_2 \cos(q_2) & m_2 l_2^2
\end{bmatrix},$%
}
\end{equation}
where \(q_1\) and \(q_2\) are the joint angles. The configuration–space manifold is endowed with the kinetic–energy Riemannian metric \(\bm{G}_{\text{ke}}(\bm{q})=\bm{M}(\bm{q})\). We visualize the manifold and metric as ellipsoids in Figure~\ref{fig:RFM}(a). Given a fixed source, we solve the Riemannian eikonal equation using the RFM approach in Algorithm~\ref{alg:RFM}, yielding the distance field (b) and geodesic flows (c). Panel (d) shows a 3D visualization of the resulting geodesics, together with a Euclidean reference path and the corresponding robot motions. Geodesics are curved paths in configuration space that minimize kinetic energy.

\paragraph{NES Solution}. Using the approach in Section~\ref{sec:nes}, we can also obtain solutions through the proposed neural Riemannian eikonal solver, visualized in Figure~\ref{fig:nes_2d}. Panels (a) and (b) show the learned distance field and associated geodesic flows. The differences with RFM in (c) are small, indicating that the network accurately approximates the Riemannian eikonal and yields results comparable to the RFM baseline. Panel (d) shows geodesics between arbitrary configuration pairs, including symmetric round trips, underscoring the model’s ability to generalize to arbitrary source–goal pairs. In contrast, RFM is single–source and must be recomputed for each new query.

\paragraph{C-NES on the Task-space Boundary.} Using the same kinetic–energy metric, we further show NES conditioned on a boundary induced by a task–space source, as visualized in Figure~\ref{fig:nes_ik}. The end–effector target is \(\bm{x}_s=(2.0,2.0)\), shown in Figure~\ref{fig:nes_ik}(a). This point admits two IK solutions, e.g., \((q_1,q_2)=(0,1.57)\) and \((1.57,-1.57)\), but these are not provided to C–NES a priori. C–NES takes \((\bm{x}_s,\bm{q}_e)\) as input; the resulting distance map is shown in Figure~\ref{fig:nes_ik}(a), and the corresponding geodesic flows in Figure~\ref{fig:nes_ik}(b). Backtracking the flow from \(\bm{q}_e\) discovers the IK solutions (red stars) and, upon integration, yields the geodesic that minimizes kinetic–energy distance. Figure~\ref{fig:nes_ik}(c) displays four motions that all reach the same end–effector position but start from different configurations. Due to redundancy, different geodesic flows terminate at different IKs. The orange trajectory follows the energetically optimal branch, leading to IK solutions distinguishing from the others.

\begin{table*}[t]
    \begin{center}
        \begin{tabular}{ccc}
            \centering
            \includegraphics[width =0.23\linewidth
            ]{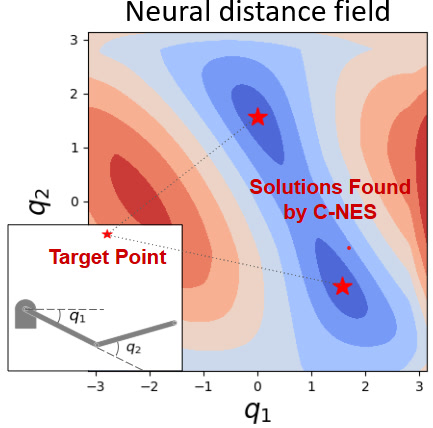}   &
            \includegraphics[width =0.22\linewidth]{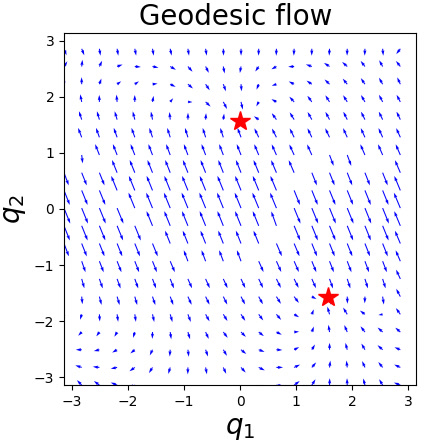}  &
            \includegraphics[width =0.49\linewidth]{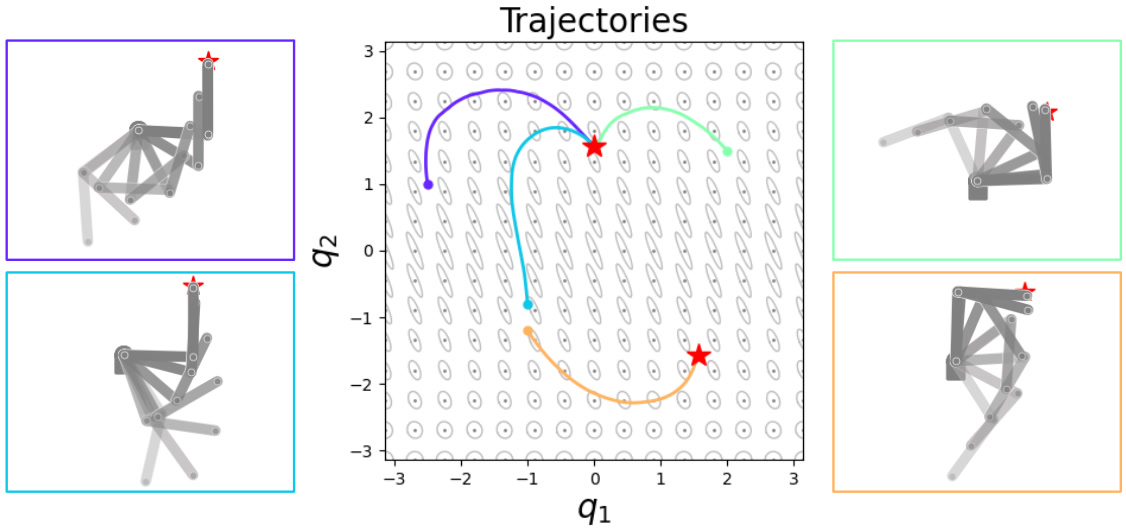}  \\
            (a) & (b) & (c) 
        \end{tabular}
        \captionof{figure}{Given a kinetic-energy metric, (a) and (b) show the distance field and geodesic flow for the target position $(2.0,2.0)$ in task space by using C-NES. Here, we do not specify the target joint angles (red stars). These joint angle targets are instead learned implicitly by the neural network. (c) shows four robot motions in task and configuration spaces (with different colors). We can observe that the motion solution for the task in orange color differs from the other three, which are automatically computed in accordance with the distance field and geodesic flow.}
        \label{fig:nes_ik}
    \end{center}
\end{table*}

\begin{figure}[t]
    \centering
    \includegraphics[width=0.95\linewidth]{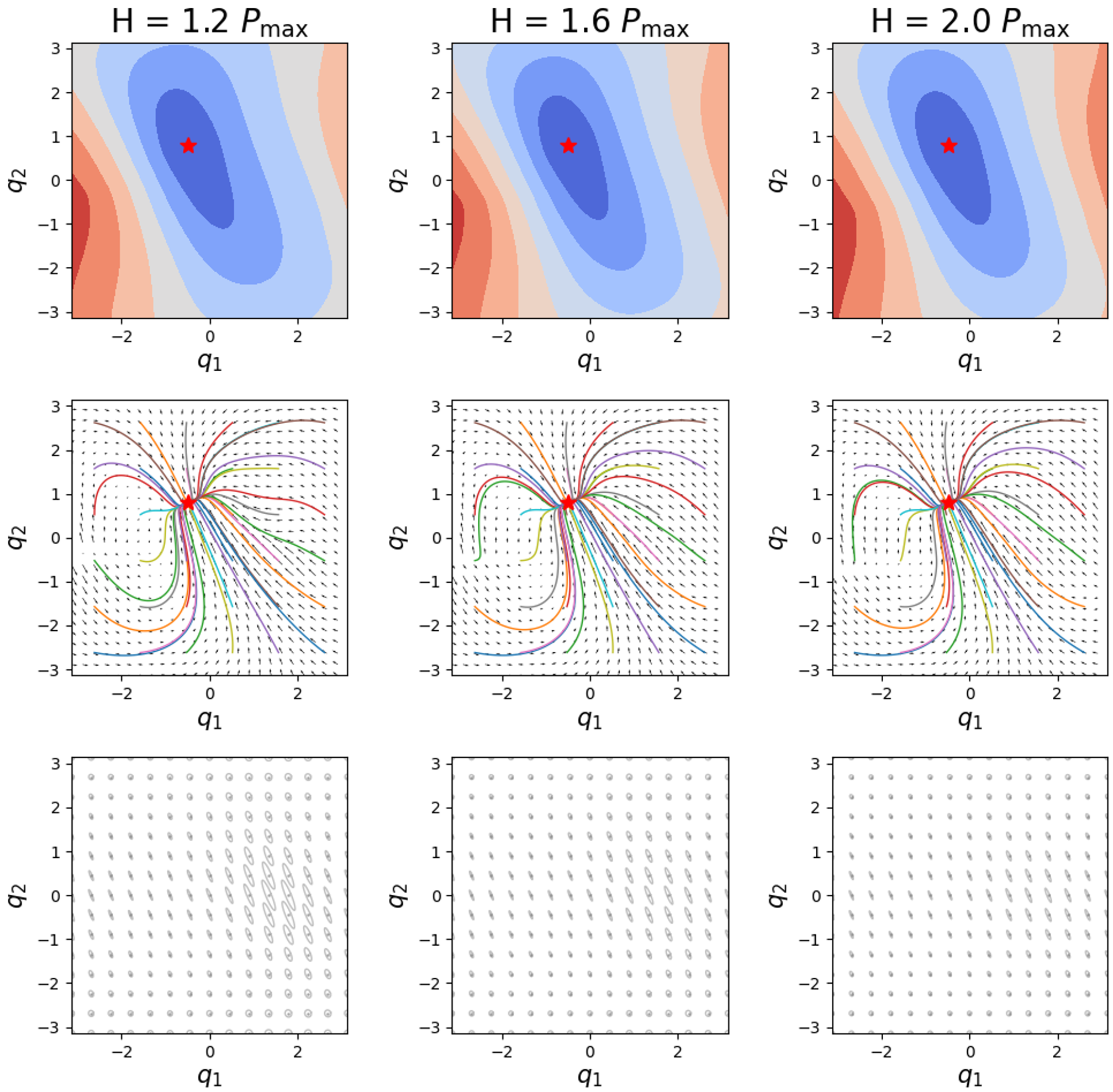}
    \caption{\textcolor{black}{C–NES conditioned on the Jacobi metric parameterized by the total energy \(H\). \textit{Top}: distance fields. \textit{Middle}: geodesics. \textit{Bottom}: configuration-space manifolds.}}
    \label{fig:jacobi_h}
\end{figure}

{\color{black}
\paragraph{C–NES on the Riemannian Metric.}
A convenient instantiation of the metric-conditioning is to use the Jacobi metric derived from Hamiltonian dynamics that considers both kinetic and potential energy (Section~\ref{sec:jacobi}). Specifically, we set the parameter in \eqref{eq:eikonal_whole_metric} to the total energy, i.e., \(\omega = H\), and train our NES conditioned on this parameter. With this choice, C–NES represents distance fields and geodesic flows at fixed energy \(H\) by simply augmenting the network input with \(H\), while keeping the training objectives unchanged. Varying \(H\) smoothly deforms the geometry through the conformal factor \(2(H-P)\). As \(H\) increases, the conformal factor \(2\!\left(H-P(\bm{q})\right)\) grows and \(\bm{G}_{\text{Jac}}(\bm{q};H)\) more closely resembles a constant scaling of \(\bm{M}(\bm{q})\). Accordingly, the resulting geodesics become closer to those induced by the kinetic–energy metric \(\bm{G}_{\mathrm{ke}}\); in the idealized limit \(H \to \infty\), they coincide. For smaller \(H\), the available kinetic energy \(T=H-P(\bm{q})\) is reduced, so the induced geodesics correspond to lower–energy motions. However, if \(H\) approaches \(P(\bm{q})\) anywhere along a path, the conformal factor \(2(H-P)\) tends to zero and the Jacobi metric becomes nearly singular, potentially causing numerical ill-conditioning. 
Figure~\ref{fig:jacobi_h} visualizes the distance fields, geodesics, and configuration–space manifolds for total energies \(H \in \{1.2,\,1.6,\,2.0\}P_{\max}\), where \(P_{\max} \) represents the maximum potential energy.
}

\begin{figure}[t]
    \centering
    \includegraphics[width=\linewidth,trim=0pt 0pt 0pt 0pt,clip]{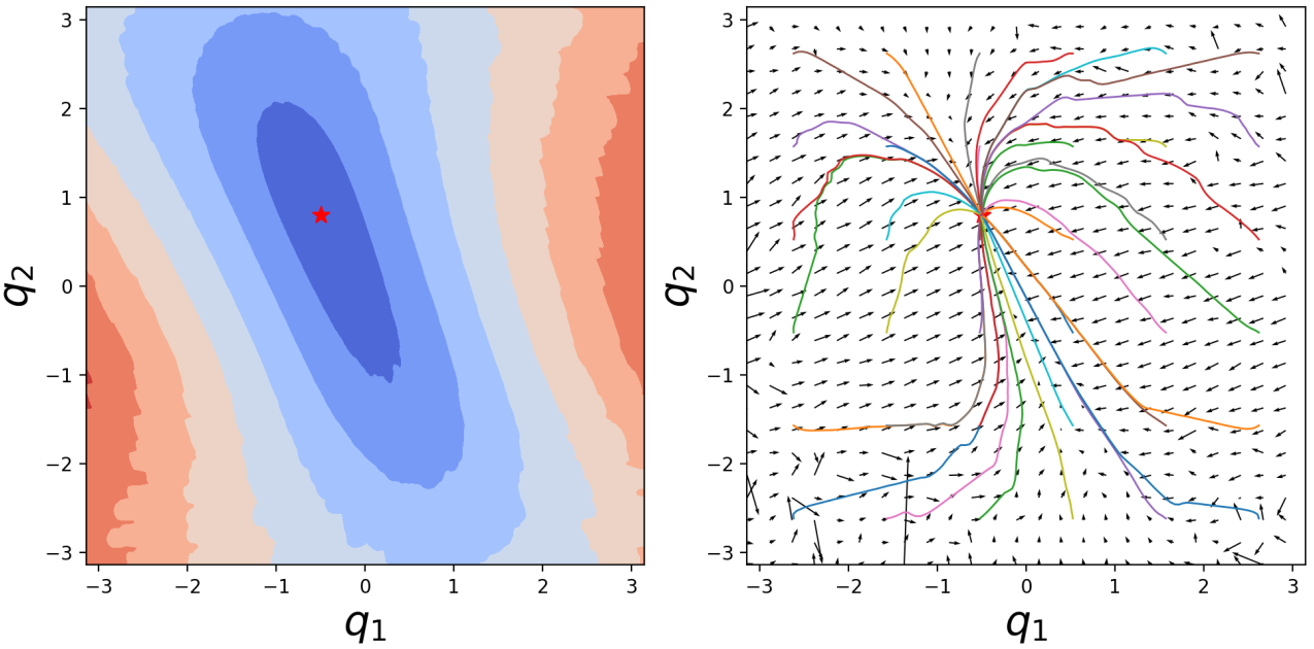}\\
    \includegraphics[width=\linewidth,trim=0pt 0pt 0pt 0pt,clip]{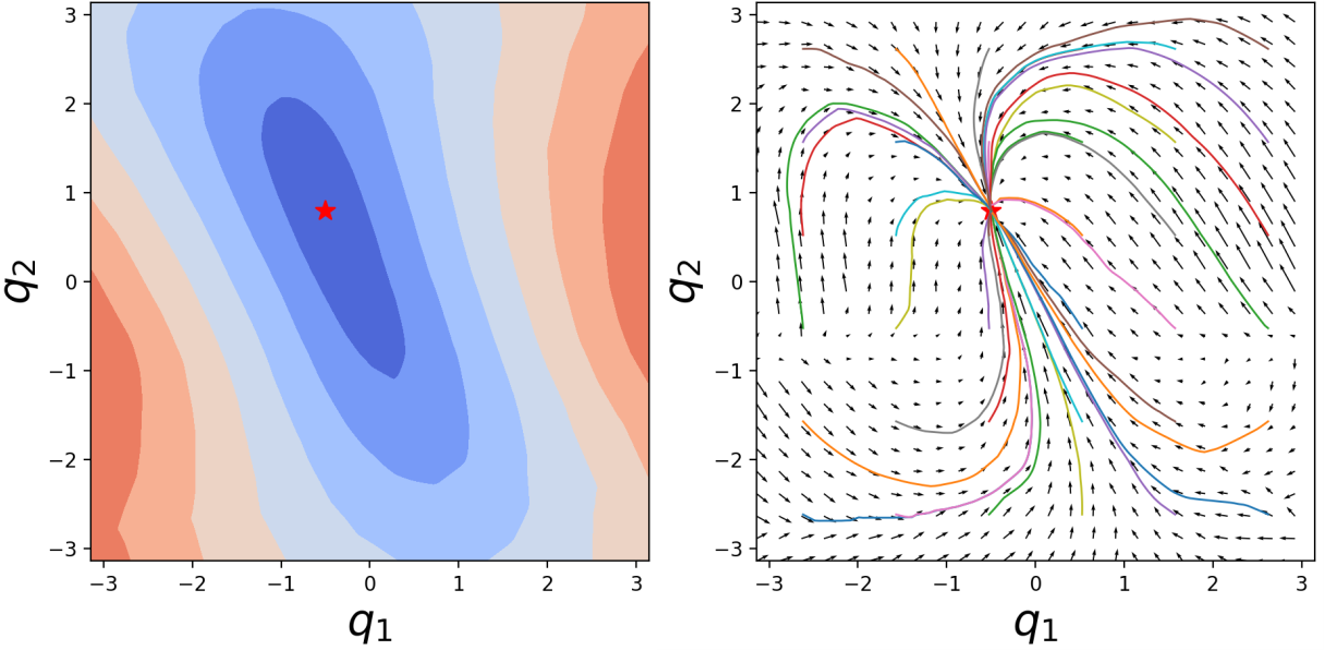}
    \caption{\textcolor{black}{Distance field (\textit{left}) and geodesics (\textit{right}). \textit{Top}: results obtained by the Geodesic ray tracing (GRT) baseline. \textit{Bottom}: supervised regression trained on GRT labels.}}
    \label{fig:compare_gofm_sl}
\end{figure}

{\color{black}
\subsection{Comparison of Distance Field-based Approaches}
We advocate solving the Riemannian eikonal equation to obtain a global, metrically consistent distance field from which geodesics are recovered by backtracking. To justify this choice, we compare NES against two baselines: (i) geodesic ray tracing (GRT) and (ii) a purely supervised learning (SL) regression of a distance field. In all methods, geodesics are extracted \emph{a posteriori} by backtracking the resulting field.

\paragraph{Geodesic Ray Tracing (GRT).}
GRT is a grid-free baseline that traces geodesic rays outward from multiple sources and, recursively, from points visited along those rays (also known as wavefront construction). From each source, we sample unit directions on the Riemannian unit sphere and integrate the geodesic ODE at unit metric speed; each ray assigns an arc length to the points it traverses. The tentative distance at any location is the point-wise minimum of accumulated arc lengths over all rays; values at non-hit locations are interpolated from nearby ray samples. GRT has two main limitations compared to the eikonal formulation~\citep{rawlinson2005fast}:
\begin{enumerate}
  \item \textbf{Optimality and consistency.} The point-wise minimum over traced rays is a sampling-dependent lower envelope. The accumulated arc length from the source to a point $x$ equals the true geodesic distance only up to the first conjugate point along that ray; beyond this point, the ray ceases to be minimizing and its length exceeds the minimal distance~\citep{LawJacobi2021}. Moreover, when directions are re-sampled at intermediate points, segments from different rays can be \emph{spliced}; such stitched paths are not guaranteed to be globally optimal and, in general, do not satisfy the eikonal PDE. Therefore, this approach does not yield a single, globally consistent distance potential.
  \item \textbf{Efficiency, coverage, and parallelization.} Achieving dense coverage requires fine angular/spatial sampling and repeated re-seeding from many intermediate points, which grows with the number of rays and integration steps~\citep{deKool2006}. Additionally, because new rays are computed iteratively based on previously traced rays, the method has sequential dependencies and data-dependent branching, causing warp divergence and irregular memory access, which hampers GPU efficiency~\citep{AilaLaine2009}. In contrast, eikonal solvers (e.g., Fast Marching) compute a dense single-source distance field in a single monotone outward pass, after which geodesics are recovered by inexpensive backtracking~\citep{KimmelSethian1998}.
\end{enumerate}

\paragraph{Supervised Learning (SL).}
A direct alternative is to learn a pairwise distance map by generating ground-truth labels on the fly and minimizing the discrepancy between predicted and labeled distances, e.g.,
\[
\mathcal{L}_{\mathrm{MSE}}
=\mathbb{E}_{(\bm{q}_s,\bm{q})}\!\left[\bigl(U_{\theta}(\bm{q}_s,\bm{q})-\hat d(\bm{q}_s,\bm{q})\bigr)^2\right],
\]
where $\hat d(\cdot,\cdot)$ is obtained by a separate procedure (e.g., solving a two-point boundary-value problem or running geodesic front propagation/FMM to create ground-truth annotations). This approach yields a fast, continuously differentiable distance oracle at inference time, but has key drawbacks: (i) \emph{computational cost}—label generation dominates training and scales with the number of source–query pairs; (ii) \emph{coverage/bias}—supervision is limited to sampled pairs and inherits any inaccuracies or sparsity of the data; (iii) \emph{no PDE structure}—the loss does not enforce the eikonal metric constraint; and (iv) \emph{flow fidelity}—even with accurate distance labels, a neural regressor can produce different gradients, which can deflect paths and yield incorrect geodesic flow. In contrast, our NES eliminates distance labels and trains by enforcing the eikonal equation and boundary conditions, producing a globally consistent field whose gradients are directly suitable for reliable geodesic backtracking.

\paragraph{Evaluation.} We evaluate GRT and SL on the same 2D Riemannian manifold with a kinetic–energy metric. Figure~\ref{fig:compare_gofm_sl} shows distance fields (\emph{left}) and backtracked geodesics (\emph{right}), in comparison to NES (Figure~\ref{fig:nes_2d}a–b). For geodesic ray tracing, we initialize 64 rays from the source with random unit directions and integrate the geodesic ODE at unit metric speed. Every 15 integration steps, we re-seed 16 new rays at the wavefront (from the tips of active rays) and record their accumulated arc lengths. We continue until the total number of rays reaches 5{,}000.
We then interpolate these samples and take the pointwise minimum across all rays to form the tentative distance field. Despite dense sampling, the field exhibits artifacts that induce non-smooth geodesics and inconsistent gradients (\textit{top row}). For supervised learning, we use the GRT field as supervision and train a neural regressor to produce a continuous distance map. This reduces visible artifacts and yields smoother flows (\textit{bottom row}), but inherits GRT’s coverage/bias and requires costly label generation. 

\paragraph{Results.}
Table~\ref{tab:GRT_sl} summarizes geodesic lengths backtracked from the learned distance fields and computation times—split into offline (data collection/training) and online (trajectory integration). RFM and NES recover near-optimal lengths with low online times. GRT yields longer paths and slower inference despite substantial offline cost. SL improves geodesic length and matches NES’s online speed, but requires even more offline time for label generation and training; the complexity of data generation further limits it to single-point eikonal formulations. Overall, eikonal methods produce dense, metrically consistent fields from which geodesics can be reliably backtracked. NES is preferred over SL because it \emph{enforces the eikonal PDE in the neural network during training} rather than regressing to sampled labels; this avoids costly label generation, reduces sampling bias, and imposes a principled, physics-based structure that promotes globally consistent distance fields and stable geodesic backtracking.
}

\begin{table}[t]
\centering
\caption{\textcolor{black}{Comparison of distance-field-based approaches}}
\label{tab:GRT_sl}
\resizebox{\linewidth}{!}{
{\color{black}
\begin{tabular}{lccc}
\toprule
Method & Geodesic Length & Runtime (offline/online) & Type \\
\midrule
RFM  & $6.44 \pm 2.93$ & \textbf{0.58s}/0.58s & Single-point \\
NES  & \textbf{$6.42 \pm 2.95$} & 5min/\textbf{0.27s}          & Source-goal \\
GRT & $6.68 \pm 3.11$ & 30min/10.6s         & Single-point \\
SL   & $6.47 \pm 2.98$ & 32min/0.27s         & Single-point \\
\bottomrule
\end{tabular}
}
}
\end{table}

\subsection{NES for Other Robot Applications}
In addition to minimum–energy motions, we present three further applications that illustrate the generalization capability of NES across broader robotic problems. These tasks instantiate different Riemannian geometries discussed in Section~\ref{sec:riemann-props}, including pullback metrics, stability–aware energy shaping, and obstacle avoidance. In all cases, the eikonal is solved on the chosen geometry, and geodesics are recovered by backtracking as in the previous sections.

\subsubsection{Task–Space Distance Minimization}
\begin{figure}[t]
    \centering
    \includegraphics[width=0.95\linewidth]{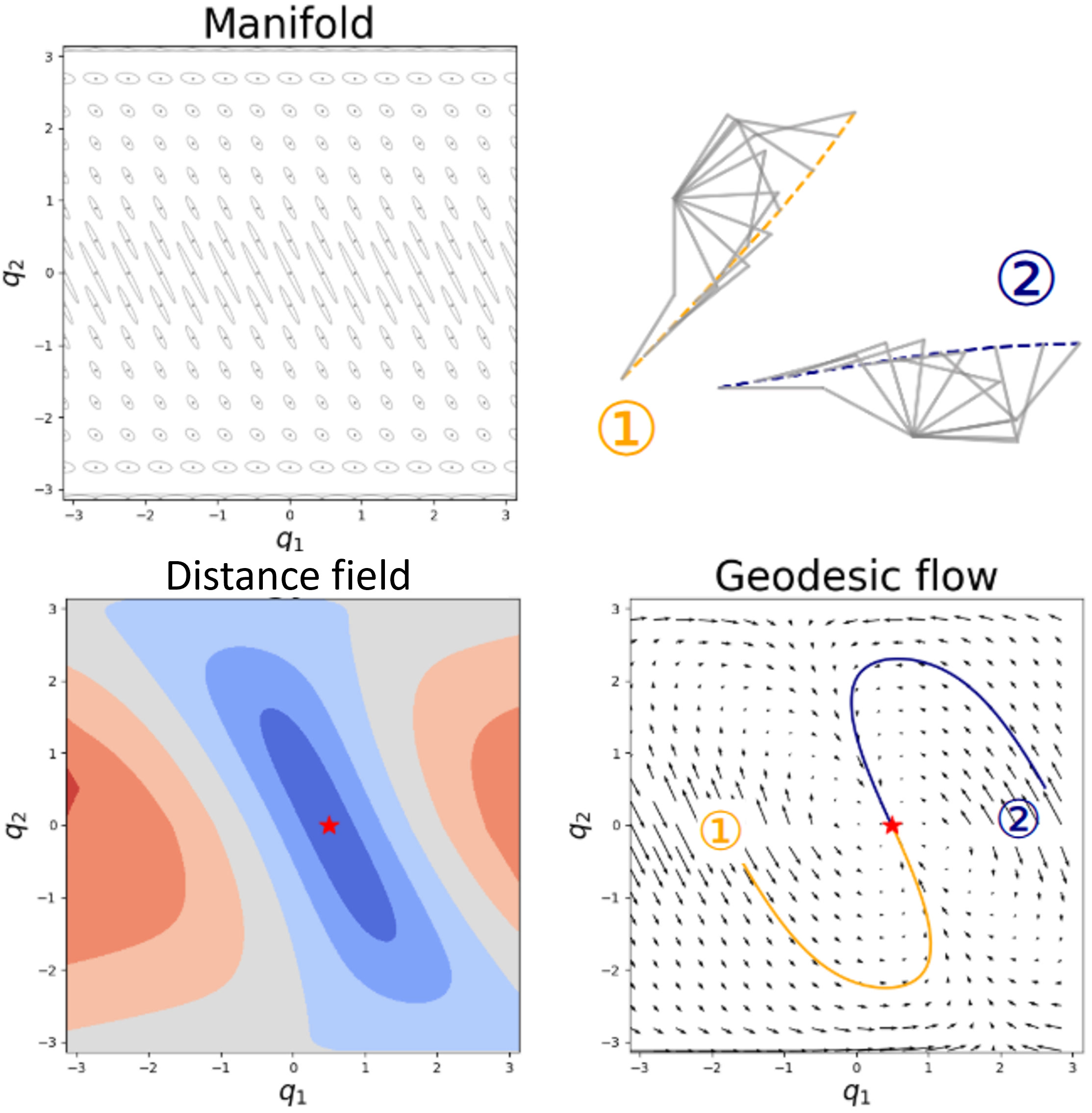}
    \caption{\textcolor{black}{Illustration of task-space distance minimizing motion generation. Geodesics on the Riemannian manifold defined by the metric $\bm{G}(\bm{q}) = \bm{J}^\top \bm{J}$ correspond to motions minimizing end-effector displacement in task space, solved by the Riemannian eikonal equation.}}
    \label{fig:min_task_space_motion}
\end{figure}

We apply our Riemannian distance field framework to generate motions that minimize \emph{task–space} displacement. Using the pullback construction from Section~\ref{sec:riemann-props} with the Euclidean task metric defined by \(\bm{G}_{\mathcal X}=\bm I\), the configuration–space metric is defined by
\[
\bm{G}_{\text{pb}}(\bm{q}) \;=\; \bm{J}(\bm{q})^{\!\top}\,\bm{J}(\bm{q}).
\]
The task–space speed induced by a joint velocity \(\dot{\bm{q}}\) is
\begin{equation}
\label{eq:task-speed}
\|\dot{\bm{x}}\|
\;=\;
\|\bm{J}\dot{\bm{q}}\|
\;=\;
\sqrt{ \dot{\bm{q}}^\top \bm{J}^\top \bm{J} \dot{\bm{q}} }
\;=\;
\|\dot{\bm{q}}\|_{\bm{J}^\top \bm{J}}.
\end{equation}
The minimum–length task–space path corresponds to geodesics under this Riemannian metric, which encodes how joint–space motions translate to task–space displacements.
This formulation minimizes the accumulated task–space displacement, which is desirable for applications such as reaching, drawing, or tool usage, in which end–effector path length matters more than joint–space path length. The Riemannian formulation ensures that among all possible paths connecting two configurations, the one that minimizes task–space travel distance is selected. Note that the task–space paths are not necessarily straight lines, especially for the non-redundant manipulators. This contrasts with task–space interpolation or null–space control, which enforces Cartesian line segments or waypoints that may be kinematically infeasible. An illustration of this task is provided in Figure~\ref{fig:min_task_space_motion}, which depicts the Riemannian manifold, the geodesic distance field and flow, as well as the resulting robot motions using the same 2D planar robot. This formulation can be further extended to incorporate task–specific metrics by choosing \(\bm{G}_{\mathcal X}\succ 0\) to prioritize motions along specific task–space directions.

\subsubsection{Stability–Aware Energy Shaping}
We embed passive stability directly into the planning geometry for a pendulum–on–cart system with two degrees of freedom: the cart position $x$ and the pendulum angle $q$, as illustrated in Figure~\ref{fig:cartpole} (\emph{top-left}).


As introduced in Section~\ref{sec:riemann-props}, we define a position-dependent Riemannian metric that scales inertia with normalized potential using \eqref{eq:passive_energy}. Since the state variables include both cart translation $x$ and pole orientation $q$, we define 
\begin{equation}
\begin{aligned}
\bm{G}_{\text{sta}}(x,q) = \bm{M}(x,q)\left[1 + \epsilon \tilde{P}(x,q)\right], \\
\tilde{P}(x,q) = \frac{P(x,q) - P_{\min}}{P_{\max} - P_{\min}} \in [0,1].
\end{aligned}
\end{equation}
A detailed description is provided in Section~\ref{sec:riemann-props}. Figure~\ref{fig:cartpole} (\emph{top–right}) shows three motions starting from different initial states, all computed under the same energy-shaped metric (\(\epsilon=2.0\)) and same eikonal formulation. As the required cart translation increases, the geodesic flow relies more on low–potential configurations: long moves lower the pole and translate near the down configuration before lifting at the goal; moderate moves lower it partially; short moves stay near upright. This behavior requires no mode switching or heuristics, because we define the Riemannian metric inflates distances in high–energy regions, trajectories naturally dwell in low–energy basins and traverse high–energy postures only when necessary. The corresponding manifold, distance field, and geodesic flow are shown in Figure~\ref{fig:cartpole} (\emph{bottom}).

\begin{table}[t]
    \centering
    \begin{tabular}{cc}
    \raisebox{0.3\height}{\includegraphics[width=0.35\linewidth]{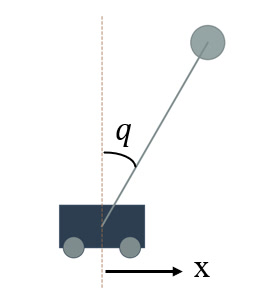}} &
    \includegraphics[width=0.6\linewidth]{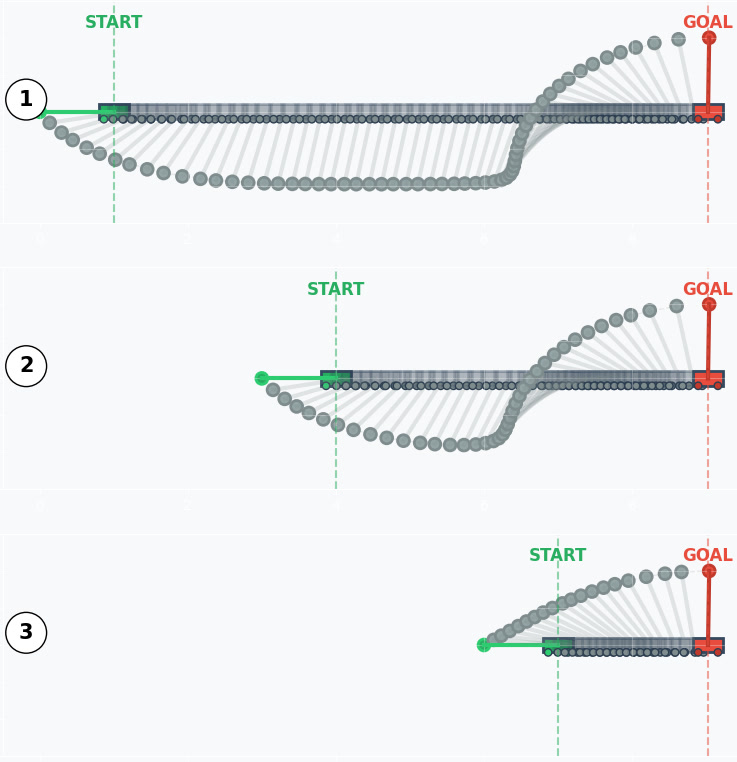} \\
    \multicolumn{2}{c}{\includegraphics[width=1.0\linewidth]{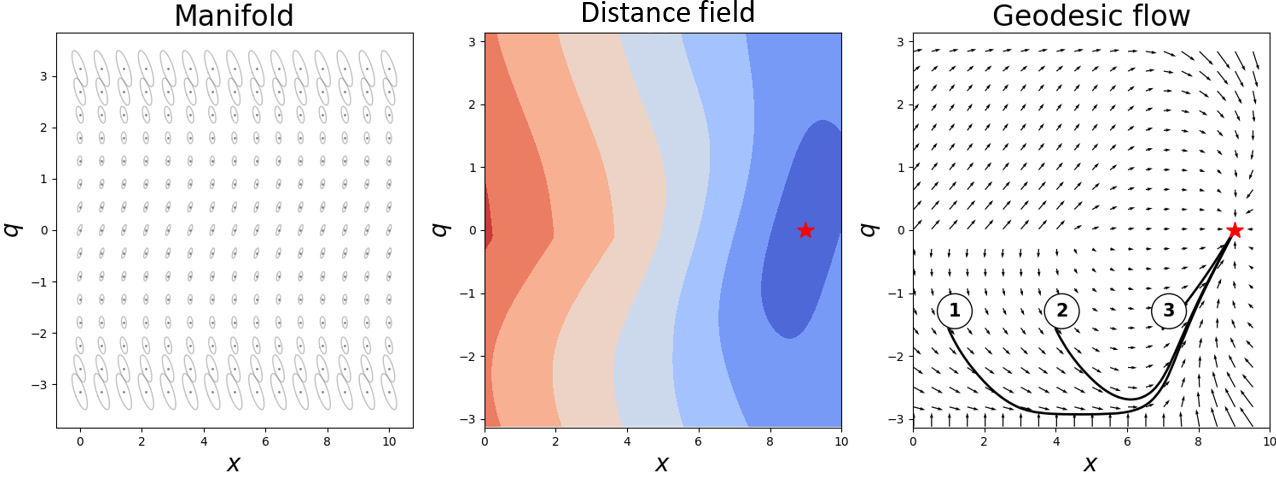}} \\
    \end{tabular}
    \captionof{figure}{\textcolor{black}{Stability-aware motion generation for the pendulum-on-cart system. \textit{Top-left}: configuration space of the system. \textit{Top-right}: representative trajectories from distinct initial conditions, computed via backtracking geodesic flow without tuning. \textit{Bottom}: the Riemannian metric, energy field, and induced geodesic flow for a fixed goal.}}
    \label{fig:cartpole}
\end{table}

\subsubsection{Collision avoidance}

Eikonal-based distance fields and geodesic flows handle collision avoidance by making obstacles impassable in the PDE. In the isotropic form \eqref{eq:eikonal}, this can be modeled by assigning zero-speed velocity to points inside obstacles $\mathcal{O}$. This yields global optimal paths that respect hard collision constraints and have shown promising results in motion planning~\citep{ni2022ntfields,ni2023progressive}. The same idea extends from isotropic to anisotropic settings by replacing the Euclidean norm with the Riemannian metric. Figure~\ref{fig:ob} illustrates the distance field (a) and flows/paths (b) on the same configuration–space manifold endowed with the kinetic–energy metric described in Section~\ref{sec:nes-meg}, now with obstacles present. Solving the eikonal equation produces a global distance field and corresponding optimal paths without additional asymptotic computational complexity.

\begin{table}[t]
    \begin{center}
        \begin{tabular}{cc}
            \centering
            \includegraphics[width =0.44\linewidth
            ]{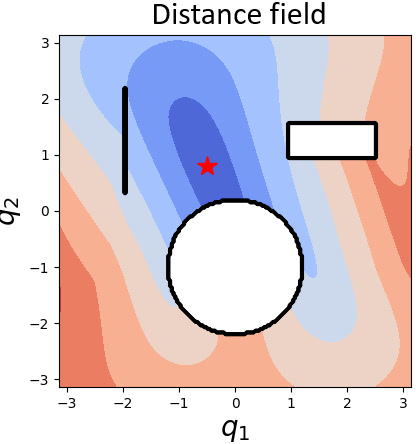}   &
            \includegraphics[width =0.45\linewidth]{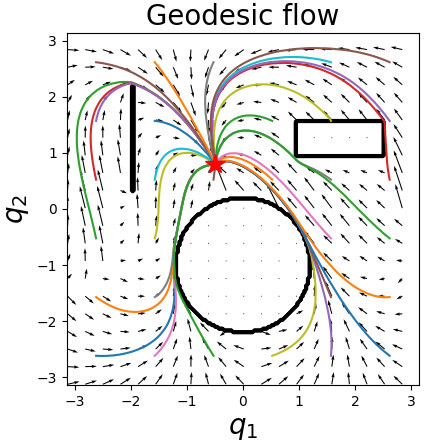}  \\

            (a) & (b) 
        \end{tabular}
        \captionof{figure}{Solution of the Riemannian eikonal equation in case of obstacles (black contours). (a) distance field. (b) Geodesic flow.}
        \vspace{-6mm}
        \label{fig:ob}
    \end{center}
\end{table}

However, because proximity to obstacles is not explicitly encoded, the resulting paths can exhibit vanishing clearance, which reduces robustness and elevates collision risk. To address this limitation, we replace the hard, non-smooth boundary (i.e., zero-speed zones) with a smoothly varying weight matrix that continuously reflects proximity to obstacles, using the Riemannian eikonal formulation described in Section~\ref{sec:riemann-props}. This formulation increases the cost of moving toward obstacles while preserving flexibility in tangential directions. As a result, paths naturally bend around obstacles, effectively maintaining smooth and safe clearance. Figure~\ref{fig:ca_reik} compares paths generated by the standard isotropic eikonal formulation (a) and our Riemannian manifold-based approach (b). The latter produces trajectories that are smoother with a more natural usage of obstacle proximity information.
\begin{table}[t]
    \begin{center}
        \begin{tabular}{cc}
            \centering
            \includegraphics[width =0.4\linewidth
            ]{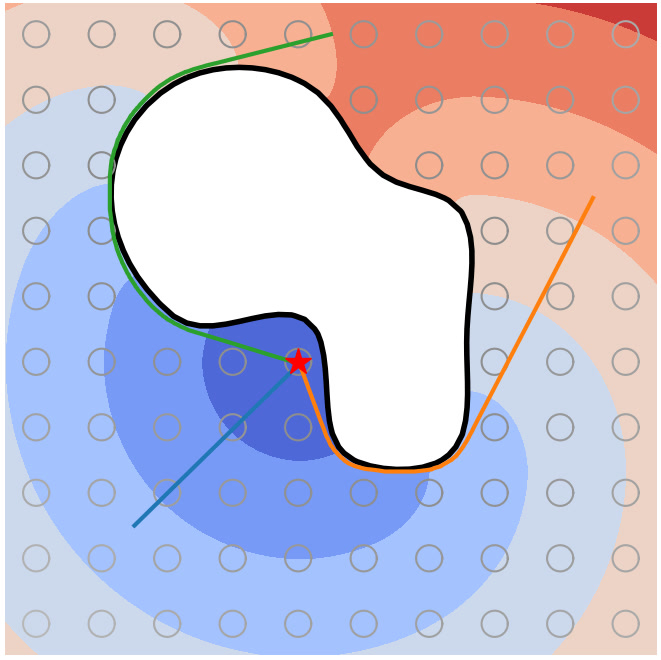}   &
            \includegraphics[width =0.4\linewidth]{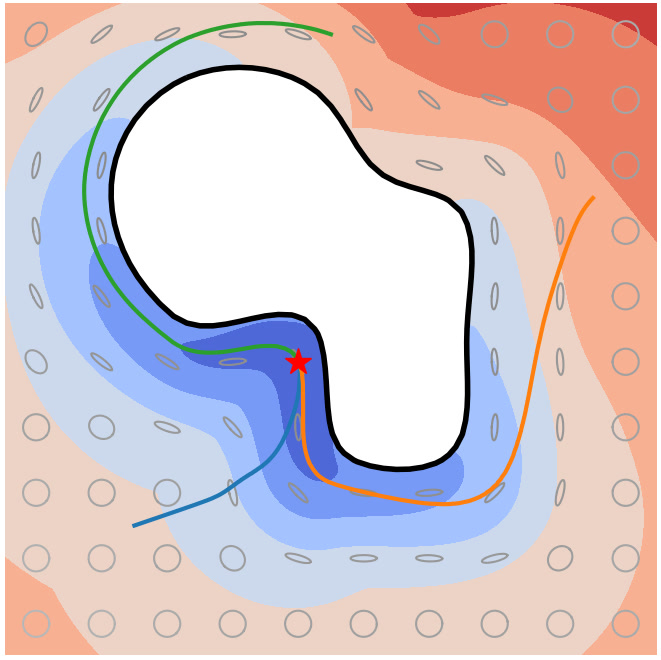}  \\
            (a) & (b) 
        \end{tabular}
        \captionof{figure}{Obstacle-avoidance comparison. (a) Trajectory generated by the standard isotropic eikonal equation. (b) Trajectory generated using a Riemannian eikonal formulation.}
        \vspace{-6mm}
        \label{fig:ca_reik}
    \end{center}
\end{table}

\section{Experiments}\label{sec:exp}
{\color{black}
We evaluate \emph{NES} on motion planning problems to demonstrate its ability to generate \emph{dynamics-aware, energy-efficient} trajectories—a long-standing challenge in robotics. From a Riemannian-geometry viewpoint, Lagrangian and Hamiltonian mechanics induce the kinetic-energy metric and the Jacobi metric respectively (Section~\ref{sec:riemann-props}). NES seeks geodesics under these metrics, yielding paths that respect system dynamics while reducing energy. The underlying robot dynamics on configuration manifolds and the connection between geodesics and optimal control are summarized in Appendix~\ref{sec:dynamics}.
}

The experiments are designed to investigate the following key questions:
\begin{itemize}
    \item \textbf{Q1:} How effective and efficient is NES in computing distance fields and geodesic paths on Riemannian manifolds?
    \item \textbf{Q2:} How does our method compare to commonly used approaches for computing geodesic paths?
    \item \textbf{Q3:} In which ways can the geodesic path on a Riemannian manifold account for reduced control inputs?
    \item \textbf{Q4:} How well can our neural Riemannian eikonal solver scale to high-dimensional robot manipulation problems?
    \item \textbf{Q5:} How effective is our C-NES approach at propagating wavefronts from joint space to task space?
    \item \textbf{Q6:} How easily can our approach be integrated into other motion optimization frameworks to facilitate dynamics-aware motion planning?
\end{itemize}

To address these questions, we conducted a series of experiments in both simulation and real-world settings by considering two robot manipulators: a 2-DoF planar robot described in Section~\ref{sec:examples}, and a 7-axis Franka robot.
The inertial mass matrix and potential energy of this robot are derived from the robot's Unified Robot Description Format (URDF) file using the Composite Rigid Body Algorithm \citep{walker1982efficient}, and we modified the implementation by \citep{urdf2casadi} to enable batch computation in PyTorch.

\subsection{Baselines}

Computing geodesics on a Riemannian manifold is challenging due to the metric’s nonlinearity and anisotropy. \textcolor{black}{In Section~\ref{sec:examples}, we demonstrate the effectiveness of NES for computing distance fields and geodesic flows, and we compare it with other distance–field–based methods. Here, we further compare NES with classical approaches that directly compute geodesics.
}

\textbf{Geodesic Shooting (GS)}: This method solves an initial value problem by numerically integrating the geodesic equation forward in time, starting from a given initial configuration and velocity. A key characteristic of GS is that it does not require explicit knowledge of the goal configuration in advance. Instead, it relies on choosing an initial velocity vector that heuristically points approximately toward the desired target. That differs from our NES method, which directly finds the geodesic path between start and goal pairs.

\textbf{Optimal Control (OC)}: This approach formulates geodesic computation as a boundary value problem and solves it through iterative optimization techniques. Specifically, we optimize the objective: 
\begin{equation}
\begin{aligned}
\label{eq:oc}
& \min_{\{\bm{u}_{\zeta}\}} \quad && \|\bm{q}_N - \bm{q}_s\|_{\bm{Q}}^2 + r \sum_{\zeta=1}^{N-1} \bm{u}_{\zeta}^\top \bm{G}(\bm{q}_{\zeta}) \bm{u}_{\zeta} \\
& \text{subject to} \quad && \bm{q}_{\zeta} = \bm{q}_{\zeta-1} + \bm{u}_{\zeta} \, d\zeta, \quad \zeta=1,\dots,N \\
& && dt = \sqrt{\bm{u}_{\zeta}^\top \bm{G}(\bm{q}_{\zeta}) \bm{u}_{\zeta}} \, d\zeta,
\end{aligned}
\end{equation}
where \(\bm{q}_{\zeta}\) is the system state indexed by the phase parameter \(\zeta\), which serves as a reparameterization of time along the trajectory corresponding to the arc-length under the Riemannian metric defined by \(\bm{G}\). The cost minimizes the weighted squared distance between the final state \(\bm{q}_N\) and the target \(\bm{q}_s\) along with a regularization term on control effort weighted by \(\bm{G}(\bm{q}_{\zeta})\). The dynamics constraint describes the discrete evolution of states with respect to \(\zeta\). The time reparameterization equation connects the physical time increment \(dt\) to the control magnitude measured by the energy metric, ensuring that \(\zeta\) parameterizes the trajectory consistently with the system’s geometry and temporal evolution. We refer to this baseline as the \textit{constant energy path} (CEP), as the optimal control approach seeks the shortest geodesics on the configuration space manifold defined by the Jacobi metric, representing constant-energy motion.


A brief overview of these baselines and our approach is listed in Table~\ref{tab:geodesic_methods_comparison}. In contrast to GS and OC, which rely on numerical optimization or iterative integration and are often sensitive to initialization, our NES learns a differentiable approximation to the geodesic distance field. Once the model is trained, it can be used to query distances and geodesic flows efficiently and can be generalized to arbitrary
start-to-goal joint configuration pairs. NES provides a closed-form policy by evaluating the analytic gradient of the learned field, enabling fast and generalizable geodesic inference across the configuration space. Unlike OC-based methods, which are typically computationally expensive and less flexible for real-time control, NES supports dynamic, efficient, and scalable geodesic computation.

\begin{table*}[h]
\centering
\small
\caption{\textcolor{black}{Comparison of Geodesic Computation Methods}}
\arrayrulecolor{black} 
\begin{tabular}{|>{\color{black}}l|>{\color{black}}c|>{\color{black}}c|>{\color{black}}c|}
\hline
\textbf{Criterion} & \textbf{Optimal Control} & \textbf{Geodesic Shooting} & \textbf{Neural Eikonal Solver (Ours)} \\ \hline
\textbf{Output} & Policy / Path & Policy / Path & Field + Policy / Path \\ \hline
\textbf{Optimality} & Local & Local & Approx. Global \\ \hline
\textbf{Online Cost} & High & Low & Low \\ \hline
\textbf{Init. Sensitivity} & High & High & Low \\ \hline
\textbf{Flexibility} & Low & High & High \\ \hline
\textbf{Differentiability} & Indirect & Indirect & Explicit \\ \hline
\end{tabular}
\arrayrulecolor{black} 
\label{tab:geodesic_methods_comparison}
\end{table*}

\subsection{Control Strategies}
\label{sec:cs}
In our experiments, we exploit a key property of NES: rather than solving a new geodesic problem for each start–goal pair, it learns offline a continuous, differentiable distance field on the configuration manifold. At runtime, we query the distance field and its analytic gradient efficiently to recover geodesic directions, enabling high rate in feedback loops. Thus, in practice, NES is not only a geodesic solver but also a compact policy/value representation that integrates naturally with standard controllers to handle constraints. In other words, NES provides high-level guidance on the manifold via the geodesic flow, while a separate low-level controller enforces actuation limits, kinematic/dynamic constraints, and tracking accuracy. This separation preserves constraint awareness encoded in the metric and distance field, and affords flexibility and adaptability in dynamic environments.

We observed that directly following the learned geodesic flow $-\bm{V}(\bm{q})$ can lead to numerical instability, especially when $\bm{q}$ is close to the target and $U(\bm{q})$ approaches zero. 
To ensure stable and robust convergence near the target in our tests, we define a configuration-dependent desired joint velocity $\dot{\bm{q}}_{\text{des}}(\bm{q})$ by blending the learned geodesic flow direction $-\bm{V}(\bm{q})$ with a simple linear vector $(\bm{q}_s-\bm{q})$ pointing directly toward $\bm{q}_s$. 
The blending weight $\lambda(\bm{q})$ is dynamically adjusted based on $U(\bm{q})$, allowing a smooth transition between global and local guidance:
\begin{equation}
\label{eq:blended_control}
\begin{aligned}
    \lambda(\bm{q})&=\frac{\beta\,U(\bm{q})}{1+\beta\,U(\bm{q})},\\
\dot{\bm{q}}_{\text{des}}(\bm{q})&=-\lambda(\bm{q})\,\bm{V}(\bm{q})+\big(1-\lambda(\bm{q})\big)\,(\bm{q}_s-\bm{q}).
\end{aligned}
\end{equation}
Here $\beta>0$ modulates the sensitivity to $U(\bm{q})$. When the robot is far from the target and $U(\bm{q})$ is large, $\lambda(\bm{q})\!\approx\!1$ and the motion follows $-\bm{V}(\bm{q})$; as $U(\bm{q})$ becomes small, $\lambda(\bm{q})\!\to\!0$ and the attraction term ensures stable convergence.

Below, we illustrate representative integrations used in our experiments, including kinematic and kinodynamic QP tracking and task-priority control via null-space projection. Other approaches, such as model predictive control (MPC)~\citep{koptev2024reactive} can also be exploited. 

\subsubsection{Quadratic Programming (QP)}

The blended velocity $\dot{\bm{q}}_{\text{des}}(\bm{q})$ specifies an instantaneous target motion.  
We compute a feasible command that respects limits and dynamics by solving a QP.  
We give kinematic (velocity) and kinodynamic (torque/acceleration) forms.

\paragraph{Kinematic QP Formulation}
At the joint velocity level, the control command is the joint velocity $\dot{\bm{q}} \in \mathbb{R}^n$. This QP is formulated to compute the feasible joint velocity $\dot{\bm{q}}$ that closely tracks the blended desired velocity $\dot{\bm{q}}_{\text{des}}(\bm{q})$, while satisfying joint velocity and position limits. The objective is to minimize the weighted squared error between the actual and desired velocities, with the cost function defined by a symmetric positive semi-definite matrix $\bm{Q}$:
\begin{equation}
\label{eq:kinematic_qp}
\begin{aligned}
    \min_{\dot{\bm{q}} \in \mathbb{R}^n} \quad & (\dot{\bm{q}} - \dot{\bm{q}}_{\text{des}}(\bm{q}))^{\top} \bm{Q} (\dot{\bm{q}} - \dot{\bm{q}}_{\text{des}}(\bm{q})), \\
    \text{s.t.} \quad & \dot{\bm{q}}_{\min} \leq \dot{\bm{q}} \leq \dot{\bm{q}}_{\max}, \\
    \quad & \bm{q}_{\min} \leq \bm{q} + \Delta t \dot{\bm{q}} \leq \bm{q}_{\max}.
\end{aligned}
\end{equation}
The weighting matrix $\bm{Q}$ allows prioritizing tracking accuracy across different joints. Joint position and velocity limits are applied to constrain the motion within the robot's physical limits.

\paragraph{Kinodynamic QP Formulation}
At the joint torque level, where the control input is $\bm{\tau} \in \mathbb{R}^n$, it is necessary to incorporate the robot's dynamics, typically modeled as $\bm{\tau} = \bm{M}(\bm{q})\ddot{\bm{q}} + \bm{C}(\bm{q}, \dot{\bm{q}})\dot{\bm{q}} + \bm{G}(\bm{q})$. To utilize the desired velocity $\dot{\bm{q}}_{\text{des}}(\bm{q})$, we first compute a nominal desired joint acceleration $\ddot{\bm{q}}_{\text{des}}$ aimed at tracking the desired velocity over time:
\begin{equation}
    \ddot{\bm{q}}_{\text{des}}(\bm{q}, \dot{\bm{q}}) = \frac{\dot{\bm{q}}_{\text{des}}(\bm{q}) - \dot{\bm{q}}}{\Delta t},
\end{equation}
where $\Delta t$ is the control time step. A nominal desired torque $\bm{\tau}_{\text{des}}$ required to achieve the target acceleration is computed using inverse dynamics:
\begin{equation}
    \bm{\tau}_{\text{des}}(\bm{q}, \dot{\bm{q}}, \ddot{\bm{q}}_{\text{des}}) = \bm{M}(\bm{q})\ddot{\bm{q}}_{\text{des}} + \bm{C}(\bm{q}, \dot{\bm{q}})\dot{\bm{q}} + \bm{G}(\bm{q}).
\end{equation}
We formulate a kinodynamic QP with both joint torque $\bm{\tau}$ and joint acceleration $\ddot{\bm{q}}$ as decision variables. This formulation allows us to explicitly include the robot's dynamics as an equality constraint while minimizing the deviation between the applied torque $\bm{\tau}$ and the nominal desired torque $\bm{\tau}_{\text{des}}$. The resulting QP is defined as:
\begin{equation}
\label{eq:kinodynamic_qp}
\begin{aligned}
    \min_{\substack{\bm{\tau} \in \mathbb{R}^n \\ \ddot{\bm{q}} \in \mathbb{R}^n}} \quad & (\bm{\tau} - \bm{\tau}_{\text{des}}(\bm{q}, \dot{\bm{q}}, \ddot{\bm{q}}_{\text{des}}))^{\top} \bm{Q} (\bm{\tau} - \bm{\tau}_{\text{des}}(\bm{q}, \dot{\bm{q}}, \ddot{\bm{q}}_{\text{des}})), \\
    \text{s.t.} \quad & \bm{\tau} - \bm{M}(\bm{q})\ddot{\bm{q}} = \bm{C}(\bm{q}, \dot{\bm{q}})\dot{\bm{q}} + \bm{G}(\bm{q}), \\
    & \bm{\tau}_{\min} \leq \bm{\tau} \leq \bm{\tau}_{\max}, \\
    & \ddot{\bm{q}}_{\min} \leq \ddot{\bm{q}} \leq \ddot{\bm{q}}_{\max}, \\
    & \dot{\bm{q}}_{\min} \leq \dot{\bm{q}} + \Delta t \ddot{\bm{q}} \leq \dot{\bm{q}}_{\max}, \\
    & \bm{q}_{\min} \leq \bm{q} + \Delta t \dot{\bm{q}} + \tfrac{1}{2}\Delta t^2 \ddot{\bm{q}} \leq \bm{q}_{\max}.
\end{aligned}
\end{equation}
In this formulation, $\bm{Q}$ is a symmetric positive semi-definite weighting matrix for the torque error objective. The constraints enforce the robot's dynamics, as well as joint torque, acceleration, velocity, and position limits. Future joint velocities and positions are approximated using first- and second-order Euler integration, respectively. The kinodynamic QP finds the optimal feasible torque and acceleration pair that satisfies the robot's dynamics and physical limits while minimizing the weighted deviation from the nominal torque required to achieve the desired acceleration derived from the blended velocity $\dot{\bm{q}}_{\text{des}}(\bm{q})$, offering a simple yet effective framework for integrating the learned geodesic flows into real-time control strategies. Furthermore, this framework is extensible and can incorporate additional constraints, such as collision avoidance~\citep{li2024representing,koptev2022neural}. 

\subsubsection{Task Prioritization}

In practice, while QP-based controllers offer flexibility for optimizing multiple objectives simultaneously, they can yield suboptimal results when objectives conflict or require strict prioritization. In weighted optimization schemes, improper tuning of weights can lead to violations of critical tasks. Task prioritization addresses this limitation by enforcing a strict hierarchy: higher-priority tasks are satisfied strictly, while lower-priority behaviors are projected into the null space of higher-priority constraints. This method is effective for simultaneous objective achievement, such as maintaining end-effector orientation while avoiding obstacles \citep{khatib1987unified,ratliff2018riemannian}.

Here we integrate the geodesic flow policies learned by NES into a task prioritization framework to enhance robotic performance while strictly satisfying the primary task. Typically, this policy is integrated as a lower-priority task, contributing to motion generation without violating higher-priority constraints. Although projecting into the null space may compromise some global optimality of the learned policy, the resulting motion remains near-optimal by leveraging the robot’s residual redundancy.

For a constraint manifold defined by $f(\bm{q})$, the joint velocity $\dot{\bm{q}}$ must satisfy:
\begin{equation}
    \bm{J}_f(\bm{q}) \dot{\bm{q}} = \bm{0},
\end{equation}
where $\bm{J}_f(\bm{q}) \triangleq \frac{\partial f}{\partial \bm{q}}$ is the constraint Jacobian. This ensures all motion remains tangent to the constraint surface. The desired geodesic flow $\dot{\bm{q}}_{\text{des}}$ is projected into the constraint null space:
\begin{equation}
    \dot{\bm{q}}_{\text{TP}} = \bm{N}_f \dot{\bm{q}}_{\text{des}},
    \label{eq:null_space_projection}
\end{equation}
where $\bm{N}_f \triangleq \bm{I} - \bm{J}_f^{\dagger} \bm{J}_f$ is the null-space projector, $\bm{J}_f^{\dagger}$ is the Moore--Penrose pseudoinverse of $\bm{J}_f$, and $\dot{\bm{q}}_{\text{TP}}$ is the task-prioritized velocity.
By integrating our geodesic flow policy through null-space projection, we achieve simultaneous constraint satisfaction and trajectory optimization. The framework ensures strict adherence to constraints $\bm{J}_f(\bm{q})\dot{\bm{q}} = \bm{0}$ while preserving the optimized trajectory in the remaining degrees of freedom. As a result, the robot's motion retains the optimality properties of the original policy wherever feasible, while ensuring that all specified constraints are strictly satisfied.

\subsection{Training Details and Evaluation Metrics}
We use a multilayer perceptron (MLP) as the backbone of our NES to train the neural Riemannian eikonal solver for the 7-DoF Franka robot. The MLP takes concatenated source-to-goal pairs as input and outputs a scalar Riemannian distance $U$, and its gradient is obtained analytically via automatic differentiation in PyTorch. To better capture the local geometric structure of the configuration space manifold, we utilize a geometry-aware sampling strategy—specifically, the Riemannian Manifold Metropolis-Adjusted Langevin Algorithm (RM-MALA)—to generate input point pairs that lie on the manifold. A detailed explanation of this sampling method is provided in Appendix~\ref{sec:geometry-aware-sampling}. During training, we randomly sample 50,000 joint configurations $\bm{q}_e$ at each epoch as wavefront boundary points, drawn from the distribution produced by the RM-MALA algorithm. The model is trained for up to $10^5$ epochs using the Adam optimizer with a learning rate of 0.001, terminating early upon convergence. Training takes approximately 5 minutes for the 2-DoF planar robot and around four hours for the 7-DoF Franka robot, using a single NVIDIA RTX 3090 GPU.

We evaluate our approach using the following metrics:

\textbf{Geodesic Length:} The geodesic length is computed via \eqref{eq:geodesic}, which is a basic metric to measure the quality of the constructed distance field.

{\color{black}
\textbf{Total Torque:} We measure control effort as
\begin{equation}
    \|\bm{\tau}\|_{\bm{R}} \triangleq \Big(\int_0^T \bm{\tau}^{\top}\bm{R}\bm{\tau}\,\mathrm{d}t\Big)^{1/2}.
\end{equation}
We report two variants: (i) \textit{norm torque} with $\bm{R}=\bm{I}$ (standard $\ell^{2}$ norm); and (ii) \textit{active torque} with $\bm{R}=\dot{\bm{q}}\dot{\bm{q}}^{\top}$, whose integrand is $(\bm{\tau}^{\top}\dot{\bm{q}})^2$, indicating torque that does mechanical work). Ideally, for geodesics under the Jacobi metric, the motion is energy-conserving, so no active work is required, and this term vanishes; see Appendix~\ref{sec:control_effort_geodesics} for a detailed discussion.
}

\textbf{Computation Time:} We measure the total computation time required to generate control inputs and produce the full trajectory.

\subsection{Results}

\subsubsection{Dynamics-aware motions generation on configuration space manifold} 

We first evaluate dynamics-aware, energy-efficient motion generation on the configuration-space manifold. Specifically, we compare NES against baselines in computing minimum-length paths on the energy-conserving manifold induced by the Jacobi metric. \textcolor{black}{As discussed in Section~\ref{sec:nes-meg}, we set the total energy to \(H = 1.2\,P_{\text{max}}\) to remain in a low-energy regime while avoiding singularities.} Table~\ref{tab:results} reports results for both 2D and 7D robotic systems. We randomly sample 100 start-goal pairs for geodesic computation and compare NES with Geodesic Shooting (GS), Optimal Control (OC), and RFM. For NES, we evaluate two tracking controllers described in Section~\ref{sec:cs}: a velocity-based QP controller (Vel) and a torque-based QP controller (Tau) for following the planned trajectories. From these experiments, we extract the following key insights:

\textbf{Energy-efficient motions via geodesic planning (Q1, Q2):} In the 2D planar simulation, NES achieves comparable geodesic lengths and lower torque consumption than RFM, and significantly outperforms both GS and OC in terms of active torque and total control effort. In the 7D case, NES generates the shortest geodesics and lowest active torques among all evaluated methods. These results validate our formulation: instead of solving dynamics-aware motion planning via trajectory optimization, we compute geodesics on a configuration-space manifold shaped by the robot's energy structure. This leverages the geometry of robot dynamics to find globally optimal solutions.

\textbf{Relation to minimum control effort (Q3):}
Our method consistently produces trajectories with lower active torque consumption by planning paths along geodesics on the energy manifold. This further results in reduced $\ell^2$ torque, a standard metric in optimal control for measuring the control effort. These results align with our theoretical analysis in Appendix~\ref{sec:control_effort_geodesics}, showing that geodesic paths on the energy-conserving manifold often correlate with lower control effort.  Notably, this performance is achieved without explicitly optimizing torque-based objectives, demonstrating that our geometric formulation inherently promotes energy-efficient behaviors.
        
\textbf{Decoupling planning and control for scalability and robustness (Q2, Q4, Q6):}
A key advantage of our framework is the clear separation between motion planning and control. Traditional optimal control tightly couples these components, solving high-dimensional nonlinear trajectory optimization problems at runtime while satisfying system constraints at every timestep. This approach is not only computationally expensive but also highly sensitive to initialization, often resulting in suboptimal solutions due to local minima. In contrast, our NES precomputes a global distance field and corresponding geodesic paths over the configuration space manifold. These geodesics encode dynamic feasibility and energy optimality, serving as reference trajectories. The control task is then reduced to a lightweight tracking problem solved by standard QP-based velocity or torque controllers. This decoupling enables real-time execution and robustness to disturbances. As evidenced in Table~\ref{tab:results}, it substantially improves computational efficiency, making our method scalable to high-DoF systems and adaptable in reactive scenarios.
    
\textbf{Flexibility and adaptability (Q1):} Unlike RFM, which relies on grid-based discretization, NES learns a continuous and differentiable representation of the distance field. RFM must solve the eikonal equation from scratch using iterative wavefront propagation for each new start point. This process severely limits its scalability and computational efficiency. In contrast, NES supports parallel computation over batch inputs, enabling rapid generalization across a wide range of initial conditions. It also avoids interpolation errors and scales effectively to high-dimensional configuration spaces. As shown in Table~\ref{tab:results}, NES performs reliably on high-dimensional platforms where RFM fails.

\begin{figure*}
    \centering
    \includegraphics[width=1.0\linewidth]{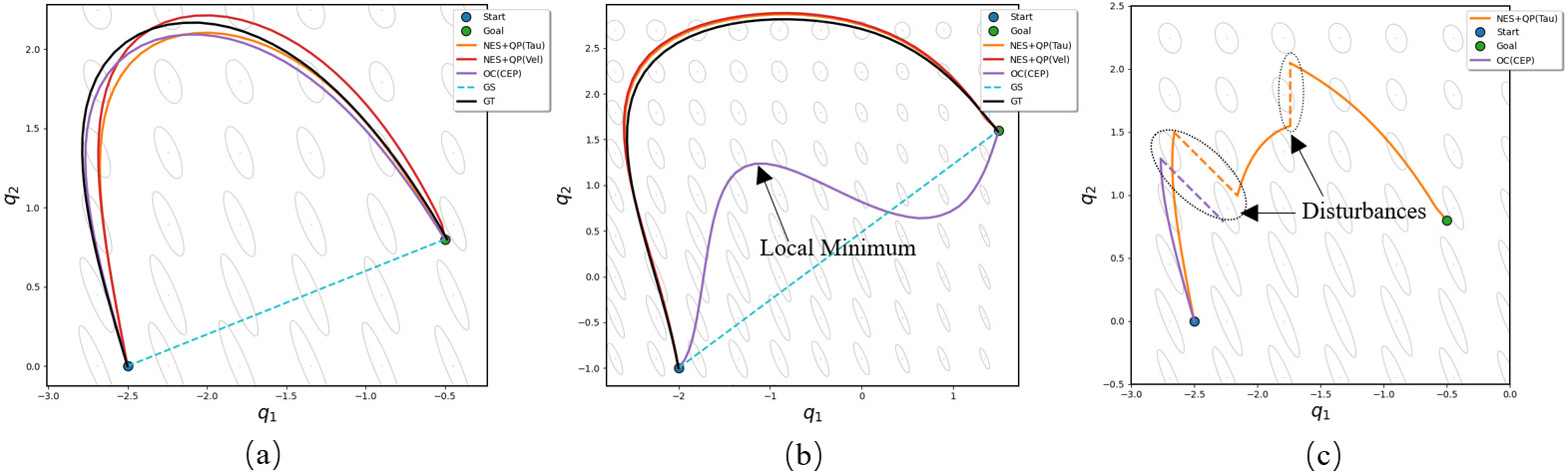}
    \caption{\textcolor{black}{Comparison between our NES with a QP controller and baseline methods. NES matches OC in short-range tasks (a), outperforms long-range cases (b) by avoiding local minima, and is robust to (c) disturbances.}}
    \label{fig:enter-label}
\end{figure*}

A comprehensive comparison of our approach and baselines is shown in Figure \ref{fig:enter-label} (\textbf{Q2}). The ground truth trajectory is derived from our RFM approach. In (a), we observe that both NES and optimal control successfully approximate the energy-optimal path if the start and goal points are close to each other.  However, if the start and goal points are far apart, as shown in (b), the optimal control method becomes trapped in a local minimum, while our NES approach still converges to the near-optimal solution. In both cases, geodesic shooting finds local minimum solutions. Figure \ref{fig:enter-label}(c) further highlights NES's robustness to disturbances. Thanks to its efficient computation and generalization ability, NES adapts to varying start-to-goal conditions and maintains reliable performance under the disturbance.

\subsubsection{Correlation with Minimum-Torque Control}
To further verify the connection between geodesics on the energy-conserving manifold and the minimum control-effort problem (\textbf{Q2, Q3}), we consider an additional optimal control problem that minimizes torque:
\begin{equation}
\label{eq:oc_mtp}
\begin{aligned}
&\min_{\{\bm{u}_t\}} \quad && \| \bm{q}_T - \bm{q}_g \|_{\bm{Q}}^{2} + \frac{r}{2}\sum_{t=1}^{T-1} 
\bm{u}_t^{\top} \bm{R}\,\bm{u}_t, \\
&\text{s.t.} \quad && \bm{q}_{t+1} = FD(\bm{q}_t,\bm{u}_t),
\end{aligned}
\end{equation}
where \(FD(\bm{q}_t,\bm{u}_t)\) denotes the system dynamics derived from \eqref{eq:eom}, and \(\bm{Q} \succeq 0\), \(\bm{R} \succ 0\) are weighting matrices. We refer to this formulation as the \textit{minimum-torque} (MT) problem. Unlike previous baselines, this formulation does not strictly enforce the Riemannian manifold constraints; instead, it serves as a proof of concept to illustrate the relationship between geodesic motions on the energy-conserving manifold and the minimization of control effort.

\begin{figure}
    \centering
    \includegraphics[width=1.0\linewidth]{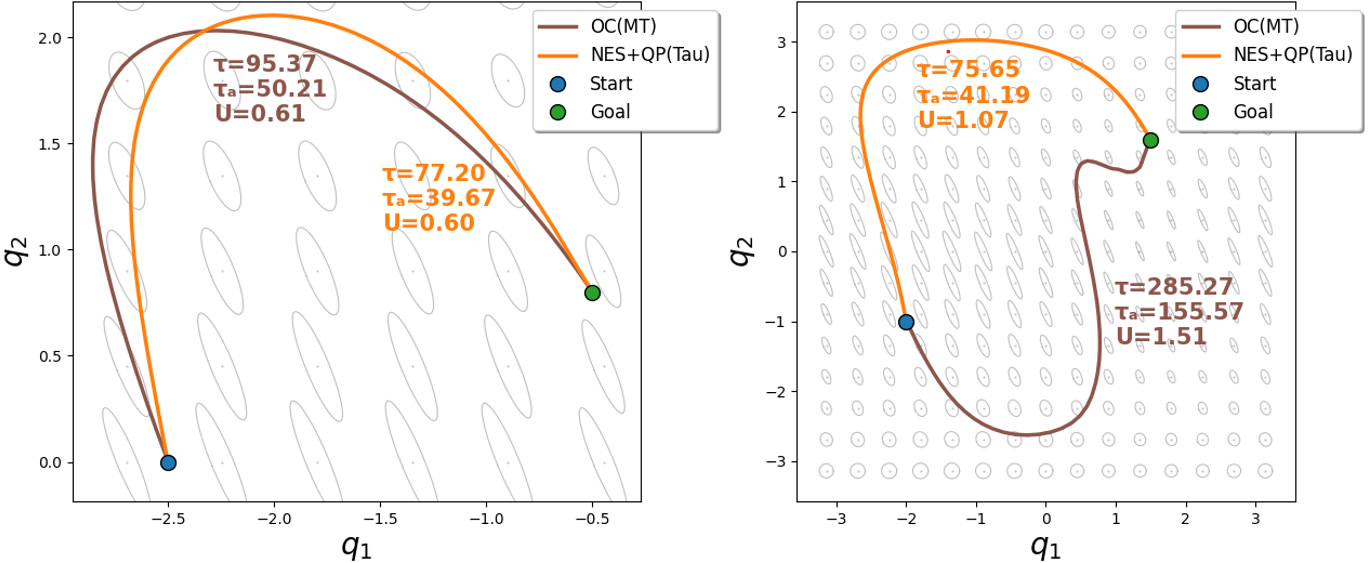}
    \caption{\textcolor{black}{Comparison between our NES with a QP controller and optimal control approach with minimum torque inputs.}}
    \label{fig:compare_mt}
\end{figure}

We compare NES with a joint-torque controller as the low-level actuator against this optimal-control formulation that penalizes control effort. Qualitative comparisons are presented in Figure~\ref{fig:compare_mt}. Our NES approach consistently generates trajectories that closely follow geodesics on the energy-conserving manifold, resulting in paths that are qualitatively similar to those obtained from the minimum-torque solution computed via the optimal-control formulation (as illustrated in Figure~\ref{fig:compare_mt}-\textit{left}). \textcolor{black}{This observation supports using geodesics on the energy-conserving manifold as a geometric prior for optimal-control objectives that penalize control effort. Although friction, modeling/tracking errors, and boundary-condition constraints introduce non-conservative energy expenditure, the geodesic path still provides a principled heuristic that serves as a high-quality initial guess close to the optimal solution.} In Figure~\ref{fig:compare_mt}-\textit{right}, NES outperforms the minimum-torque solution obtained from optimal control. A critical limitation of the optimal-control formulation is its reliance on manual tuning of the weighting matrices \(\bm{Q}\) and \(\bm{R}\), which introduces subjectivity and can yield suboptimal solutions if not tuned carefully. In contrast, NES derives a control policy inherently aligned with the system’s dynamics, enabling adaptive and efficient energy use without heuristic parameter tuning.

\begin{table*}[ht]
\centering
\caption{\textcolor{black}{Comparison of Baselines on Energy-Efficient Motion Generation}}
\arrayrulecolor{black} 
\resizebox{\linewidth}{!}{
\begin{tabular}{|>{\color{black}}c|>{\color{black}}c>{\color{black}}c>{\color{black}}c>{\color{black}}c|>{\color{black}}c>{\color{black}}c>{\color{black}}c>{\color{black}}c|}
\hline
\multicolumn{1}{|>{\color{black}}c|}{\multirow{2}{*}{\textbf{Method}}} & \multicolumn{4}{>{\color{black}}c|}{\textbf{Planar Robot (2D)}} & \multicolumn{4}{>{\color{black}}c|}{\textbf{Franka Robot (7D)}} \\ \cline{2-9}
\multicolumn{1}{|>{\color{black}}c|}{} & \textbf{Geodesic Length} & \textbf{L2 Torque} & \textbf{Active Torque} & \textbf{Time (Policy/Path)} & \textbf{Geodesic Length} & \textbf{L2 Torque} & \textbf{Active Torque} & \textbf{Time (Policy/Path)} \\ \hline

\textbf{GS} & 0.99 $\pm$ 0.37 & 88.7 $\pm$ 26.5 & 55.4 $\pm$ 19.5 & \textbf{0.001/0.18} & 8.73 $\pm$ 2.16 & 246.3 $\pm$ 82.8 & 58.9 $\pm$ 24.6 & \textbf{0.001/0.18} \\ 

\textbf{OC} & 0.91 $\pm$ 0.34 & 74.7$\pm$ 20.6 & 48.2$\pm$ 15.9 & 4.43/4.43 & 8.29 $\pm$ 2.34 & 201.6 $\pm$ 68.3 & 52.0 $\pm$ 21.2 & 6.14/6.14 \\ 
\textbf{RFM} & \textbf{0.80 $\pm$ 0.22} & 55.8 $\pm$ 14.6 & 38.9 $\pm$ 9.7 & 0.58/0.58 & - & - & - & - \\ 
\textbf{NES+QP (Vel)} & 0.81 $\pm$ 0.22 & \textbf{51.6 $\pm$ 11.1} & \textbf{34.2 $\pm$ 7.5} & 0.013/0.27 & \textbf{7.69 $\pm$ 2.07} & 128.9 $\pm$ 32.0 & 30.6 $\pm$ 11.6 & 0.016/0.31 \\ 
\textbf{NES+QP (Tau)} & 0.81 $\pm$ 0.22 & 53.9 $\pm$ 13.0 & 35.6 $\pm$ 8.1 & 0.021/0.40 & 7.74 $\pm$ 2.10 & \textbf{118.0 $\pm$ 29.3} & \textbf{30.5 $\pm$ 11.4} & 0.025/0.51 \\ \hline
\end{tabular}
}
\arrayrulecolor{black} 
\label{tab:results}
\end{table*}

\subsubsection{Conditioned on Task-space Boundaries}
We further evaluate the performance of our proposed C-NES method in mapping joint space boundaries to task space (\textbf{Q5}). For the planar robot, we consider only the end-effector position. For the Franka robot, NES is extended to handle position and orientation separately, with:

\begin{equation}
\begin{aligned}
    \label{eq: nn_ik}
    U^{\text{pos}}_{\bm{\theta}}(\bm{x}_s, \bm{q}_e) &= \lVert f^{\text{pos}}(\bm{q}_e) - \bm{x}^{\text{pos}}_s \rVert \; \sigma \big( u^{\text{pos}}_{\bm{\theta}}(\bm{x}_s, \bm{q}_e) \big), \\
    U^{\text{ori}}_{\bm{\theta}}(\bm{x}_s, \bm{q}_e) &= \arccos\big( f^{\text{ori}}(\bm{q}_e)^\trsp \bm{x}^{\text{ori}}_s \big) \; \sigma \big( u^{\text{ori}}_{\bm{\theta}}(\bm{x}_s, \bm{q}_e) \big),
\end{aligned}
\end{equation}
where \( \lVert f^{\text{pos}}(\bm{q}_e) - \bm{x}^{\text{pos}}_s \rVert \) denotes the Euclidean distance in \( \mathbb{R}^3 \), and \( \arccos\big( f^{\text{ori}}(\bm{q}_e)^\trsp \bm{x}^{\text{ori}}_s \big) \) corresponds to the geodesic distance on the \( \mathcal{S}^3 \) manifold. Here, \( f^{\text{ori}}(\bm{q}_e) \) and \( \bm{x}^{\text{ori}}_s \) are the unit quaternions representing the orientation. The total loss is defined as the sum of the losses and the gradient flow is computed as a linear combination of gradients from \( U^{\text{pos}}_{\bm{\theta}} \) and \( U^{\text{ori}}_{\bm{\theta}} \), ensuring that both position and orientation contribute to the result.

A qualitative visualization of the planar robot is shown in Figure~\ref{fig:nes_ik}. In this example, the target position of the end-effector is set to $\bm{x}_s = (2.0,2.0)$, as shown in (a). This inverse kinematics problem has two possible solutions for the joint angles: $q_1=0,q_2 =1.57$ and $q_1=1.57,q_2 =-1.57$. Rather than explicitly solving the inverse kinematics problem, our approach generates the geodesic flow that iteratively guides the solution towards the minimal geodesic length, as shown in (a) and (b). In (c), we visualize four different robot trajectories that all reach the same end-effector position but originate from different joint configurations. Due to the system's redundancy, multiple joint configurations can map to the same task space position, resulting in different geodesic flows starting from various joint angles. Consequently, the orange trajectory represents an optimal energy path that terminates at a different joint configuration, distinguishing it from the other three.

Quantitative results are presented in Table~\ref{tab:results_IK}. We compare our C-NES method against two baseline methods. The first is a geodesic shooting approach guided by Gauss--Newton (GN), a second-order optimization method commonly used in inverse kinematics. GN optimizes joint configurations for the inverse kinematics problem. The second baseline is an optimal control (OC) formulation, where the original terminal cost term $\|\bm{q}_N - \bm{q}_s\|_{\bm{Q}}^2$ in \eqref{eq:oc} is replaced with a forward kinematics cost $\|f(\bm{q}_N) - \bm{x}_s\|_{\bm{Q}}^2$. The RFM method fails under this setup due to severe grid distortions introduced by the nonlinear forward kinematics function, which make solving the eikonal equation intractable. These results highlight the effectiveness of C-NES in computing geodesic paths that map joint configurations to end-effector poses, resulting in energy-efficient motions from joint space to task space. Importantly, our formulation extends the classical eikonal equation to incorporate nonlinear forward kinematics, re-framing the inverse kinematics problem as a hybrid optimization problem that jointly balances task-space accuracy and energy optimality. The network learns a global distance field that encodes energy-efficient solutions in task space, leveraging the eikonal equation constraint $\|\nabla U\|_{\bm{G}} = 1$ to capture globally shortest paths across the configuration manifold. 

\begin{table}[h]
\centering
\caption{Baseline Comparison for Geodesic Lengths in Dynamics-aware Motion Generation in Task Space}
\label{tab:results_IK}
\resizebox{\linewidth}{!}{
\begin{tabular}{cccc}
\hline
Method & GN+GS & OC & C-NES \\
\hline
Planar Robot (2D) & $0.54 \pm 0.20$ & $0.50 \pm 0.18$ & \textbf{$0.43 \pm 0.12$} \\
Franka Robot (7D) & $6.92 \pm 1.71$ & $6.27 \pm 1.60$ & \textbf{$5.45 \pm 1.47$} \\
\hline
\end{tabular}
}
\end{table} 

\begin{table}[htbp]
\centering
\caption{Inference time (s) of NES across Different Batch Sizes.}
\label{tab:time_cost}
\resizebox{\linewidth}{!}{
\begin{tabular}{|c|c|c|c|c|c|c|}
\hline
\textbf{Batch Size} & 1 & 10 & \textbf{$10^2$} & \textbf{$10^3$} & \textbf{$10^4$} & \textbf{$10^5$} \\ \hline
CPU             & \textbf{0.010 }           &\textbf{ 0.012}             & \textbf{0.014  }               & 0.035                 & 0.249                 & 3.131                 \\ \hline
GPU             & 0.027            & 0.028             & 0.028                 & \textbf{0.028}                 & \textbf{0.031}                 & \textbf{0.121}                 \\ \hline
\end{tabular}
}
\end{table}

\subsubsection{Data and Computational Efficiency}
We further validate the data and computational efficiency of our neural Eikonal solver for geometry-aware motion generation (\textbf{Q1, Q4}). We analyze both training efficiency and inference time. 

Figure \ref{fig:training_efficiency} presents the training performance of our neural solver, evaluated by the number of data points required and total training time. For the planar robot, the network converges within 5 minutes using up to $3.0 \times 10^8$ data points. In contrast, training the Franka robot model takes approximately 4 hours and requires $5.0 \times 10^9$ data points. In both robot experiments, training time scales linearly with the size of the training dataset.

Table \ref{tab:time_cost} summarizes the inference time per forward pass across varying batch sizes. The reported times include neural network evaluation, inertial mass matrix, computation of the inertial mass matrix and potential energy, and automatic differentiation to estimate the geodesic flow. Overall, inference times remain low and stable across both CPU and GPU platforms, even for batch sizes up to $10^3$, indicating strong suitability for real-time applications. The GPU’s parallelism further enhances efficiency at larger batch sizes.
The majority of computational bottlenecks lies in inverting the matrices to retrieve geodesic flows. Notably, for a batch size of 1, CPU inference completes in under 0.01 seconds, supporting high-frequency updates of source-to-goal pairs that are critical for reactive and real-time planning and control.

\begin{table}[t]
    \begin{center}
        \begin{tabular}{cc}
            \centering
            \includegraphics[width =0.48\linewidth
            ]{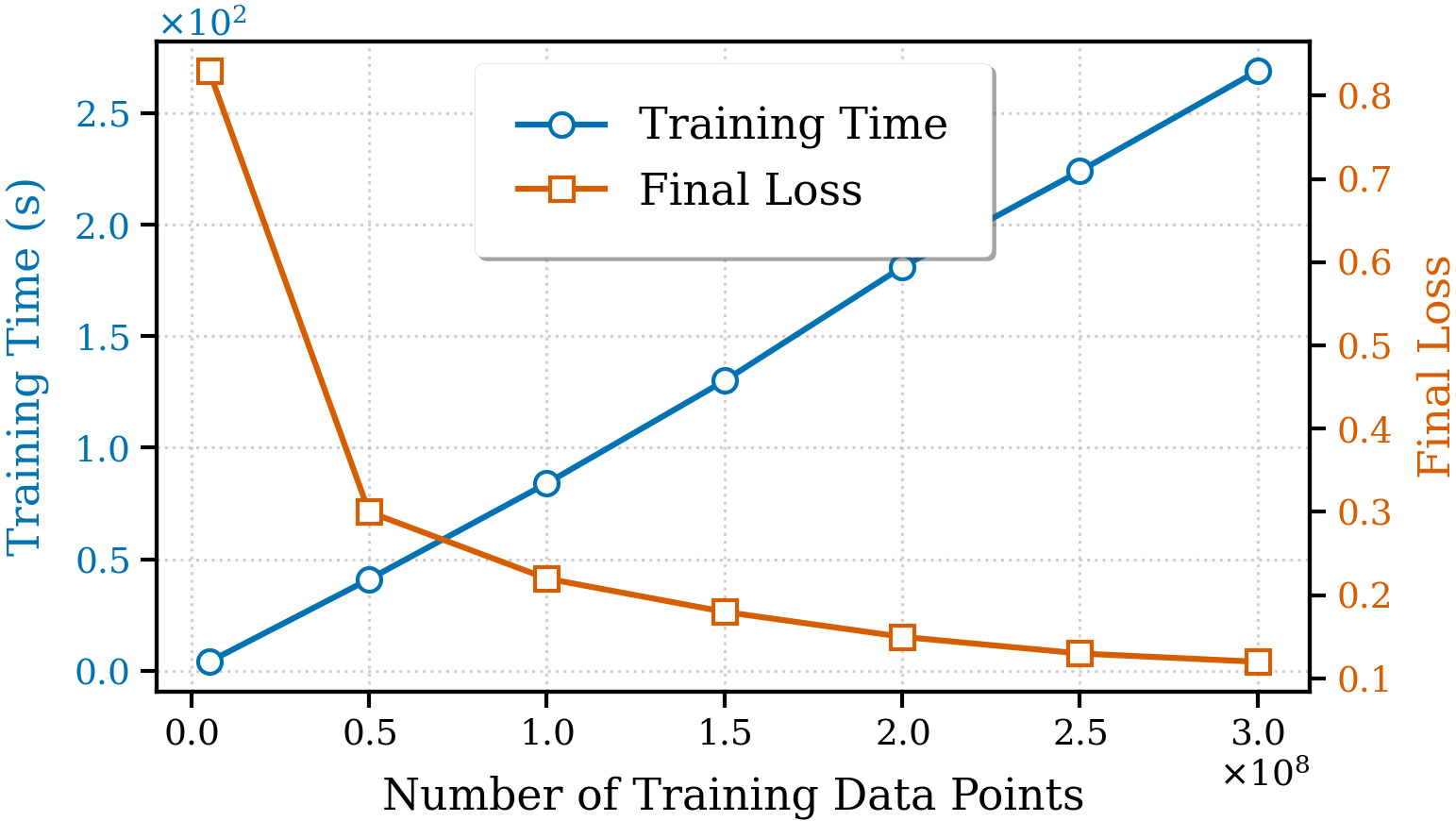}   &
            \includegraphics[width =0.48\linewidth]{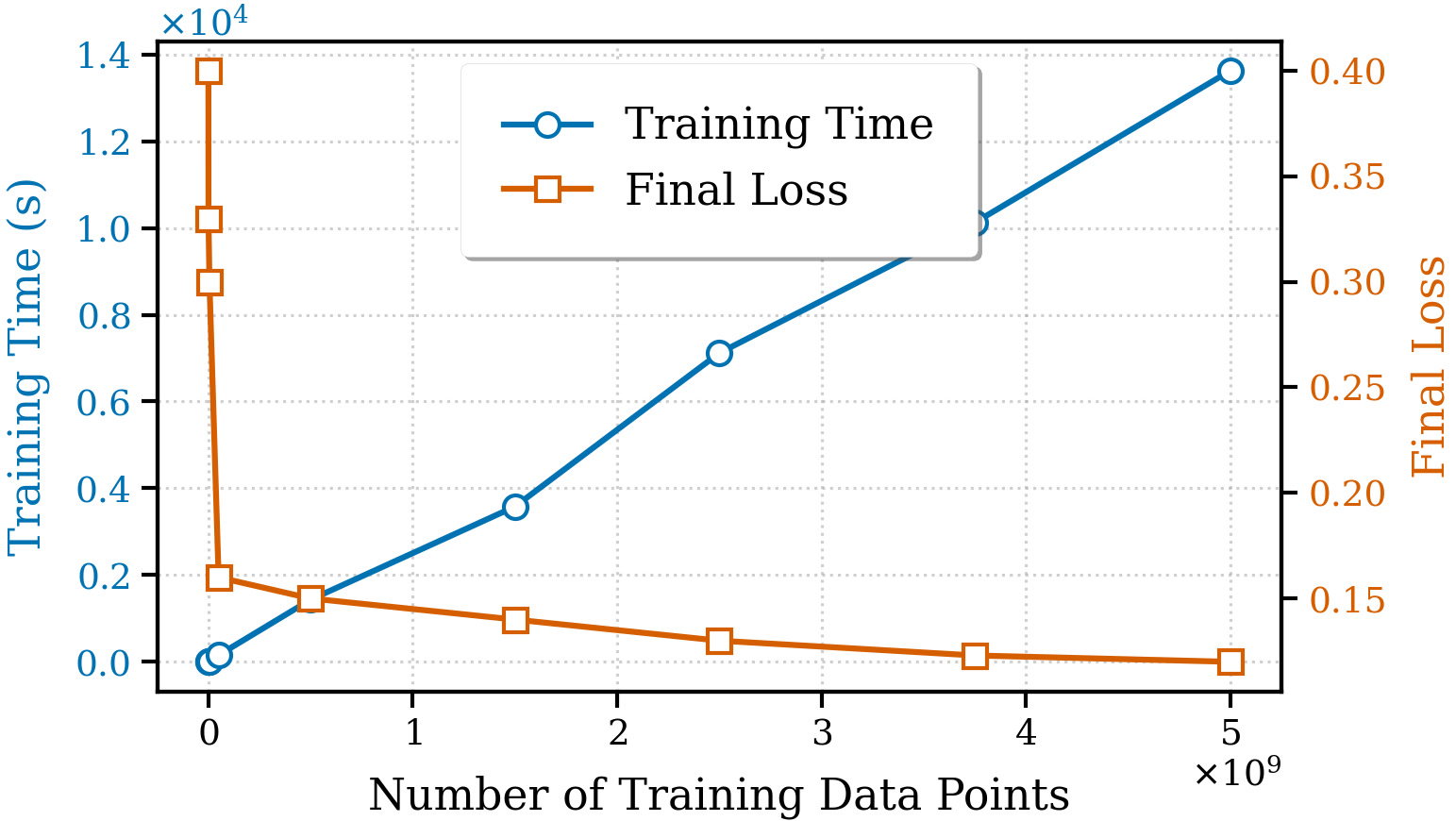}  \\
            (a) Planar Robot & (b) Franka Robot
        \end{tabular}
        \captionof{figure}{\textcolor{black}{Evaluation of efficiency for the 2D planar robot and Franka robot.}}
        \label{fig:training_efficiency}
    \end{center}
\end{table}

\subsubsection{Task-prioritized Energy-efficient Motion Policy}
Previous experiments have demonstrated NES's capability to generate energy-efficient motions that can be effectively tracked using a QP controller. In this section, we further evaluate the effectiveness of NES within task-prioritized control frameworks, focusing on its ability to preserve natural energy conservation behavior while satisfying additional constraints (\textbf{Q6}). Specifically, we consider two representative scenarios: constrained motion and obstacle avoidance. The constrained motion scenario involves task-space requirements, such as maintaining a desired end-effector position and orientation. In contrast, obstacle avoidance requires generating collision-free trajectories to ensure safe operation in cluttered environments. The experimental setups for both scenarios are described in detail below:

\subsubsection{Constrained motion} In this setup, the robot’s end-effector is restricted to moving within a specified plane while maintaining an orthogonal orientation. The constraint requires the end-effector to remain in a plane parallel to the horizontal plane at $x=0.3$, with the vertical axis of the end-effector aligned consistently throughout the motion. To achieve energy-efficient movement, our NES policy is designed to operate within the null-space of the primary task. The Jacobian associated with this constraint defines a subspace of the robot’s full Jacobian, allowing NES to optimize energy usage without violating the primary task requirements.
 
\subsubsection{Obstacle avoidance} This scenario focuses on achieving energy-efficient motion while preventing collisions with surrounding obstacles. Building on prior work~\citep{li2024representing}, we utilize a distance function and its gradient to monitor proximity to obstacles. When no obstacles are within the predefined safety margin, the robot follows the NES-generated energy-efficient policy. However, if an obstacle enters a predefined safety zone, the collision avoidance strategy constrains NES within the null-space, ensuring safe, collision-free trajectories without compromising the primary task.

\begin{table}[h]
    \begin{center}
        \begin{tabular}{cc}
            \centering
            \includegraphics[width =0.45\linewidth]{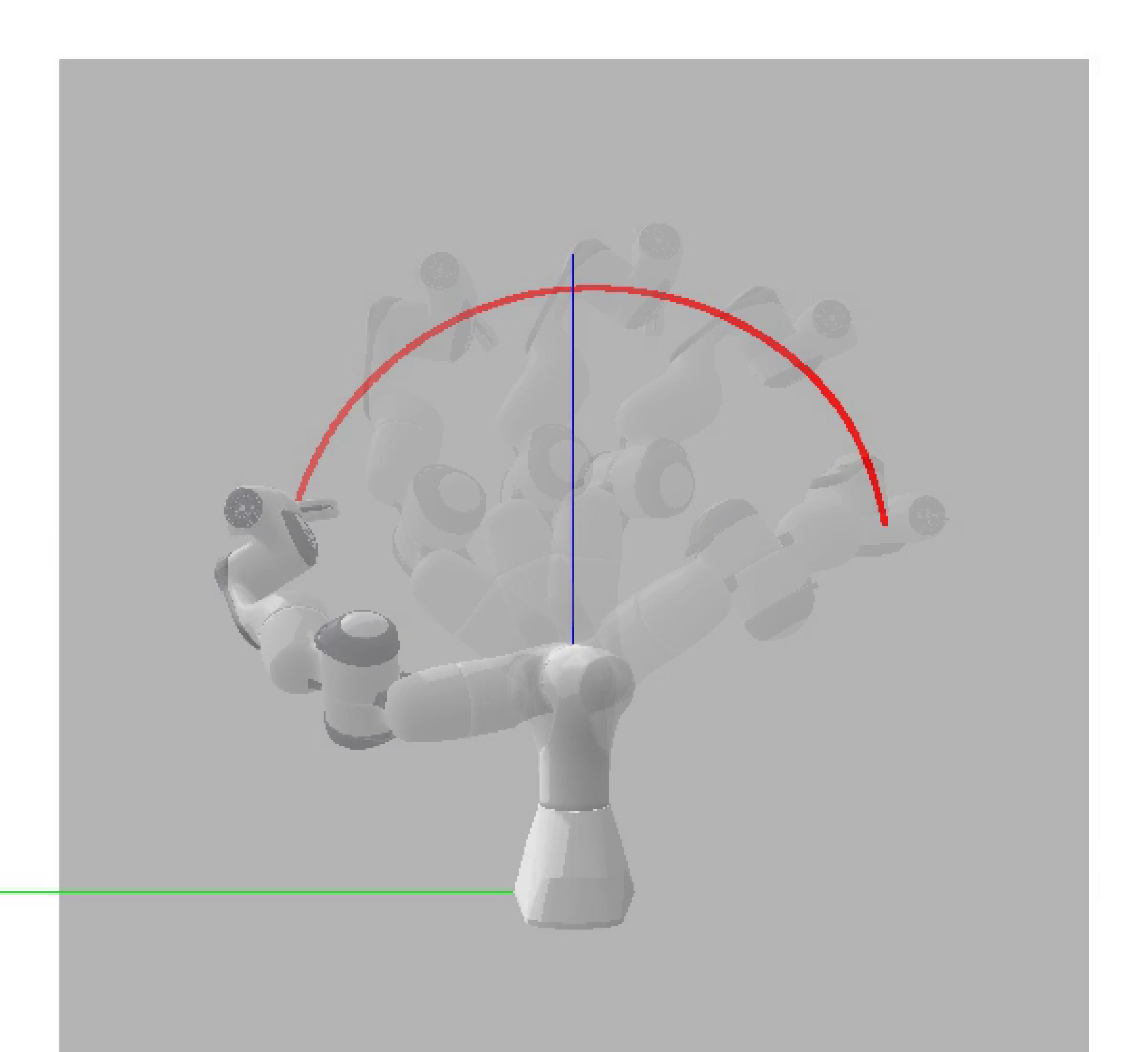}   &
            \includegraphics[width =0.45\linewidth]{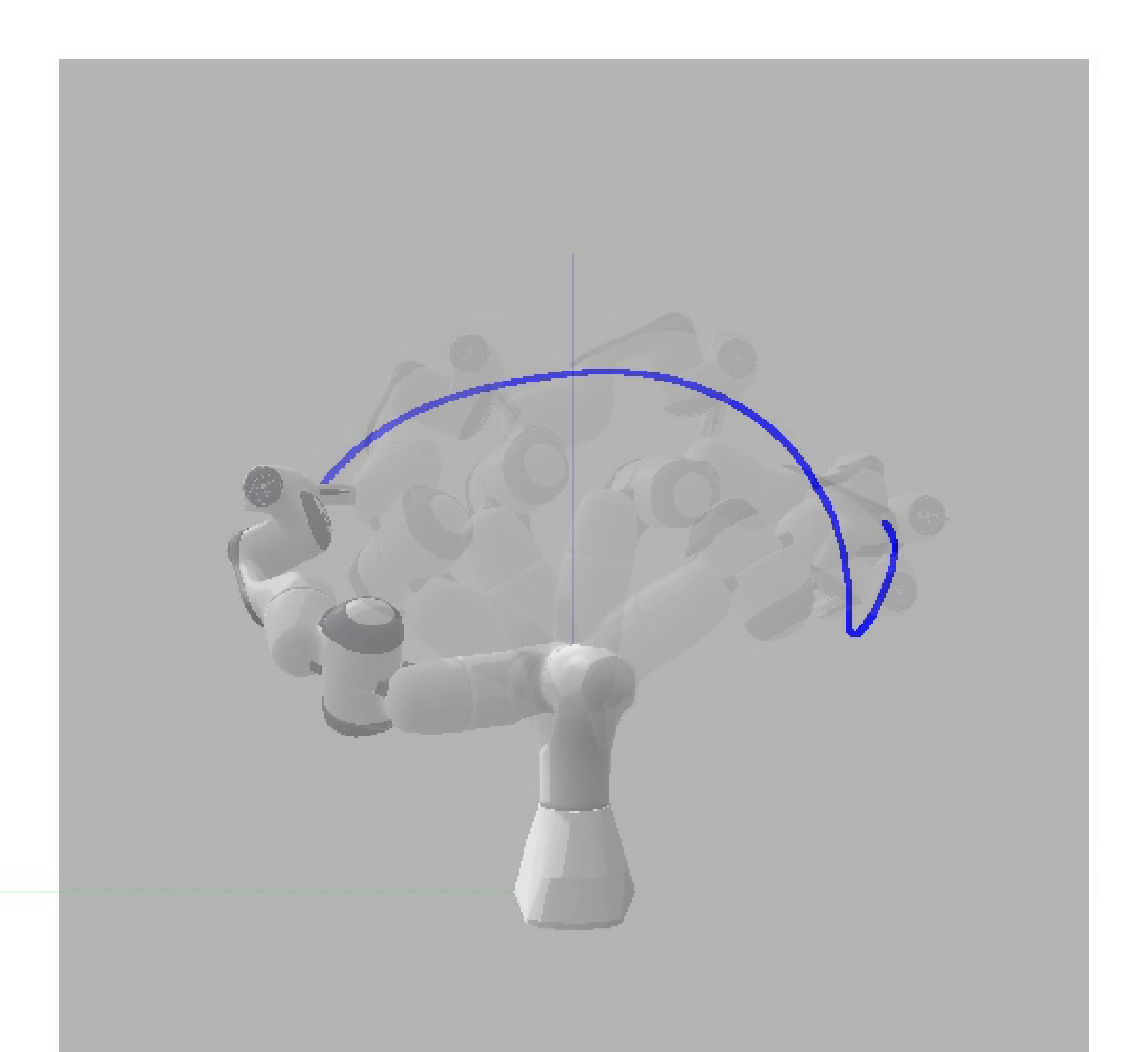}  \\
            \includegraphics[width =0.45\linewidth
            ]{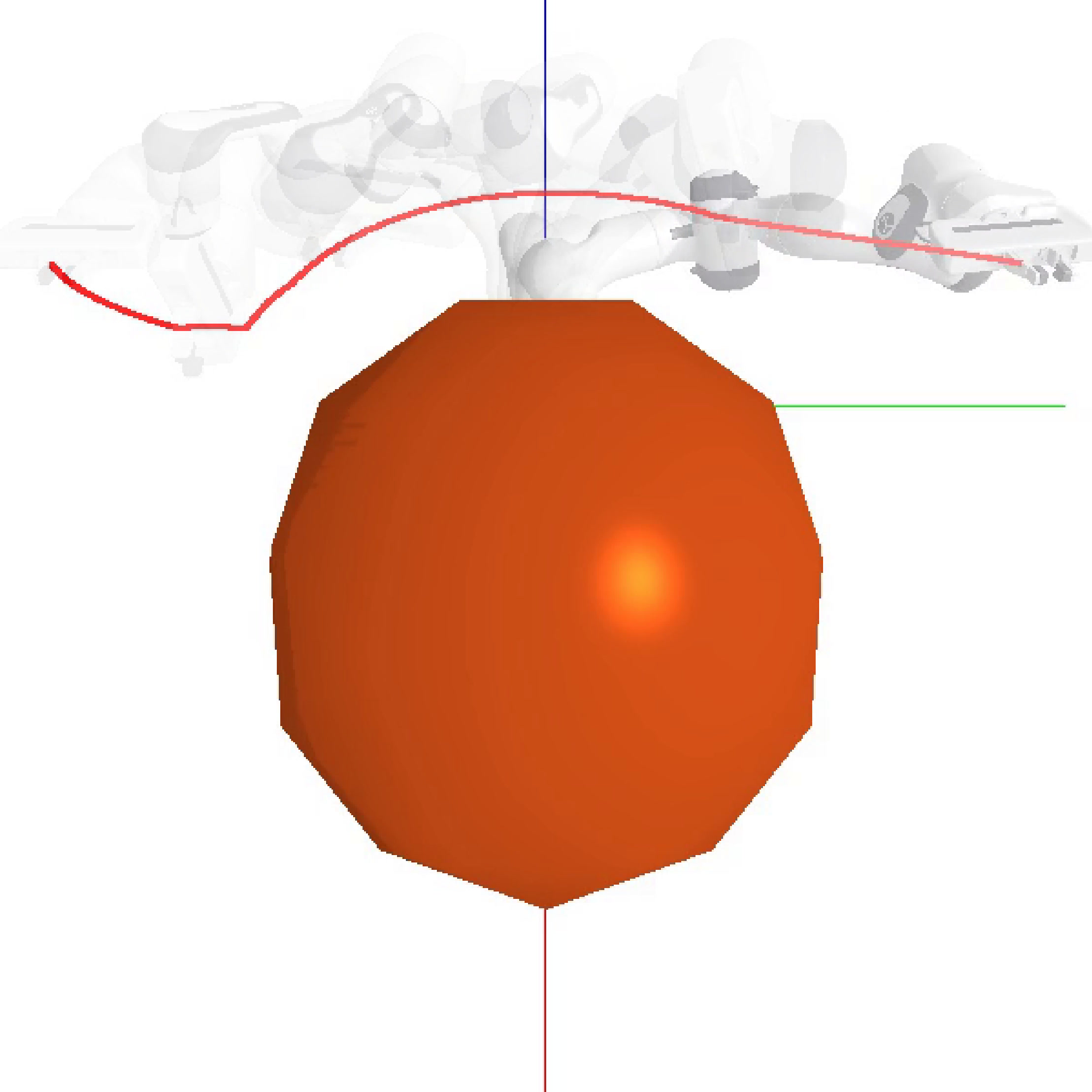}   &
            \includegraphics[width =0.45\linewidth]{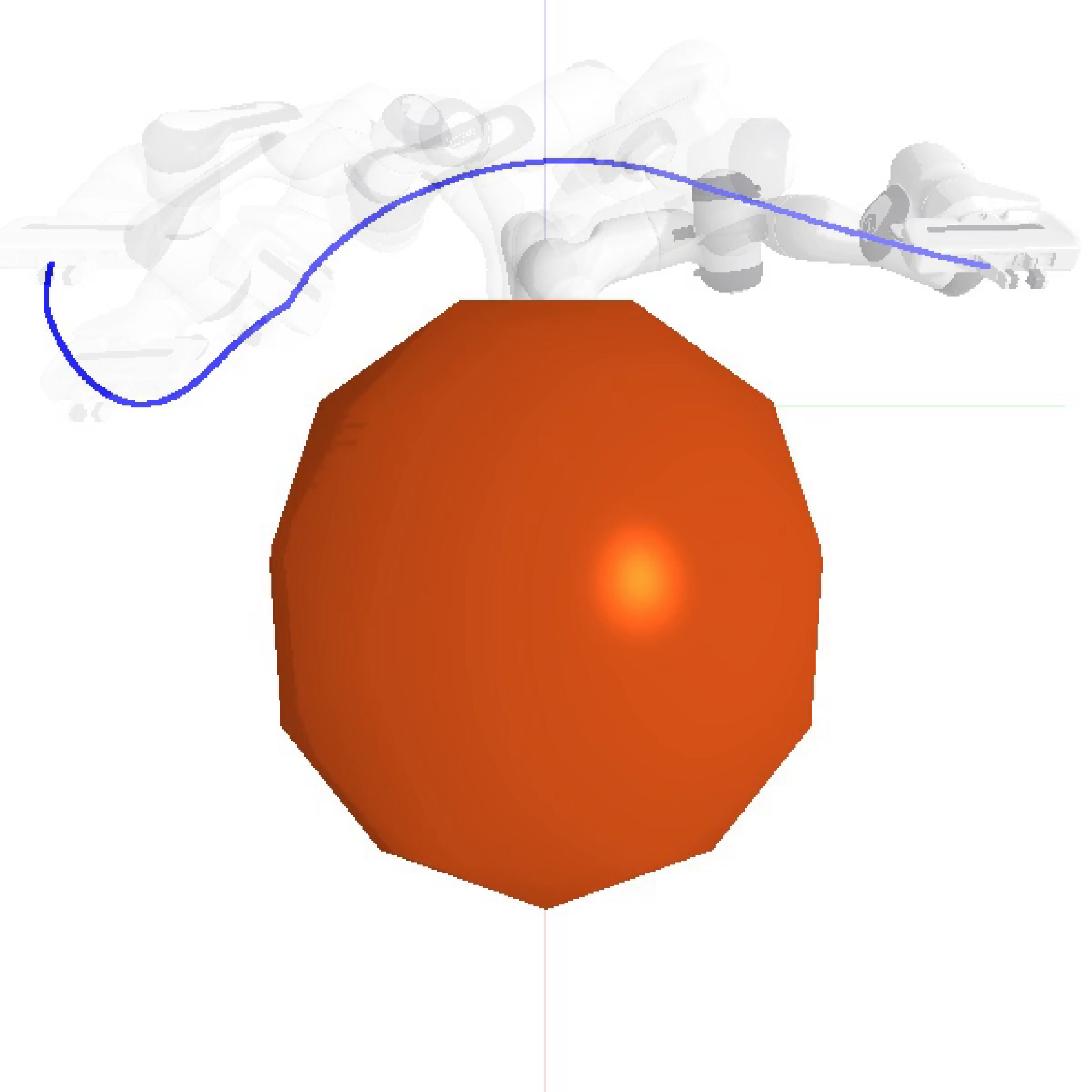}  \\
        \end{tabular}
        \captionof{figure}{Trajectories generated by task-prioritized energy-efficient motion policy and baseline. Top: Constrained motion in which the end-effector is perpendicular to a plane. Bottom: Obstacle avoidance. Red and blue curves depict the GS and NES paths operating in the null-space of each principal task.}
        \label{fig:tp-NES}
    \end{center}
\end{table}

We begin by randomly sampling 100 joint configuration pairs that satisfy the above two constraints. For each pair, we apply the NES algorithm to compute energy-aware policies. We use geodesic shooting as the baseline because it is also a reactive approach. Both NES and geodesic shooting policies are projected into the null-space of the primary tasks. Table \ref{table:tp-nes} presents the experimental results on average geodesic lengths for each method. As expected, introducing constraints will increase geodesic lengths on the configuration space manifold as the space of feasible solutions is restricted. Despite this, the NES-generated motion policies continue to outperform geodesic shooting in terms of energy efficiency. Figure~\ref{fig:tp-NES} compares trajectories of task-prioritized motions with and without our energy-aware motion policy, demonstrating the effectiveness and flexibility of NES in combination with other motion policies for real-time, adaptive, and energy-efficient robot control. 

\begin{table}[]
\centering
\caption{Geodesic lengths for task-prioritized energy-aware motion policies under the Jacobi metric.}
\label{table:tp-nes}
\resizebox{\linewidth}{!}{
\begin{tabular}{|c|c|c|c|}
\hline
Method           & No Constraint           & Obstacle                & Constrained Motion        \\ \hline
GS + TP   & 7.94$\pm$1.67           & 8.52$\pm$1.83           & 8.75$\pm$1.83             \\ \hline
NES + TP      & \textbf{7.67$\pm$1.70}           & \textbf{7.86$\pm$1.85}           & \textbf{8.20$\pm$1.86}             \\ \hline

\end{tabular}
}
\end{table}

\subsubsection{Robot Experiments}

Finally, we conduct robotic experiments to demonstrate the efficacy of our approach; refer to the accompanying video (\textbf{Q4}). In the experiments, we employ the QP controller with joint velocity inputs to track the geodesic flow generated by the NES framework. Two scenarios are evaluated for energy-conserving paths: with and without gravity compensation. In the first scenario, gravity compensation is applied externally via the controller, so the Riemannian metric reflects only \textit{kinetic energy}. In the second, both kinetic and potential energy are incorporated into the geodesic computation, resulting in full energy conservation (Jacobi metric). 

\begin{figure}
    \centering
    \includegraphics[width=1.0\linewidth]{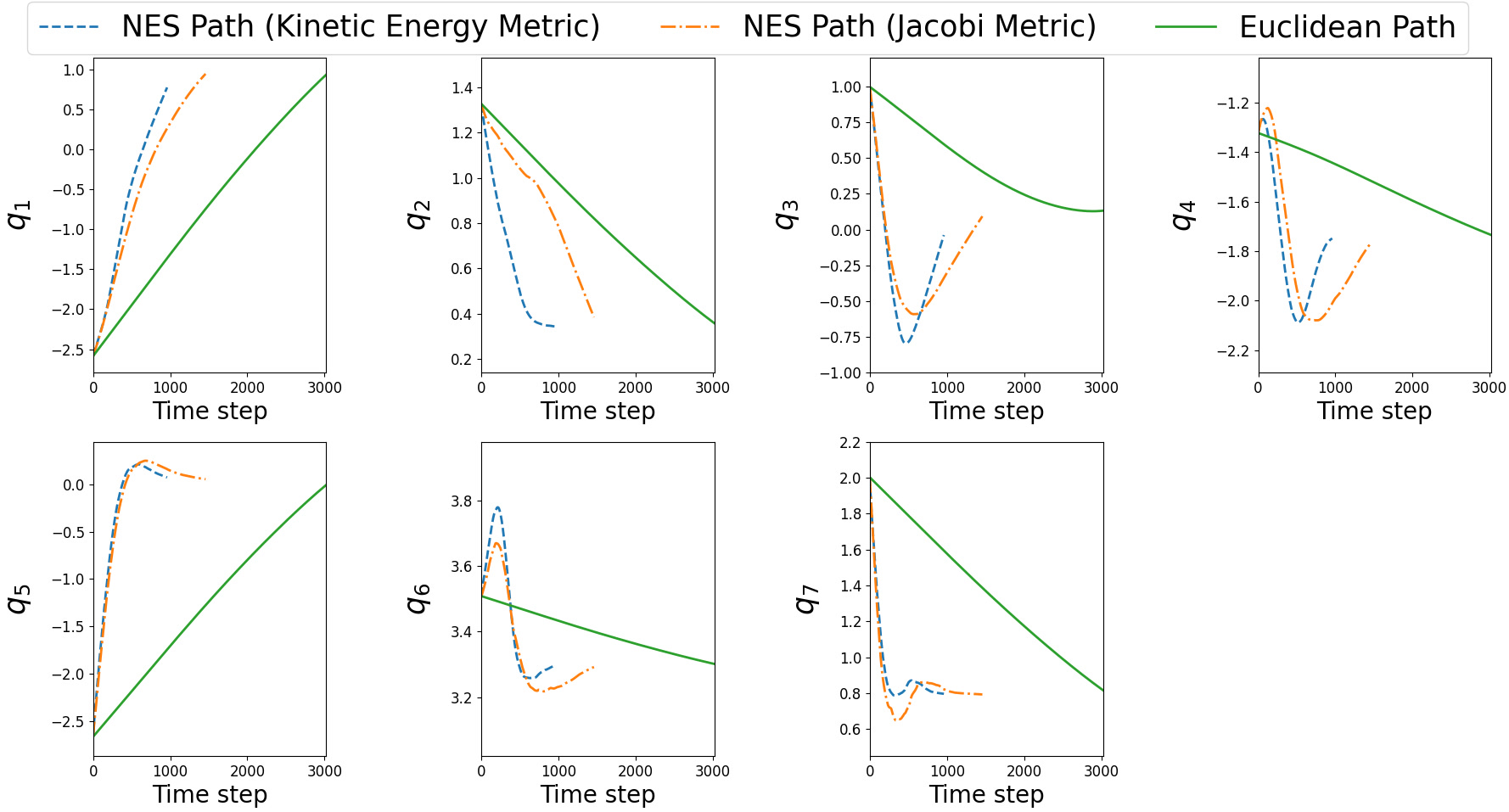}
    \caption{Joint positions for Euclidean and NES-generated paths. }
    \label{fig:nes_path_7d}
\end{figure}

\begin{figure}
    \centering
    \includegraphics[width=\linewidth,trim=2pt 2pt 2pt 12pt,clip]{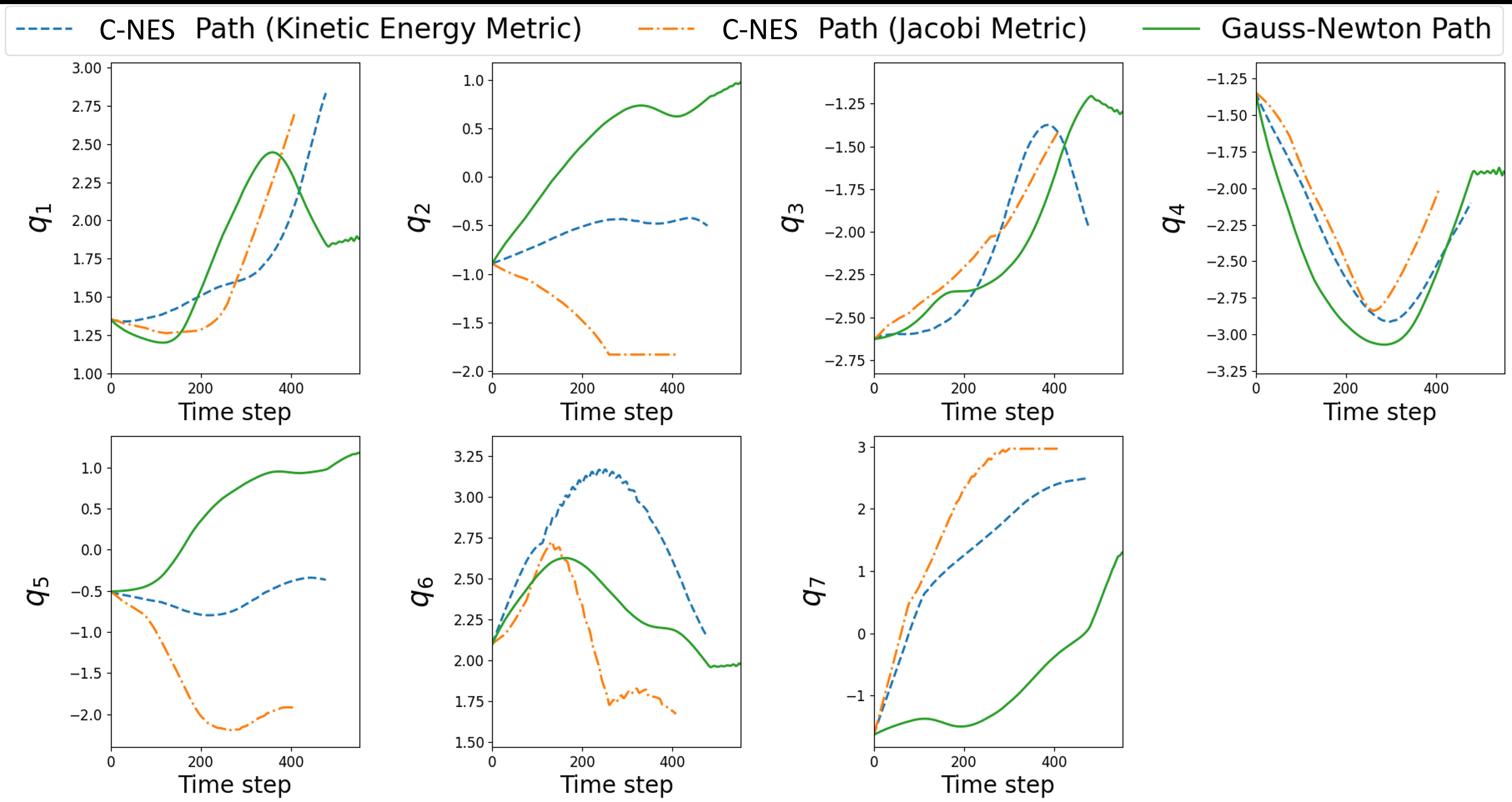}
    \caption{Joint positions for inverse kinematics solved by Gauss-Newton and C-NES with kinetic energy metric and Jacobi metric. }
    \label{fig:nes_path_7d_ik}
\end{figure}

Figure~\ref{fig:config_planner} presents key frames of the robot trajectories generated by NES under both metrics. For comparison, we include the results from the geodesic shooting (GS) approach, where the initial velocity is directed toward the goal point and scaled to satisfy the manifold constraints. On both manifolds, GS produces a straight-line Euclidean path with differences lying in joint velocities to satisfy the underlying Riemannian metric constraint. To visualize differences more clearly, Figure~\ref{fig:nes_path_7d} plots the joint positions for all three trajectories. Although all trajectories share the same start and goal configurations, the NES-generated paths exhibit more natural behavior with curved trajectories in configuration space, better reflecting energy-efficient motion consistent with the geometry of the underlying manifold. Similarly, robot motions and joint positions for task-space motion generation are visualized in Figures \ref{fig:IK_planner} and \ref{fig:nes_path_7d_ik}. Although the target joint configuration is unknown, the gradient flow produced by our C-NES allows the robot to reach the solution iteratively along the path of minimal geodesic distance. Due to the redundancy of the robot, the resulting trajectories differ and converge to distinct final joint configurations, all satisfying the same task-space goal. 


\begin{table}[t]
    \centering
    \begin{tabular}{c}
        \includegraphics[width=0.9\linewidth]{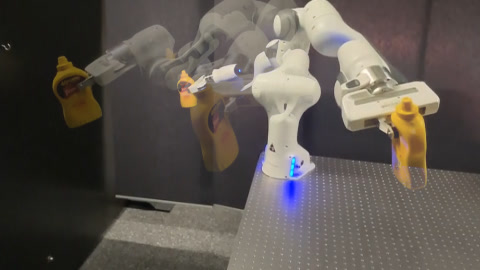} \\
        (a) NES-generated path with kinetic energy metric. \\
        \includegraphics[width=0.9\linewidth]{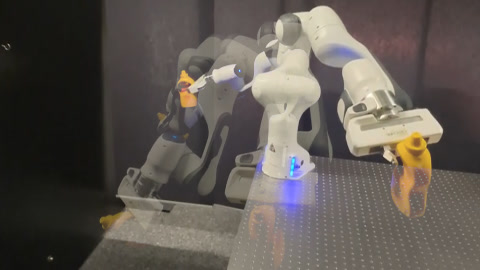} \\
        (b) NES-generated path with Jacobi metric. \\
        \includegraphics[width=0.9\linewidth]{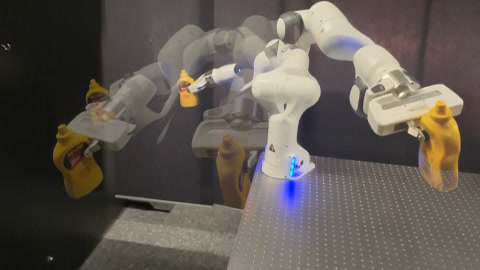} \\
        (c) Euclidean path. \\
    \end{tabular}
   \captionof{figure}{Snapshots of robot motions for C-space path planning. Initial and final frames are displayed in solid color. Intermediary frames are transparent.}
    \label{fig:config_planner}
\end{table}

We further visualize the variations in energy and torque during the robot’s motion to assess the energy efficiency of our approach. Figure~\ref{fig:energy_torque_table} (a) shows the energy and torque variations on the Riemannian manifold with Jacobi metric (\textit{solid lines}), compared to the Euclidean path (\textit{dashed lines}). On the manifold equipped with the Jacobi metric, the total energy—comprising both kinetic and potential components—is expected to remain constant. Rather than directly moving toward the goal, the NES-generated path initially reduces the robot's potential energy, converting it into kinetic energy and resulting in a curved path. Therefore, we observe lower energy and reduced torque consumption. Similarly, Figure~\ref{fig:energy_torque_table} (b) illustrates kinetic energy-efficient trajectories for the task-space motion generation problem using C-NES. In this case, the kinetic energy remains constant at each time step. It exhibits more effective utilization of kinetic energy compared to the geodesic shooting path with Gauss-Newton optimization. While it may require higher torque at the beginning to maintain kinetic energy, it compensates by reducing energy and torque demands later in the motion, ultimately producing a more efficient trajectory to reach the goal.

To further evaluate performance at scale, we simulate 100 randomly sampled start-to-goal configuration pairs and record energy and torque cost during trajectory execution. Results are summarized in Table~\ref{tab:energy_torque}. At each time step, joint positions and velocities are used to compute the energy, while the total torque is obtained by summing the actuator torques applied throughout the trajectory execution. Across both Riemannian metrics, NES consistently achieves lower overall energy cost and reduced torque inputs compared to GS. Its variant, C-NES, also outperforms GS with Gauss-Newton optimization in task space. These results highlight the strength of the NES framework in producing dynamics-aware, energy-efficient motions that are well-suited for real-world control and planning tasks.

\begin{table}[t]
    \centering
    \begin{tabular}{c}
        \includegraphics[width=0.8\linewidth]{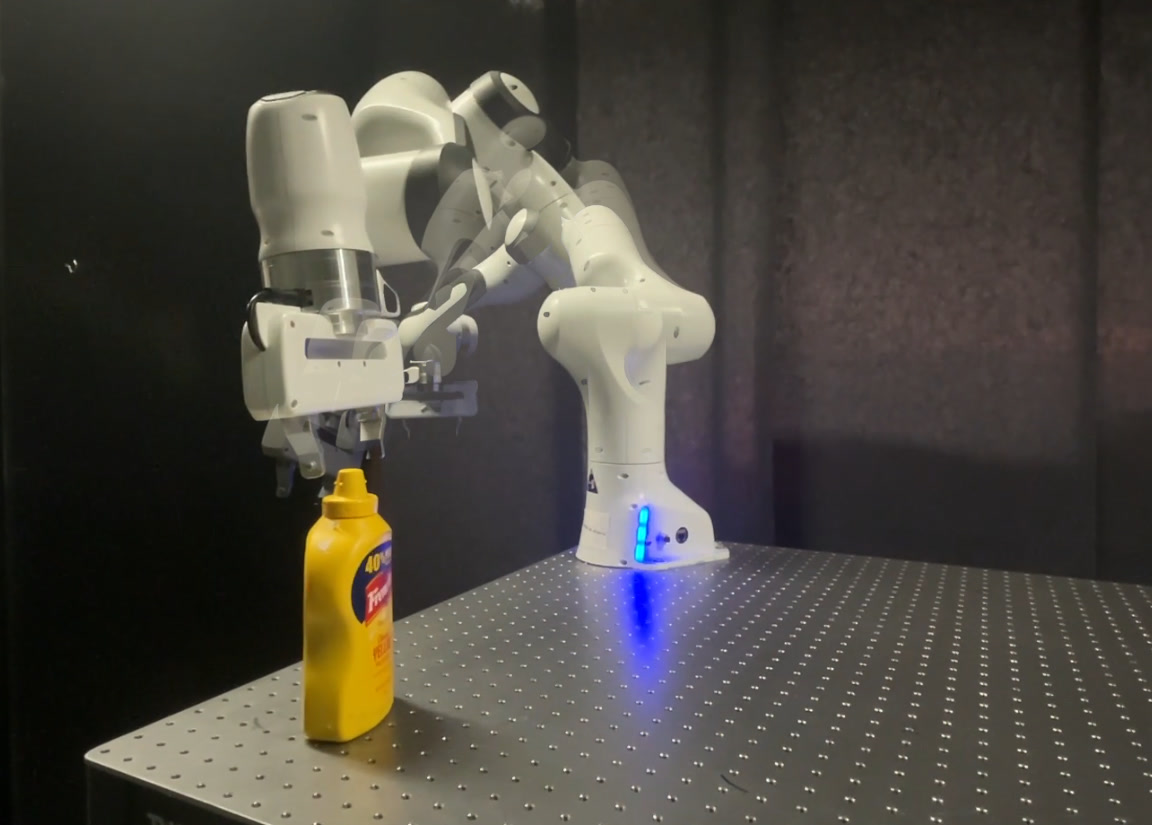} \\
        (a) The solution of C-NES with the kinetic energy metric. \\
        \includegraphics[width=0.8\linewidth]{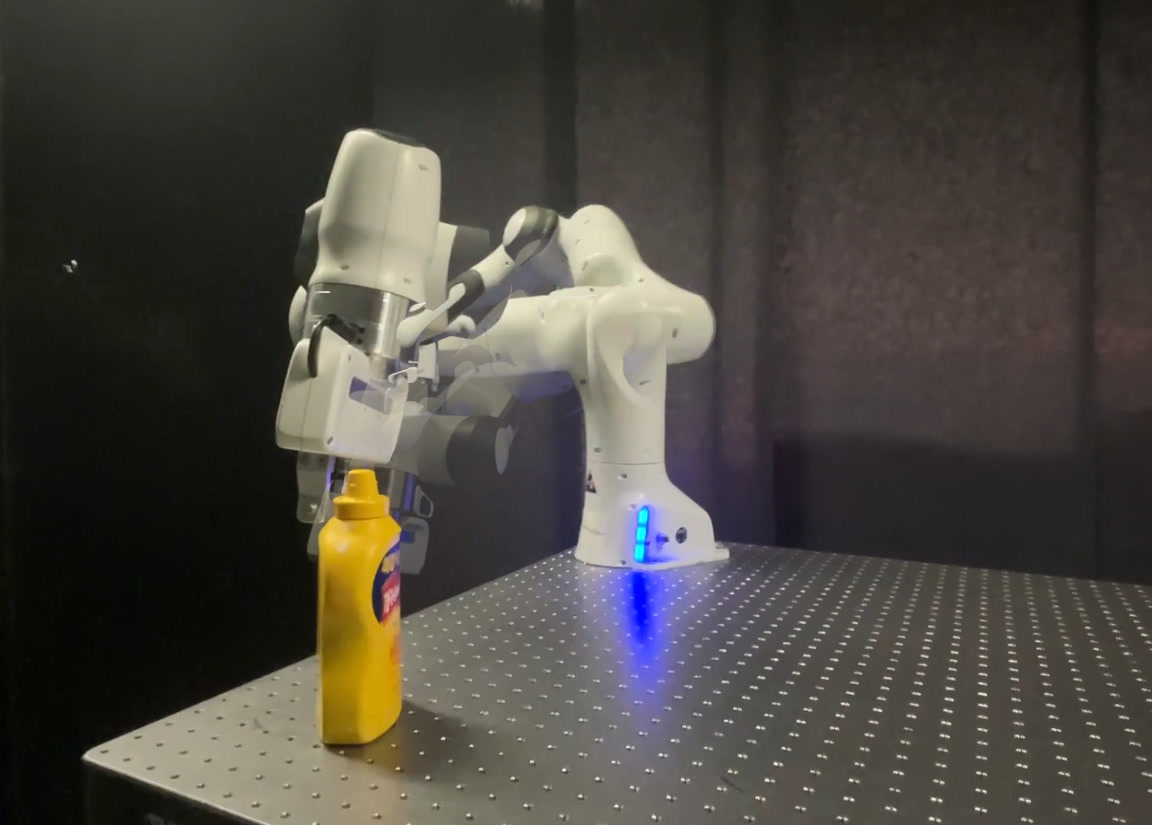} \\
        (b) The solution of C-NES with the Jacobi metric. \\
        \includegraphics[width=0.8\linewidth]{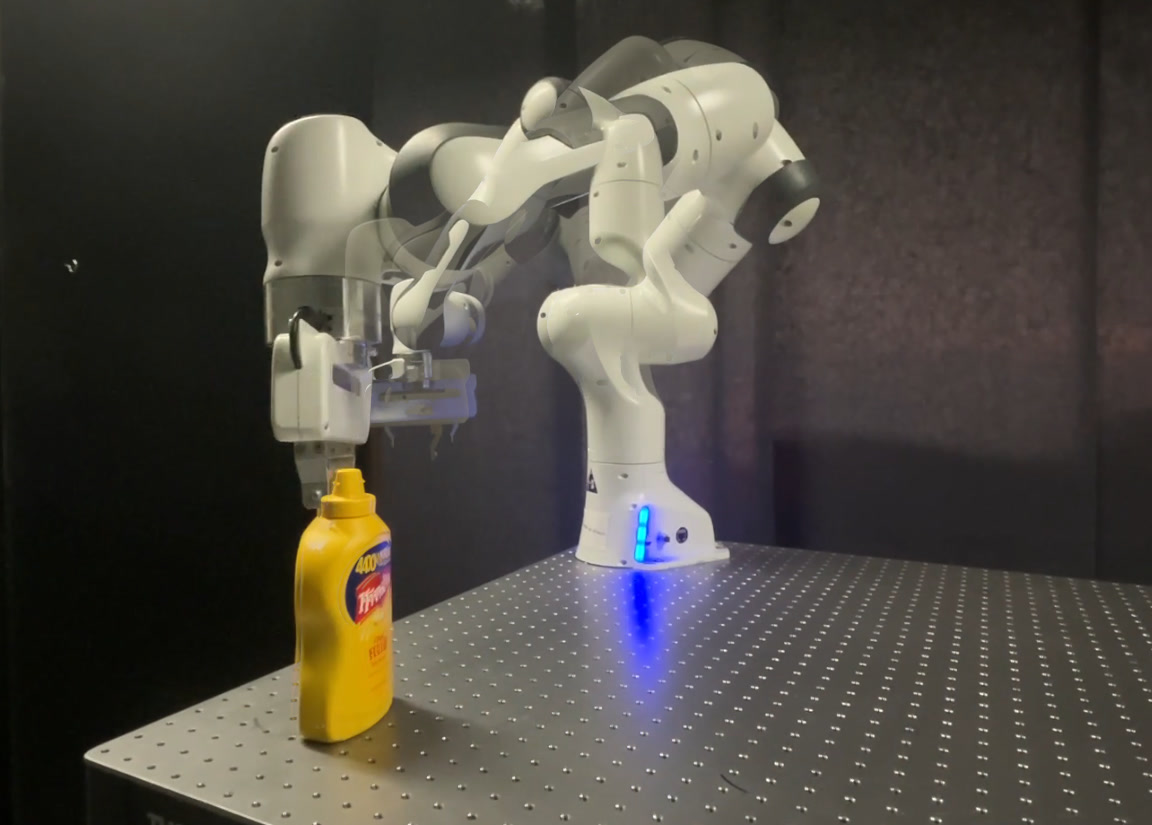} \\
        (c) The solution of the Gauss-Newton method. \\
    \end{tabular}
    \captionof{figure}{Snapshots of robot motions for inverse kinematics. Initial and final frames are displayed in solid color. Intermediary frames are transparent.}
    \label{fig:IK_planner}
\end{table}

\begin{table}[htbp]
    \centering
    \begin{tabular}{c}
        \includegraphics[width=0.9\linewidth]{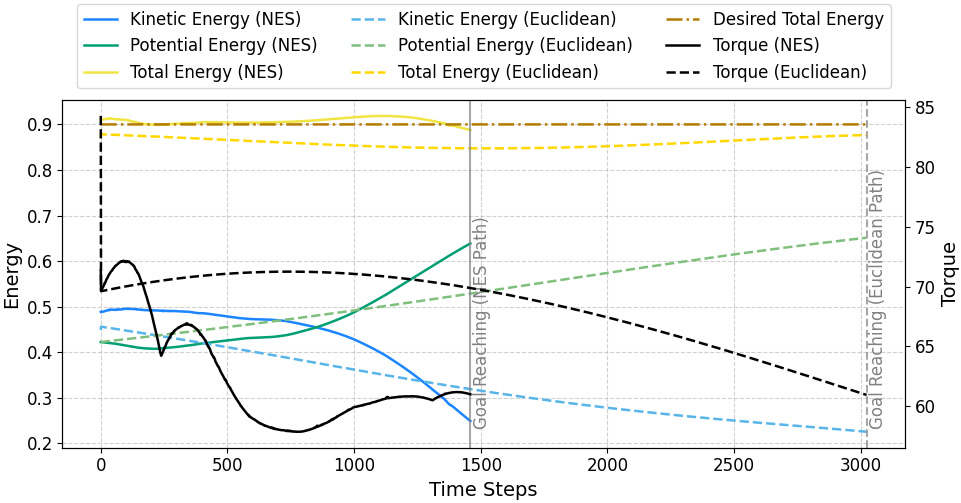} \\[2mm]
        (a) NES and Euclidean paths with Jacobi metric. \\[4mm]
        \includegraphics[width=0.9\linewidth,trim=2pt 2pt 2pt 15pt,clip]{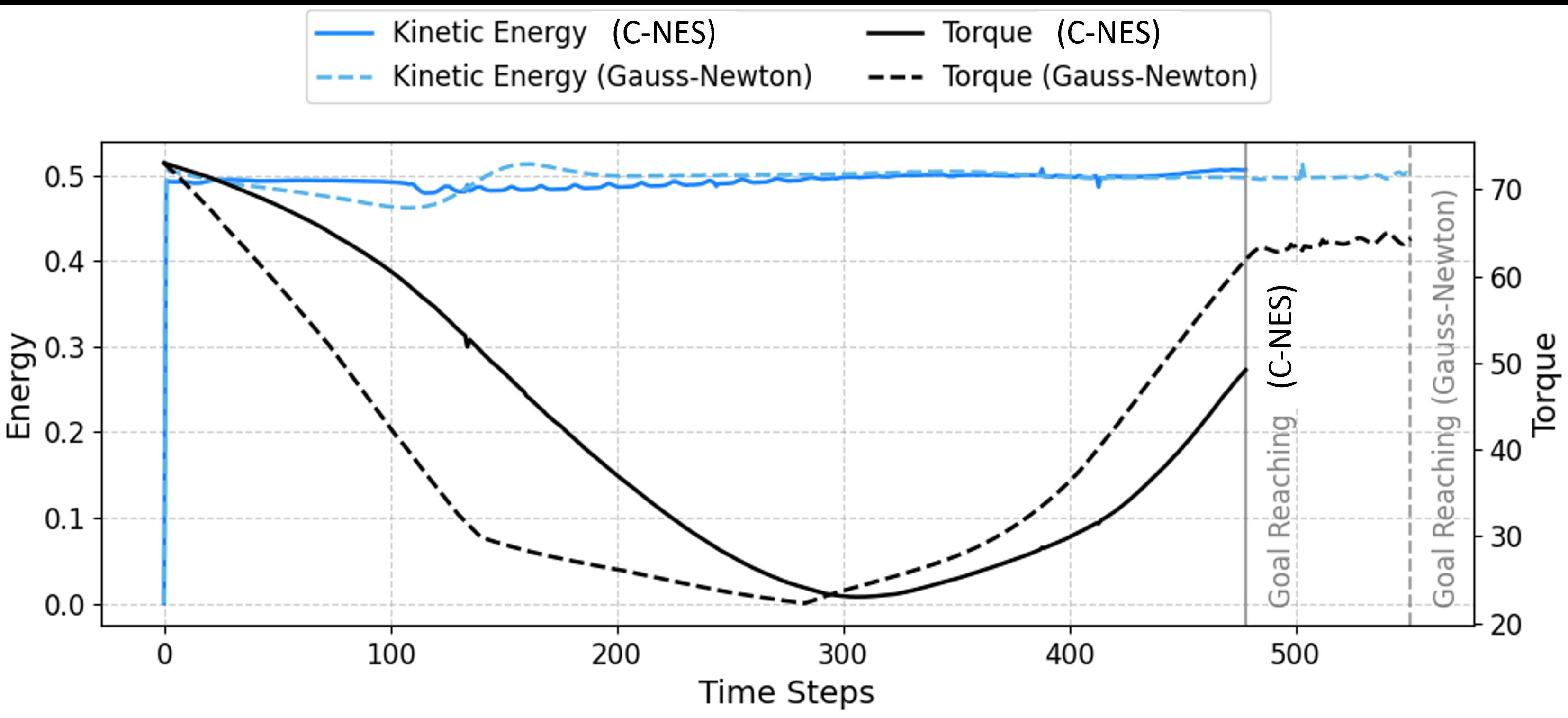} \\[2mm]
        (b) C-NES and Gauss-Newton paths with \\ kinetic energy metric. \\
    \end{tabular}
    \captionof{figure}{Variation of energy and torque among different paths and Riemannian metrics.}
    \label{fig:energy_torque_table}
\end{table}

\begin{table}[]
    \centering
    \caption{Total Energy and torque cost during the robot's motion.}
    \label{tab:energy_torque}
    \resizebox{\linewidth}{!}{
\begin{tabular}{|cc|cc|cc|}
\hline
\multicolumn{2}{|c|}{\multirow{2}{*}{}}                                         & \multicolumn{2}{c|}{Kinetic Energy Metric}             & \multicolumn{2}{c|}{Jacobi Metric}                    \\ \cline{3-6} 
\multicolumn{2}{|c|}{}                                                          & \multicolumn{1}{c|}{Energy} & Torque           & \multicolumn{1}{c|}{Energy}   & Torque          \\ \hline
\multicolumn{1}{|c|}{\multirow{2}{*}{C-Space}}          & GS & \multicolumn{1}{c|}{$1.53\pm 0.62$}  & $13.4\pm 6.89$   & \multicolumn{1}{c|}{$2.82 \pm 0.85$} & $12.8 \pm 5.17$ \\ \cline{2-6} 
\multicolumn{1}{|c|}{}                                  & NES            & \multicolumn{1}{c|}{$0.62 \pm 0.18$} & $5.14\pm 1.92$   & \multicolumn{1}{c|}{$1.42 \pm 0.46$} & $6.68 \pm 2.54$ \\ \hline \hline
\multicolumn{1}{|c|}{\multirow{2}{*}{Task-space}}              & GS + GN   & \multicolumn{1}{c|}{$0.20\pm 0.09$}  & $1.56 \pm 0.62$  & \multicolumn{1}{c|}{$0.51 \pm 0.23$} & $2.12 \pm 1.08$ \\ \cline{2-6} 
\multicolumn{1}{|c|}{}                                  & C-NES        & \multicolumn{1}{c|}{$0.15\pm 0.07$}  & $1.26 \pm 0.56$  & \multicolumn{1}{c|}{$0.37 \pm 0.11$} & $1.75 \pm 0.88$ \\ \hline
\end{tabular}
}
\end{table}
\textcolor{black}{
It is important to note that, in practical mechanical systems, actuators cannot perfectly share or recycle power. Negative mechanical work is often dissipated locally, and opposed closed-force loops can generate internal forces with near-zero net mechanical work while still incurring nontrivial electrical losses. Consequently, geodesic optimality (minimal length under the chosen Riemannian metric) does not, in general, coincide with minimal execution effort. We therefore use geodesics as a principled heuristic: a global, geometry-aware reference that can guide the robot toward lower-energy motions and inform trajectory generation. Although not universally aligned with all actuator cost models, this approach is grounded in a well-established geometric framework and yields geometrically optimal paths for analyzing and planning robot motion when both kinetic and potential energy are modeled. In addition, the total energy $H$ can be provided as a conditioning input to NES, allowing the geodesic flow to adapt dynamically to changes in energy, as described in Section~\ref{sec:nes-meg}.
}

\section{Conclusion}
\label{sec:conclusion}
\textcolor{black}{
In this article, we presented an approach to compute Riemannian distance functions by formulating the problem as wavefront propagation and solving the corresponding \emph{Riemannian eikonal equation}.} We proposed a neural Riemannian eikonal solver (NES) that learns a mesh-free, continuous, and differentiable implicit field, enabling efficient queries of distances and geodesic flows in high-dimensional configuration spaces, with globally consistent distance and flow fields. In addition, we introduced NES variants that can be conditioned on boundary data and/or on Riemannian metrics. Our method integrates with planning, control, and optimization as a principled geometric prior. We demonstrated the effectiveness of the approach in constructing distance fields and geodesic flows across kinematics, dynamics, motion planning, and control problems—including high-dimensional manipulators—with quantitative and qualitative comparisons to established baselines. Finally, we conducted extensive experiments on energy-aware manifolds induced by kinetic and potential energy. The resulting geodesics provide informative guidance toward lower-energy and lower-torque motions, highlighting the practicality of our approach.

A limitation of our approach lies in training the neural Riemannian eikonal solver. Once trained, the model can be reused across multiple queries, enabling efficient and real-time applications. However, the lack of prior data and the strongly anisotropic nature of the Riemannian metric still pose significant challenges for the training. While our formulation theoretically allows for global optimal solutions, its practical performance can in some cases unexpectedly fall short. Future work will focus on improving the robustness and accuracy of the training process to enhance overall reliability. Additionally, we plan to extend our framework to broader metric spaces beyond Riemannian manifolds. The eikonal equation itself can be generalized to other types of partial differential equations relevant to robotics, including those arising in dynamic systems modeling~\citep{lutter2018deep}, value function computation in dynamic programming, and reinforcement learning. These extensions could unlock new applications for geometry-aware learning in complex robotic environments. 

\section{Acknowledgments}
This work was supported by the China Scholarship Council (No.202204910113), by the Swiss National Science Foundation (HORACE project), and by the State Secretariat for Education, Research and Innovation in Switzerland for participation in the European Commission's Horizon Europe Program with the INTELLIMAN project (HORIZON-CL4-Digital-Emerging Grant 101070136) and the SESTOSENSO project (HORIZON-CL4-Digital-Emerging Grant 101070310). We also 
thank Dr James Hermus for insightful discussions and helpful suggestions.

\bibliographystyle{SageH}
\bibliography{references}

\clearpage

\appendix
\section*{Appendix}

\section{Robot Dynamics on Configuration Space Manifolds}\label{sec:dynamics}
\subsection{Lagrangian Mechanics}
The Lagrangian is a scalar function defined on the tangent bundle of the configuration manifold $G$, given by:
\begin{equation}
\label{eq:lagrangian}
L\left(\bm{q}, \dot{\bm{q}}\right) = T\left(\bm{q}, \dot{\bm{q}}\right) - P\left(\bm{q}\right),
\end{equation}
where $\bm{q}$ is the configuration, $\dot{\bm{q}}$ is the velocity, $T$ is the kinetic energy, and $P$ is the potential energy. The kinetic energy is typically expressed as:
\begin{equation}
    \label{eq:kinetic_energy}
    T\left(\bm{q}, \dot{\bm{q}}\right) = \frac{1}{2} \dot{\bm{q}}^{\top} \bm{M}(\bm{q}) \dot{\bm{q}},
\end{equation}
with $\bm{M}(\bm{q})$ representing the configuration-dependent inertia matrix.
The action functional $S$, defined over a trajectory $\bm{q}(t)$ between times $t_0$ and $t_1$, integrates the Lagrangian along the path:
\begin{equation}
\label{eq:action}
S\left(\bm{q}\right) = \int_{t_0}^{t_1} L\left(\bm{q}(t), \dot{\bm{q}}(t)\right) \, \mathrm{d}t.
\end{equation}
When potential energy is ignored, this functional reduces to the kinetic-energy functional, with $\bm{M}(\bm{q})$ acting as the kinetic-energy metric. According to the principle of stationary action, the physical trajectory is one that makes the action stationary, i.e., $\delta S = 0$. This leads to the Euler–Lagrange equations:
\begin{equation}
\label{eq:euler_lagrange}
\frac{\mathrm{d}}{\mathrm{d}t} \left( \frac{\partial L}{\partial \dot{\bm{q}}} \right) - \frac{\partial L}{\partial \bm{q}} = \bm{0}.
\end{equation}
Substituting the Lagrangian into the Euler–Lagrange equations yields the system’s second-order dynamics:
\begin{equation}
\label{eq:eom}
\bm{M}(\bm{q}) \ddot{\bm{q}} + \bm{C}(\bm{q}, \dot{\bm{q}})\dot{\bm{q}} + \bm{g}(\bm{q}) = \bm{0},
\end{equation}
where $\bm{C}(\bm{q}, \dot{\bm{q}})\dot{\bm{q}}$ groups the Coriolis and centrifugal terms arising from $\dot{\bm{M}}(\bm{q}) \dot{\bm{q}}$ and $\frac{\partial \bm{M}}{\partial \bm{q}}$.
$\bm{g}(\bm{q}) = \frac{\partial P}{\partial \bm{q}}$ is the generalized gravitational force.

\subsubsection{Derivation of Equation of Motion}
\label{sec:eom}
We derive the equation of motion from Equation \eqref{eq:euler_lagrange}. The partial derivative of the kinetic energy $T$ with respect to $\dot{q}_i$ is:
\begin{equation}
\frac{\partial T}{\partial \dot{q}_i} = \sum_j M_{ij} \dot{q}_j.
\end{equation}
Taking the total time derivative yields:
\begin{equation}
\label{eq:T_t}
\frac{d}{dt} \left( \frac{\partial T}{\partial \dot{q}_i} \right) = \sum_j M_{ij} \ddot{q}_j + \sum_{j,k} \frac{\partial M_{ij}}{\partial q_k} \dot{q}_k \dot{q}_j.
\end{equation}
Next, we compute the partial derivative of $T$ with respect to $q_i$:
\begin{equation}
\frac{\partial T}{\partial q_i} = \frac{1}{2} \sum_{j,k} \frac{\partial M_{jk}}{\partial q_i} \dot{q}_j \dot{q}_k.
\end{equation}
The kinetic part of the Euler–Lagrange equation is then given by:
\begin{equation}
\label{eq:kinetic_el}
\begin{aligned}
    &\frac{d}{dt} \left( \frac{\partial T}{\partial \dot{q}_i} \right) - \frac{\partial T}{\partial q_i} \\
    &= \sum_j M_{ij} \ddot{q}_j + \sum_{j,k} \left( \frac{\partial M_{ij}}{\partial q_k} - \frac{1}{2} \frac{\partial M_{jk}}{\partial q_i} \right) \dot{q}_j \dot{q}_k \\
    &= \sum_j M_{ij} \ddot{q}_j + \sum_{j,k} \left(\frac{1}{2} \frac{\partial M_{ij}}{\partial q_k} + \frac{1}{2} \frac{\partial M_{ik}}{\partial q_j} - \frac{1}{2} \frac{\partial M_{jk}}{\partial q_i} \right) \dot{q}_j \dot{q}_k.
    \end{aligned}
\end{equation}
Since the potential energy $P$ depends only on $\bm{q}$, we have:
\begin{equation}
\frac{\partial P}{\partial \dot{q}_i} = 0, \quad \frac{\partial P}{\partial q_i} = g_i.
\end{equation}
Combining the kinetic and potential terms, the full equation becomes:
\begin{equation}
\sum_j M_{ij} \ddot{q}_j + \sum_{j,k} \Gamma_{i,jk} \dot{q}_j \dot{q}_k + g_i = 0,
\end{equation}
where the Christoffel symbol $\Gamma_{i,jk}$ is defined as:
\begin{equation}
\Gamma_{i,jk} = \frac{1}{2} \left(
\frac{\partial M_{ij}}{\partial q_k} +
\frac{\partial M_{ik}}{\partial q_j} -
\frac{\partial M_{jk}}{\partial q_i}
\right).
\end{equation}
The term $\sum_{j,k} \Gamma_{i,jk} \dot{q}_j \dot{q}_k$ corresponds to the Coriolis and centrifugal forces, i.e., the $i$-th component of $\bm{C}(\bm{q}, \dot{\bm{q}}) \dot{\bm{q}}$.
In vector form, this yields the standard form of the robot dynamics \citep{siciliano2008springer}:
\begin{equation}
\bm{M}(\bm{q}) \ddot{\bm{q}} + \bm{C}(\bm{q}, \dot{\bm{q}}) \dot{\bm{q}} + \bm{g}(\bm{q}) = \bm{0}.
\end{equation}

\subsection{Hamiltonian Mechanics}
The Hamiltonian function \( H(\bm{q}, \bm{p}) \) represents the total energy of a mechanical system~\citep{lutter2023combining}. It is defined as the sum of kinetic and potential energy.
\begin{equation}
\label{eq:total_energy}
H(\bm{q}, \bm{p}) = T(\bm{q}, \dot{\bm{q}}) + P(\bm{q}) = \frac{1}{2} \bm{p}^\top \bm{M}^{-1}(\bm{q}) \bm{p} + P(\bm{q}),
\end{equation}
where \( \bm{p} \in \mathbb{R}^n \) is the generalized momentum, given by
\begin{equation}
\bm{p} = \frac{\partial L}{\partial \dot{\bm{q}}} = \bm{M}(\bm{q}) \dot{\bm{q}}.
\end{equation}
The Hamiltonian and Lagrangian are related via a Legendre transform:
\begin{equation}
H(\bm{q}, \bm{p}) = \bm{p}^\top \dot{\bm{q}} - L(\bm{q}, \dot{\bm{q}}),
\end{equation}
where \( \dot{\bm{q}} = \bm{M}^{-1}(\bm{q}) \bm{p} \). Substituting the Euler–Lagrange equations yields Hamilton's equations:
\begin{equation}
\label{eq:hamiltons_equations}
\dot{\bm{q}} = \frac{\partial H}{\partial \bm{p}}, \qquad
\dot{\bm{p}} = -\frac{\partial H}{\partial \bm{q}}.
\end{equation}
These equations describe the evolution of the system in phase space \( (\bm{q}, \bm{p}) \). In the absence of non-conservative forces, the Hamiltonian \( H \) is conserved, reflecting conservation of total mechanical energy.

\subsubsection{Geometric Formulation and Maupertuis' Principle}
When \( H(\bm{q}, \bm{p}) \) is constant along a trajectory, the action integral can be reformulated in a time-independent form. This leads to \textit{Maupertuis' Principle}~\citep{albu2022can}:
\begin{equation}
\label{eq:maupertuis}
S_{M}(\bm{q}) = \int_{\bm{q}_1}^{\bm{q}_2} \bm{p}^\top d\bm{q},
\end{equation}
which states that the true path connecting $\bm{q}_1$ and $\bm{q}_2$ corresponds to a minimum or a saddle point of the action functional. This reformulation provides a geometric viewpoint on robot dynamics, enabling the study of energy-conserving paths using tools from Riemannian geometry and algebraic topology. In this setting, constant-energy trajectories correspond to geodesics under the \emph{Jacobi metric}~\citep{casetti2000geometric}
\begin{equation}
\label{eq:jacobi_metric}
\bm{G}_\text{Jac}(\bm{q}) = 2 \big( H - P(\bm{q}) \big) \bm{M}(\bm{q}),
\end{equation}
which is a conformal scaling of the kinetic-energy metric.

\subsubsection{Derivation of Jacobi metric}
To derive the Jacobi metric, we can reparameterize the trajectory \( \bm{q}(t) \) using the arc-length \( s \), with
\begin{equation}
    \bm{q}' = \frac{d\bm{q}}{ds}, \quad \dot{\bm{q}} = \frac{d\bm{q}}{dt} = \frac{d\bm{q}}{ds} \frac{ds}{dt} = \bm{q}' \frac{ds}{dt}.
\end{equation}
Then, the kinetic energy becomes
\begin{equation}
\begin{aligned}
T &= \frac{1}{2} \dot{\bm{q}}^\top \bm{M}(\bm{q}) \dot{\bm{q}} 
= \frac{1}{2} \left( \bm{q}' \frac{ds}{dt} \right)^\top \bm{M}(\bm{q}) \left( \bm{q}' \frac{ds}{dt} \right) \\
&= \frac{1}{2} \left( \frac{ds}{dt} \right)^2 \bm{q}'^\top \bm{M}(\bm{q}) \bm{q}'.
\end{aligned}
\end{equation}
By energy conservation,
\begin{equation}
H = T + P(\bm{q}) \quad \Rightarrow \quad T = H - P(\bm{q}).
\end{equation}
Equating both expressions for \( T \), we get:
\begin{equation}
\frac{1}{2} \left( \frac{ds}{dt} \right)^2 \bm{q}'^\top \bm{M}(\bm{q}) \bm{q}' = H - P(\bm{q}).
\end{equation}
Solving for \( dt \) gives:
\begin{equation}
dt = \sqrt{ \frac{ \bm{q}'^\top \bm{M}(\bm{q}) \bm{q}' }{ 2(H - P(\bm{q})) } } \, ds.
\end{equation}
The action integral becomes:
\begin{equation}
\label{eq:jacobi_action}
\begin{aligned}
S_{M}(\bm{q}) 
&= \int \bm{p}^\top d\bm{q} 
= \int \bm{p}^\top \bm{q}' \, ds 
= \int \dot{\bm{q}}^\top \bm{M}(\bm{q}) \bm{q}' \, ds \\
&= \int_{s_1}^{s_2} \left( \bm{q}' \frac{ds}{dt} \right)^\top \bm{M}(\bm{q}) \bm{q}' \, ds \\
&= \int_{s_1}^{s_2} \left( \frac{ds}{dt} \right) \bm{q}'^\top \bm{M}(\bm{q}) \bm{q}' \, ds.
\end{aligned}
\end{equation}
Substituting the expression for \( dt \), we have:
\begin{equation}
\begin{aligned}
\label{eq:jacobi_metric_eq}
S_{M}(\bm{q}) &= \int_{s_1}^{s_2} \sqrt{ \bm{q}'^\top \left( 2(H - P(\bm{q})) \bm{M}(\bm{q}) \right) \bm{q}' } \, ds \\ 
&= \int_{s_1}^{s_2} \left\| \bm{q}' \right\|_{\bm{G}_\text{Jac}(\bm{q})} \, ds.
\end{aligned}
\end{equation}
Thus, energy-conserving paths correspond to geodesics under the Jacobi metric \( \bm{G}_\text{Jac}(\bm{q}) \), offering a geometric interpretation of conservative dynamics \citep{casetti2000geometric}.

\subsection{Geodesics as Optimal Control Solutions}
\label{sec:control_effort_geodesics}
We now explore the geometric connection between energy conservation and optimal control strategies that minimize control input. Specifically, we show how minimizing control input leads to energy-preserving trajectories consistent with geodesic motion under the Jacobi metric. The time derivative of the Hamiltonian expression in \eqref{eq:total_energy} is
\begin{equation}
\label{eq:hamiltonian_derivative}
\begin{aligned}
\dot{H} 
&= \frac{d}{dt} \left( \frac{1}{2} \dot{\bm{q}}^\top \bm{M}(\bm{q}) \dot{\bm{q}} \right) + \frac{d}{dt} P(\bm{q}) \\
&= \dot{\bm{q}}^\top \bm{M}(\bm{q}) \ddot{\bm{q}} + \frac{1}{2} \dot{\bm{q}}^\top \dot{\bm{M}}(\bm{q}) \dot{\bm{q}} + \dot{\bm{q}}^\top \nabla_{\bm{q}} P(\bm{q}).
\end{aligned}
\end{equation}
Substituting the dynamics from \eqref{eq:eom} with external torque $\bm{\tau}$:
\begin{equation}
    \bm{M}(\bm{q}) \ddot{\bm{q}} + \bm{C}(\bm{q}, \dot{\bm{q}}) \dot{\bm{q}} + \bm{g}(\bm{q}) = \bm{\tau},
\end{equation}
we obtain:
\begin{equation}
\begin{aligned}
\dot{H} 
&= \dot{\bm{q}}^\top \left[ \bm{\tau} - \bm{C}(\bm{q}, \dot{\bm{q}}) \dot{\bm{q}} - \bm{g}(\bm{q}) \right] 
+ \frac{1}{2} \dot{\bm{q}}^\top \dot{\bm{M}}(\bm{q}) \dot{\bm{q}} 
+ \dot{\bm{q}}^\top \bm{g}(\bm{q}) \\
&= \dot{\bm{q}}^\top \bm{\tau} - \dot{\bm{q}}^\top \bm{C}(\bm{q}, \dot{\bm{q}}) \dot{\bm{q}} + \frac{1}{2} \dot{\bm{q}}^\top \dot{\bm{M}}(\bm{q}) \dot{\bm{q}}.
\end{aligned}
\end{equation}
Using the skew-symmetry property for rigid-body systems~\citep{spong2006robot}
\begin{equation}
    \dot{\bm{q}}^\top \left( \dot{\bm{M}}(\bm{q}) - 2\bm{C}(\bm{q}, \dot{\bm{q}}) \right) \dot{\bm{q}} = 0,
\end{equation}
we conclude that
\begin{equation}
    \dot{H} = \dot{\bm{q}}^\top \bm{\tau}.
\end{equation}
This equation represents the \textit{instantaneous power input} to the system. When \( \dot{H} = 0 \), the torque \( \bm{\tau} \) is orthogonal to the velocity \( \dot{\bm{q}} \), meaning it does not change total energy—only a redirection of motion, consistent with geodesic motion under the Jacobi metric~\eqref{eq:jacobi_metric}. For a proof of the skew-symmetry property, see Appendix \ref{sec:skew}.

From the perspective of classical optimal control, minimizing control effort is typically formulated as
\begin{equation}
\label{eq:min_tau}
    \min_{\bm{\tau}}  \int_0^T \bm{\tau}^\top \bm{R} \bm{\tau} \, \mathrm{d}t,
\end{equation}
where \( \bm{R} \succeq 0 \) is a symmetric weighting matrix. Most often, this matrix penalizes control effort uniformly in all directions, i.e., the identity matrix. However, this formulation does not distinguish between torque that alters system energy and torque that merely redirects motion. To reflect the underlying physics more faithfully, we can consider minimizing the square of the instantaneous power input
\begin{equation}
\label{eq:min_power}
    \min_{\bm{\tau}} \int_0^T \dot{H}^2 \, \mathrm{d}t = \min_{\bm{\tau}} \int_0^T \left( \dot{\bm{q}}^\top \bm{\tau} \right)^2 \, \mathrm{d}t.
\end{equation}

This criterion penalizes only the component of torque that performs mechanical work (i.e., the portion aligned with the system velocity). This aligns with the standard optimal control problem \eqref{eq:min_tau} with $\bm{R} = \dot{\bm{q}} \dot{\bm{q}}^\top$. In contrast to conventional \( \ell^2 \) minimization that treats all directions equally, this criterion is more natural and physically grounded. As a result, geodesics under the Jacobi metric naturally emerge as solutions to this optimal control problem, highlighting the deep connection between energy-efficient control and geometric mechanics.

\section{Derivation of Skew-Symmetric Matrix} 
\label{sec:skew}
Assume the $(i,j)$-th element of the matrix \( \dot{\bm{M}}(\bm{q}) - 2\bm{C}(\bm{q}, \dot{\bm{q}}) \) is denoted as \( n_{ij} \). From Equations~\eqref{eq:T_t} and~\eqref{eq:kinetic_el}, we obtain:
\begin{equation}
\begin{aligned}
n_{ij} &= \frac{\partial M_{ij}}{\partial q_k} \dot{q}_k - \left( \frac{\partial M_{ij}}{\partial q_k} + \frac{\partial M_{ik}}{\partial q_j} - \frac{\partial M_{jk}}{\partial q_i} \right) \dot{q}_k \\
&= \left( \frac{\partial M_{jk}}{\partial q_i} - \frac{\partial M_{ik}}{\partial q_j} \right) \dot{q}_k.
\end{aligned}
\end{equation}
By interchanging the indices \( i \) and \( j \), we obtain:
\begin{equation}
n_{ji} 
= \left( \frac{\partial M_{ik}}{\partial q_j} - \frac{\partial M_{jk}}{\partial q_i} \right) \dot{q}_k = -n_{ij}.
\end{equation}
Therefore, the matrix \( \dot{\bm{M}}(\bm{q}) - 2\bm{C}(\bm{q}, \dot{\bm{q}}) \) is skew-symmetric.

\section{Laplace-Beltrami Operator}
\label{sec:laplace-beltrami operator}

The Laplace-Beltrami operator is a second-order differential operator that generalizes the classical Laplace operator from Euclidean space to Riemannian manifolds. This extension is essential for analyzing functions on curved spaces, as it accounts for the geometry of the manifold. The Laplace-Beltrami operator allows us to compute the divergence of the gradient in more general spaces and is given by the expression
\begin{equation}
\label{eq: laplace_beltrami}
       \Delta f = \frac{1}{\sqrt{|\bm{G}|}}\frac{\partial}{\partial_i} \left( \sqrt{|\bm{G}|}G^{ij} \frac{\partial}{\partial_j}f \right),
\end{equation}

where $G^{ij}$ are the components of the inverse of the metric tensor and $|\bm{G}|$ is the determinant of $\bm{G}$. 

An alternative, more compact form of the Laplace-Beltrami operator can be derived from \eqref{eq: laplace_beltrami}:
\begin{equation}
\label{eq: laplace_beltrami_compact} 
\resizebox{\linewidth}{!}{$
\begin{aligned}
\Delta f &= \frac{1}{\sqrt{|\bm{G}|}}\frac{\partial}{\partial_i} \left( \sqrt{|\bm{G}|}G^{ij} \frac{\partial}{\partial_j}f \right) \\
&= (\frac{\partial}{\partial_i} G^{ij}) \frac{\partial}{\partial_j} f + G^{ij} \frac{1}{\sqrt{|\bm{G}|}} \left(\frac{\partial}{\partial_i} \sqrt{|\bm{G}|} \, \frac{\partial}{\partial_j} f \right) + G^{ij} \frac{\partial^2}{\partial_i\partial_j} f \\
&= (\frac{\partial}{\partial_i} G^{ij}) \frac{\partial}{\partial_j} f + G^{ij} \frac{1}{\sqrt{|\bm{G}|}} \left( \frac{1}{2} (\frac{\partial}{\partial_i} G_{kl}) G_{kl} \sqrt{|\bm{G}|} \, \frac{\partial}{\partial_j} f \right) + G^{ij} \frac{\partial^2}{\partial_i\partial_j} f \\
&= (\frac{\partial}{\partial_i} G^{ij}) \frac{\partial}{\partial_j} f + \frac{1}{2} G^{ij} ( \frac{\partial}{\partial_i} G_{kl}) G^{kl}  \frac{\partial}{\partial_j} f + G^{ij} \frac{\partial^2}{\partial_i\partial_j} f \\
&= G^{ij} \frac{\partial^2}{\partial_i\partial_j}f - G^{ij} \bm{\Gamma}^k_{ij} \frac{\partial}{\partial_k} f \\
&= G^{ij} \left(\frac{\partial^2}{\partial_i\partial_j}f - \bm{\Gamma}^k_{ij} \frac{\partial}{\partial_k}f \right).
\end{aligned}
$}
\end{equation}

\section{Geometry-Aware Sampling}
\label{sec:geometry-aware-sampling}
We present the Riemannian Manifold Metropolis-Adjusted Langevin Algorithm (RM-MALA) for geometry-aware sampling on the Riemannian manifold. A detailed algorithm is shown in Algorithm \ref{alg:RMMALA}.

\subsection{Target Probability Density Function (PDF)}
In each tangent space, an infinitesimal space is induced by the Riemannian metric $
    d\mathcal{M}(\bm{q}) = \sqrt{\left| \bm{G}(\bm{q})\right|}d\bm{q}$,
bridging the target probability density function (PDF) $\rho(\bm{q})$ with respect the Lebesgue measurement $d\bm{q}$, to the PDF $p(\bm{q})$ with respect to $d\mathcal{M}(\bm{q})$ by
\begin{equation}
    \label{eq: bridge}
    \rho(\bm{q}) = p(\bm{q})\sqrt{\left| \bm{G}(\bm{q})\right|}.
\end{equation}

\subsection{Sampling on the Riemannian manifold}
The objective is to sample variables on the Riemannian manifold from the PDF $\rho(\bm{q})$, while taking into account the local geometric structure. Given the Riemannian metric tensor, we adopt the Metropolis-adjusted Langevin Monte Carlo algorithm on the Riemannian Manifold \citep{girolami2011riemann}. The algorithm describes the Langevin diffusion process on the Riemannian manifold in a stochastic differential equation (SDE)
\begin{equation}
    \label{eq: mala}
    d\bm{q}(t) = \frac{1}{2}\Tilde{\nabla}_{\bm{q}}\mathcal{L}(\bm{q}(t)) + d\Tilde{\bm{b}}(t),
\end{equation}
with $\Tilde{\nabla}_{\bm{q}}\mathcal{L}(\bm{q}(t))$ representing the natural gradient equipped by the Riemannian metric tensor $
\Tilde{\nabla}_{\bm{q}} \mathcal{L}(\bm{q}) = \bm{G}^{-1}(\bm{q}) \; \nabla_{\bm{q}}\mathcal{L}(\bm{q}),
$
where
\begin{equation}
    \label{eq: log_desire}
   \mathcal{L}(\bm{q}) = \log\rho(\bm{q}).
\end{equation}
The equation of the Brownian motion $d\Tilde{\bm{b}}(t)$ is given by
\begin{equation}
    \label{eq: brownian}
    \resizebox{\linewidth}{!}{$
    \begin{aligned}
    d\Tilde{\bm{b}}_i(t) &= \frac{1}{\sqrt{|\bm{G}(\bm{q}(t))|}} \sum_{j=1}^{D} \frac{\partial}{\partial \bm{q}_j} \left( (\bm{G}^{-1}(\bm{q}(t)))_{ij} \sqrt{|\bm{G}(\bm{q}(t))|} \right) dt \\
    &\quad + \left( \sqrt{\bm{G}^{-1}(\bm{q}(t))} d\bm{b}(t) \right)_{\!i},
    \end{aligned}
    $}
\end{equation}
where $\bm{b}(t)$ is the normal Brownian motion.
Assuming $p(\bm{q})$ is a constant, the natural gradient is expressed using \eqref{eq: bridge}:
\begin{equation}
    \label{eq: ng}
    \Tilde{\nabla}_{\bm{q}} \mathcal{L}(\bm{q}) = \bm{G}^{-1}(\bm{q}) \nabla_{\bm{q}} \sqrt{\left| \bm{G}(\bm{q})\right|}.
\end{equation}
In \eqref{eq: mala}, $d\Tilde{\bm{b}}(t)$ defines the Brownian motion on the Riemannian manifold.

After applying the first-order Euler integration with the fixed step size $\epsilon$ to the SDE \eqref{eq: mala}, we have
\begin{equation}
    \label{eq: euler}
    \bm{q}(t\!+\!1) = \bm{\mu}(\bm{q}(t), \epsilon) + \left(\epsilon \sqrt{\bm{G}^{-1}(\bm{q}(t))} \bm{z}(t)\right),
\end{equation}
where $\bm{\mu}(\bm{q}(t), \epsilon)$ is the mean of the Gaussian distribution associated with the sampled variable
\begin{equation}
    \label{eq: mean}
    \resizebox{\linewidth}{!}{$
    \begin{aligned}
    \bm{\mu}(\bm{q}(t), \epsilon) &= \bm{q}_{i}(t) + \frac{\epsilon^2}{2} \Big(\bm{G}^{-1}\big(\bm{q}(t)\big) \nabla_{\bm{q}} \mathcal{L}\big(\bm{q}(t)\big)\Big)_i  \\
&\quad - \epsilon^2 \sum_{j=1}^{D} \left( \Big(\bm{G}^{-1}\big(\bm{q}(t)\big)\Big)\frac{\partial \bm{G}\big(\bm{q}(t)\big)}{\partial \bm{q}_j} \bm{G}^{-1}\big(\bm{q}(t)\big) \right)_{ij} \\
&\quad + \frac{\epsilon^2}{2} \sum_{j=1}^{D} \Big(\bm{G}^{-1}\big(\bm{q}(t)\big)\Big)_{ij} \operatorname{Tr} \left( \Big(\bm{G}^{-1}\big(\bm{q}(t)\big)\Big) \frac{\partial G\big(\bm{q}(t)\big)}{\partial \bm{q}_j} \right) \\
    \end{aligned}
    $}
    \end{equation}
and $\bm{z}$ is a random variable sampled from the standard normal distribution $\bm{z} \sim \mathcal{N}\left(\bm{z}|\bm{0}, \bm{I}\right)$.

Finally, the probability of the sampled variable follows the Gaussian distribution
\begin{equation}
\label{eq: sample_dist}
p\left(\bm{q}(t\!+\!1)|\bm{q}(t)\right) \!= \!
\mathcal{N}\left(\bm{q}(t+1)|\bm{\mu}(\bm{q}(t), \epsilon), \;\epsilon^2 \bm{G}^{-1}\left(\bm{q}(t)\right)\right).
\end{equation} 
The acceptance of the sampled variable is finally calculated with
\begin{equation}
\label{eq: accept}
\begin{aligned}
    \alpha = &\mathcal{L}(\bm{q}(t+1)) + \log p(\bm{q}(t)|\bm{q}(t+1)) -  \mathcal{L}(\bm{q}(t)) \\
             & - \log p(\bm{q}(t+1)|\bm{q}(t)).
\end{aligned}
\end{equation}

\begin{algorithm}[t]
\caption{RM-MALA}
\label{alg:RMMALA}
\KwIn{
$\bm{G}(\bm{q})$: Riemannian metric function\\
$\mathcal{L}(\bm{q})$: Proposed likelihood function\\
$n_\text{burn}$: Burn-in steps \\
$n_\text{sample}$: Sample steps \\
$\epsilon$: Step size 
}
\KwOut{$Q$: A set of sampled points $\bm{q}$}

\textbf{Initialization:}\\
\textit{set} Random points $\bm{q} \in \left[-\pi, \pi \right)$\\

\textbf{Sampling Step:}\\
\For{i from 1 to $n_\text{burn} + n_\text{sample}$}{
Sample new point $\bm{q}_\text{new}$ through \eqref{eq: euler} \\
\If{$\bm{q}_\text{new} \notin \left[-\pi, \pi \right)$}{
         $\bm{q}_\text{new} = \text{arctan2}\left(\sin\left(\bm{q}_\text{new}\right), \cos\left(\bm{q}_\text{new}\right)\right)$
        }
Calculate the proposed log-likelihood $\log\mathcal{L}(\bm{q})$ and $\log\mathcal{L}(\bm{q}_\text{new})$ through \eqref{eq: log_desire} \\
Calculate the transition likelihood $p\left(\bm{q}_\text{new}|\bm{q}\right)$ and $p\left(\bm{q}|\bm{q}_\text{new}\right)$ through \eqref{eq: sample_dist} \\
Calculate the acceptance ratio $\alpha$ through \eqref{eq: accept}\\ 
Draw a random number $t \in [0, 1)$ \\
\If{$e^{\alpha} \geq t$}{
         $\bm{q} = \bm{q}_\text{new}$ \\
         \If{$i \ge n_\text{burn}$}{
        Add $\bm{q}_\text{new}$ into set $Q$
        }
        }
\Else{$\bm{q}_\text{new}$ is not accepted}
}
\end{algorithm}
\end{document}